%% file: thesis.tex
\documentclass[12pt,reqno,oneside]{amsbook}

\usepackage[english]{babel}  

\usepackage{amsmath, amssymb, amsthm}

\usepackage{listings} 

\usepackage{setspace}

\usepackage{graphicx}

\usepackage{booktabs}  

\usepackage{multirow}  

\usepackage{tabularx}

\usepackage{comment}

\usepackage{float} 

\usepackage{chngcntr}
\counterwithout{figure}{chapter}
\counterwithout{table}{chapter}

\makeatletter
\providecommand{\@dotsep}{2} 
\renewcommand{\@tocline}[7]{
  \relax
  \ifnum #1>\c@tocdepth\else
    \par \addpenalty\@secpenalty\addvspace{#2}%
    \begingroup \hyphenpenalty\@M
      \@ifempty{#4}{\@tempdima\csname r@tocindent\number#1\endcsname\relax}{\@tempdima#4\relax}%
      \parindent\z@ \leftskip#3\relax \advance\leftskip\@tempdima\relax
      \rightskip\@pnumwidth plus4em \parfillskip-\@pnumwidth
      #5\leavevmode\hskip-\@tempdima #6\nobreak
      \leaders\hbox{$\m@th\mkern \@dotsep mu.\mkern \@dotsep mu$}\hfill
      \hbox to\@pnumwidth{\@tocpagenum{#7}}\par
      \nobreak
    \endgroup
  \fi}
\makeatother

\usepackage{enumitem}


\usepackage{geometry}
\theoremstyle{definition}

\usepackage[pdfusetitle]{hyperref}

\begin{document}

\newcommand{\thetitle}{Spectral Geometry for Deep Learning: Compression and Hallucination Detection via Random Matrix Theory}

\newcommand{\institution}{University of Illinois Chicago} 
\newcommand{\degree}{Master of Science}
\newcommand{\programname}{Computer Science}
\newcommand{\committee}{Amit Ranjan Trivedi, Chair and Advisor \\ Sathya Ravi \\ Sourav Medya\\ Marco Brambilla, Politecnico di Milano}
\newcommand{\theauthor}{Davide Ettori}
\newcommand{\priordegrees}{B.S. in Computer Engineering, Politecnico di Milano, Milan, Italy, 2023}
\newcommand{\graduationyear}{2026}

\author{\theauthor}
\title{\thetitle}
\date{\today}

\frontmatter

\begin{titlepage}
    \centering
	{\Large {\textbf{\thetitle}}}  \par
    \vspace{2cm}
    {BY\par}
    \vspace{0.5cm}
    {\theauthor}\\
    {\priordegrees}\\
    \vspace{2cm}
	THESIS\\
	\vspace{1cm}
	Submitted as partial fulfillment of the requirements\\ \vspace{0.2cm}
	for the degree of {\degree} in {\programname} \\ \vspace{0.2cm}
	in the Graudate College of the \\  \vspace{0.2cm}
	University of Illinois Chicago, {\graduationyear}\\
	\vspace{1cm}
	Chicago, Illinois
		
	\raggedright
    \vfill
    Defense Committee: \\
    \committee \\
\end{titlepage}

\setcounter{page}{2} 

\input{sections/other/_0_accessibility_statement.tex}






\tableofcontents
\listoffigures
\listoftables

\input{sections/other/__abbrs.tex}

\input{sections/other/_4_summary}

\mainmatter

\doublespacing 

\input{sections/1_introduction/0_index}

\input{sections/2_background/0_index}

\input{sections/3_related/0_index}

\input{sections/4_methodology/0_index}

\input{sections/5_conclusion/0_index}

\input{sections/other/_7_appendix.tex}

\backmatter 
\singlespacing

\input{sections/other/_8_biblio.tex}
\end{document}

%% file: sections/other/_0_accessibility_statement.tex
\chapter*{Accessibility Statement}
An accessible EPUB version of this document can be obtained by contacting Davide Ettori at \href{mailto:detto3@uic.edu}{detto3@uic.edu}

%% file: sections/other/__abbrs.tex
%
\chapter*{List of Abbreviations}

\begin{description}

    \item[BBP] Baik–Ben Arous–Péché
\vspace{0.5cm}
    \item[CNN] Convolutional Neural Network
\vspace{0.5cm}
    \item[FNN] Feed Forward Neural Network
\vspace{0.5cm}
    \item[GRU] Gated Recurrent Unit
\vspace{0.5cm}
    \item[KD] Knowledge Distillation
\vspace{0.5cm}
    \item[KL] Kullback–Leibler
\vspace{0.5cm}
    \item[LLM] Large Language Model
\vspace{0.5cm}
    \item[LSTM] Long Short Term Memory
\vspace{0.5cm}
    \item[MP] Marčenko–Pastur
\vspace{0.5cm}
    \item[OOD] Out of Distribution
\vspace{0.5cm}
    \item[RMT] Random Matrix Theory
\vspace{0.5cm}
    \item[RNN] Recurrent Neural Network
\vspace{0.5cm}
    \item[TW] Tracy–Widom 

\end{description}

%% file: sections/other/_4_summary.tex
\chapter*{Summary}


The rapid growth of deep learning and large language models has transformed artificial intelligence across natural language processing, computer vision, and multimodal reasoning. These advances, however, come with two persistent challenges: reliability and efficiency. Models can produce hallucinations, misinterpret out-of-distribution inputs, and exhibit unstable behavior that undermines trust in high-stakes domains such as healthcare, law, and finance. At the same time, their increasing scale demands enormous computational resources, making them difficult to deploy efficiently and sustainably.

This thesis develops a unifying framework grounded in Spectral Geometry and Random Matrix Theory (RMT) to address both reliability and efficiency in modern deep learning systems. By studying the eigenvalue dynamics of hidden activations, we show that spectral statistics provide a compact and interpretable lens on model behavior, capable of distinguishing structured, causal representations from noise-like variability. Two core contributions are presented within this framework.

The first contribution, \textbf{EigenTrack}, is a real-time detector of hallucinations and out-of-distribution behavior in large language and vision-language models. EigenTrack converts streaming hidden activations into spectral signatures, entropy, eigenvalue gaps, and divergence from the Marchenko–Pastur baseline, and models their temporal evolution with lightweight recurrent classifiers. This approach enables early detection of reliability failures before they manifest in model outputs, while offering interpretable insights into the underlying representation dynamics. EigenTrack achieves state-of-the-art performance across diverse architectures, demonstrating that spectral features provide stable and generalizable signals for monitoring model trustworthiness.

The second contribution, \textbf{RMT-KD}, is a principled method for compressing deep networks through random matrix theoretic knowledge distillation. By identifying outlier eigenvalues in activation spectra as carriers of causal information, RMT-KD progressively projects networks onto lower-dimensional subspaces while preserving accuracy through iterative self-distillation. This process yields models that are significantly more compact and energy-efficient, yet remain dense and hardware-friendly. RMT-KD attains state-of-the-art trade-offs between compression and accuracy, outperforming traditional pruning and heuristic low-rank methods by relying on statistically grounded criteria rather than ad hoc thresholds.

Together, these contributions establish spectral geometry as a coherent foundation for both diagnosing uncertainty and guiding compression in large-scale neural networks. By linking eigenvalue statistics to representation quality, the thesis demonstrates how Random Matrix Theory can provide interpretable, mathematically principled tools to make large language models and related architectures more reliable, efficient, and trustworthy. This spectral perspective opens new avenues for understanding deep learning systems, bridging theoretical insights with practical methods for safe and sustainable AI deployment.

%% file: sections/1_introduction/0_index.tex
\chapter{Introduction}
\label{chap:introduction}
In recent years, deep learning and large language models have become central to advances in artificial intelligence. Transformer-based architectures now underpin applications ranging from natural language understanding and text generation to computer vision and multimodal reasoning. These models have shown remarkable capabilities, often surpassing human performance on benchmark tasks, and are increasingly being adopted in domains such as healthcare, law, finance, and education. Their rapid adoption highlights both their transformative potential and the growing reliance of modern society on machine intelligence.

However, the success of these systems comes with critical challenges. On the one hand, large language models often suffer from reliability issues: they can produce hallucinations, generate plausible but incorrect outputs, and fail dramatically when faced with out-of-distribution inputs. Such behavior undermines trust in AI, particularly in safety-critical contexts where decisions must be both accurate and justifiable. On the other hand, the efficiency of these models is a major obstacle. State-of-the-art architectures contain billions of parameters, requiring enormous computational resources for training and inference. This makes them difficult to deploy on edge devices, costly to maintain in large-scale production, and unsustainable in terms of energy consumption.

Addressing these dual challenges of reliability and efficiency requires new methods that are not only empirically effective but also grounded in mathematical principles. One promising perspective arises from spectral geometry and Random Matrix Theory (RMT), which provide a rigorous way to analyze the eigenvalue spectra of hidden activations in neural networks. Eigenvalue statistics can reveal whether a model is encoding structured, causal information or drifting toward noise-like behavior, offering a compact and interpretable description of its internal dynamics. By leveraging these spectral insights, it becomes possible to both diagnose failures in real time and guide principled approaches to model compression.

This thesis develops such a spectral framework for modern deep learning and large language models. It introduces two contributions built on Random Matrix Theory: EigenTrack, a method for real-time hallucination and out-of-distribution detection, and RMT-KD, a knowledge distillation approach for efficient network compression. Together, these methods demonstrate that spectral analysis of neural representations can provide a unifying foundation for making AI systems simultaneously more trustworthy and more efficient.

\input{sections/1_introduction/1_goals}
\input{sections/1_introduction/2_structure}

%% file: sections/1_introduction/1_goals.tex
\section{Goals and Research Questions}
\label{intro:goals}

The inspiration for this research arises from prior work conducted at the AEON Lab at the University of Illinois Chicago, where spectral methods were first explored as a tool to understand the behavior of large neural networks. Building on this foundation, the present thesis investigates how spectral geometry and Random Matrix Theory can provide interpretable, mathematically grounded solutions to the pressing challenges of modern artificial intelligence. The need for such approaches is clear: large language models and other deep architectures are increasingly deployed in critical applications, yet they remain prone to hallucinations, brittle under distributional shifts, and prohibitively costly to scale. These limitations highlight the necessity of frameworks that not only improve performance but also enhance trustworthiness and efficiency in a principled manner.

The central goals of this thesis are therefore twofold. First, to develop methods that detect and anticipate failures in large language and vision-language models by analyzing the spectral dynamics of their hidden activations. Second, to design compression strategies that reduce the computational burden of deep networks without sacrificing accuracy, guided by rigorous statistical laws rather than heuristics. These goals naturally lead to the following research questions: Can spectral geometry serve as a compact and interpretable signal of reliability in large-scale models? And can Random Matrix Theory provide a principled foundation for identifying and retaining the causal structure necessary for efficient compression? Addressing these questions is the focus of this thesis, with the broader objective of advancing the reliability and sustainability of modern deep learning.

%% file: sections/1_introduction/2_structure.tex
\section{Thesis Structure}
\label{intro:struc}

In Chapter 2 we provide the necessary background to this research, reviewing the theoretical foundations of spectral geometry, random matrix theory, and the fundamentals of deep neural networks and large language models. This chapter introduces the mathematical tools that will be applied throughout the thesis.  
Chapter 3 surveys related work, presenting prior approaches to hallucination detection, out-of-distribution recognition, and model compression. We compare black-box, grey-box, and white-box detection methods, as well as traditional compression techniques such as pruning, low-rank approximations, and standard knowledge distillation.  
Chapter 4 introduces the first main contribution of this thesis, \textit{EigenTrack}, and describes in detail its methodology, design, and experimental evaluation for hallucination and out-of-distribution detection in large language and vision-language models.  
Chapter 5 presents the second contribution, \textit{RMT-KD}, and explains how Random Matrix Theory can be used to guide causal knowledge distillation for efficient network compression. This chapter also discusses its empirical results across natural language processing and computer vision benchmarks.  
Finally, Chapter 6 concludes the thesis by summarizing the key contributions, outlining limitations, and suggesting directions for future research at the intersection of spectral theory and deep learning.

%% file: sections/2_background/0_index.tex
\chapter{Background}
\label{chap:background}

In this chapter we review the theoretical and technological foundations of our work. We first outline modern deep learning architectures, focusing on large language models, convolutional networks, and vision-language systems. We then introduce the mathematical basis of spectral geometry and random matrix theory. Next, we discuss two central challenges in today’s models: reliability, including hallucinations and out-of-distribution errors, and efficiency, with emphasis on compression and knowledge distillation. We also survey prior applications of spectral methods in vision and language models, and conclude with a unified perspective that links efficiency and reliability through spectral geometry.

\input{sections/2_background/1_foundations}

\input{sections/2_background/2_spectral}

\input{sections/2_background/3_rmt}

\input{sections/2_background/4_hallucination}

\input{sections/2_background/5_compression}

\input{sections/2_background/6_spectral}

\input{sections/2_background/7_bridge}

%% file: sections/2_background/1_foundations.tex
\section{Background: Foundations of Deep Learning Architectures}
\label{background:foundations_deep_learning}

\subsection*{From Machine Learning to Deep Learning}

Machine Learning (ML) develops algorithms that learn patterns from data rather than following explicit rules. Classical ML methods such as decision trees, support vector machines, and k-nearest neighbors rely on handcrafted features and work well on structured data. Deep Learning (DL), a subfield of ML, uses deep neural networks to automatically learn hierarchical representations from raw inputs. Enabled by large datasets and GPU computing, DL has driven breakthroughs in computer vision, natural language processing, and multimodal tasks \cite{lecun2015deep,goodfellow2016deep}.

\subsection{Neural Networks and Representation Learning}

Feed Forward Neural Networks (FNNs) consist of only linear layers of interconnected neurons, defined as $h^{(l)} = \sigma(W^{(l)} h^{(l-1)} + b^{(l)})$, that apply weighted sums and nonlinear activations \cite{goodfellow2016deep}. Training with backpropagation allows them to approximate highly complex functions. Hidden layers serve as representations, mapping raw data to informative latent features. Modern networks are usually overparameterized, with more parameters than training examples. Surprisingly, this often improves generalization, stability, and representation power \cite{zhang2017understanding,belkin2019reconciling}.

\begin{figure}[h]
    \centering
    \includegraphics[width=0.75\linewidth]{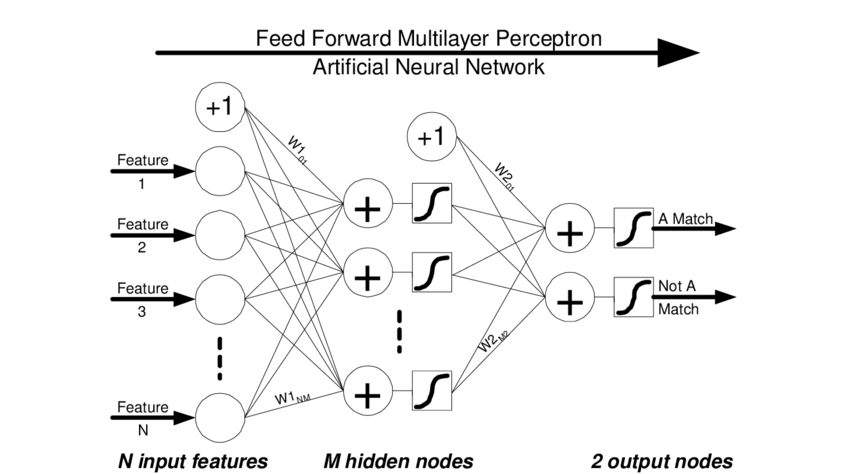}
    \caption[Neural network (NN)]{Neural network (NN): features pass through hidden layers and nonlinear activations, producing the output.}
    \label{fig:nn_arch}
\end{figure}

\subsection*{Convolutional Neural Networks (CNNs)}

CNNs transformed computer vision by exploiting spatial locality with many convolutional filters following this equation $h_{i,j}^{(l)} = \sigma\!\left(\sum_{m,n} W_{m,n}^{(l)} \, x_{i+m,\,j+n}^{(l-1)} + b^{(l)}\right)$. Stacked convolution and pooling layers extract increasingly abstract features, while linear layers perform classification. Landmark models such as AlexNet, VGG, and ResNet demonstrated the scalability of CNNs \cite{krizhevsky2012imagenet,he2016deep}.

\begin{figure}[h]
    \centering
    \includegraphics[width=0.75\linewidth]{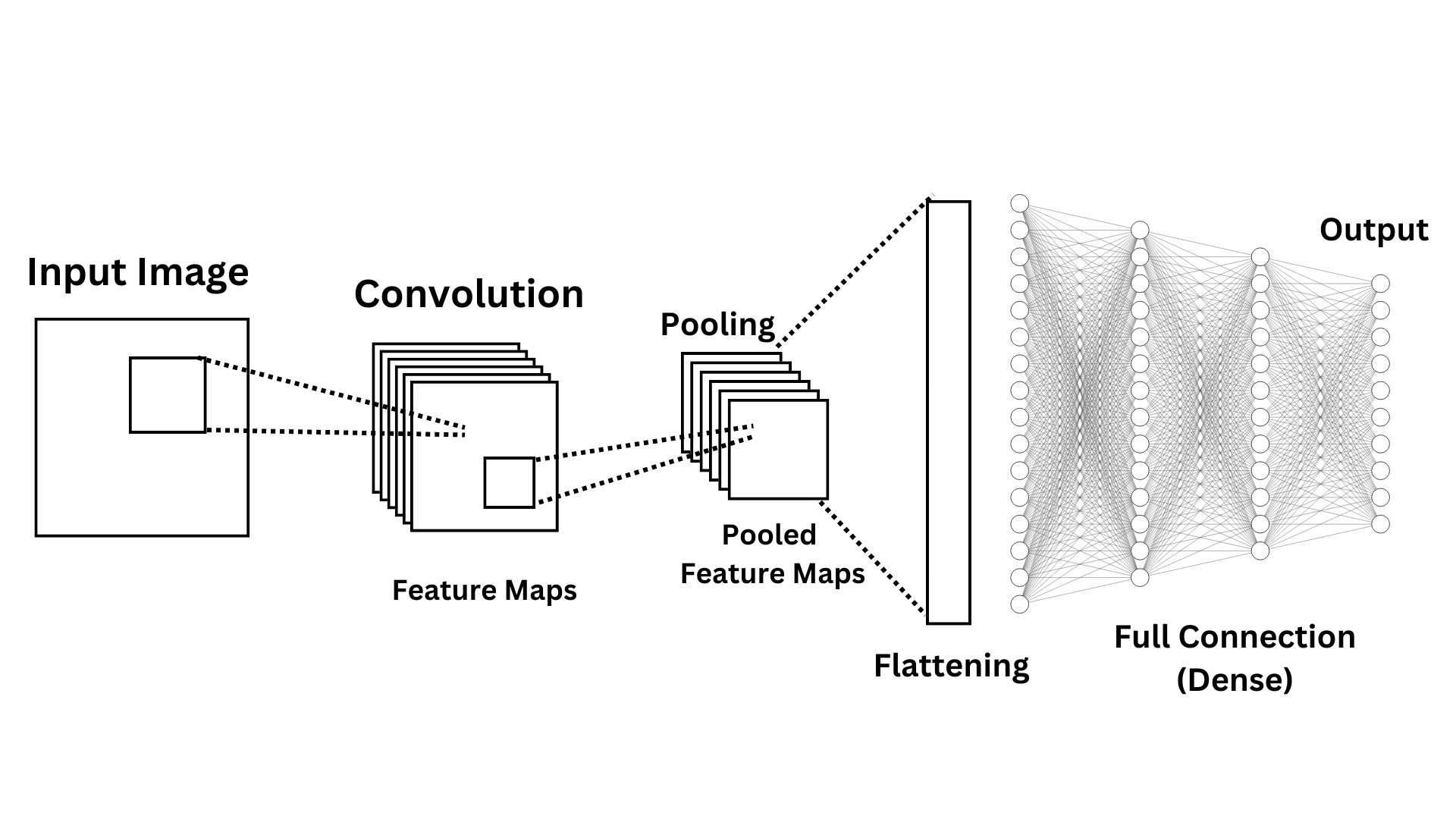}
    \caption[CNN architecture]{CNN: convolutional filters extract features, pooling layers reduce resolution, and linear layers classify. (Stanford CS231 Notes)}
    \label{fig:cnn_arch}
\end{figure}

\subsection*{Recurrent Neural Networks (RNNs)}
\label{background:rnn}

Recurrent neural networks (RNNs) are designed to process sequential data by maintaining a hidden state that evolves over time. Given an input sequence $\{x_t\}_{t=1}^T$, a vanilla RNN updates its hidden state as $h_t = \tanh(W_{xh} x_t + W_{hh} h_{t-1} + b_h)$, and produces an output $y_t = W_{hy} h_t + b_y$. While conceptually simple, standard RNNs suffer from vanishing and exploding gradients, which limits their ability to capture long-range dependencies.

To address this, gated architectures such as the Gated Recurrent Unit (GRU) and Long Short-Term Memory (LSTM) were introduced. A GRU introduces two gates: the reset gate $r_t = \sigma(W_r x_t + U_r h_{t-1} + b_r)$ and the update gate $z_t = \sigma(W_z x_t + U_z h_{t-1} + b_z)$. The candidate hidden state is $\tilde{h}_t = \tanh(W_h x_t + U_h (r_t \odot h_{t-1}) + b_h)$, and the final hidden state is $h_t = (1 - z_t) \odot h_{t-1} + z_t \odot \tilde{h}_t$. These mechanisms allow the GRU to selectively retain or overwrite information over time.

The LSTM further extends this idea by maintaining a separate memory cell $c_t$ alongside the hidden state $h_t$. It employs three gates: the input gate $i_t = \sigma(W_i x_t + U_i h_{t-1} + b_i)$, the forget gate $f_t = \sigma(W_f x_t + U_f h_{t-1} + b_f)$, and the output gate $o_t = \sigma(W_o x_t + U_o h_{t-1} + b_o)$. The memory cell is updated as $c_t = f_t \odot c_{t-1} + i_t \odot \tanh(W_c x_t + U_c h_{t-1} + b_c)$, and the hidden state is given by $h_t = o_t \odot \tanh(c_t)$. By explicitly controlling how information is added, forgotten, and exposed, LSTMs can capture dependencies over long sequences more effectively than vanilla RNNs or GRUs.

\subsection{Transformers and Large Language Models}

Transformers, introduced by Vaswani et al. \cite{vaswani2017attention}, rely on self-attention to capture long-range dependencies without recurrence. The attention mechanism computes similarity scores between all pairs of tokens in a sequence, allowing each token to weight and aggregate information from every other token. This enables the model to focus on relevant context regardless of distance. Their scalability has enabled Large Language Models (LLMs) such as BERT \cite{devlin2019bert}, GPT \cite{radford2019language}, and LLaMA \cite{touvron2023llama}. This architecture has been shown to be effective also for multimodal data.

\begin{figure}[h]
    \centering
    \includegraphics[width=0.75\linewidth]{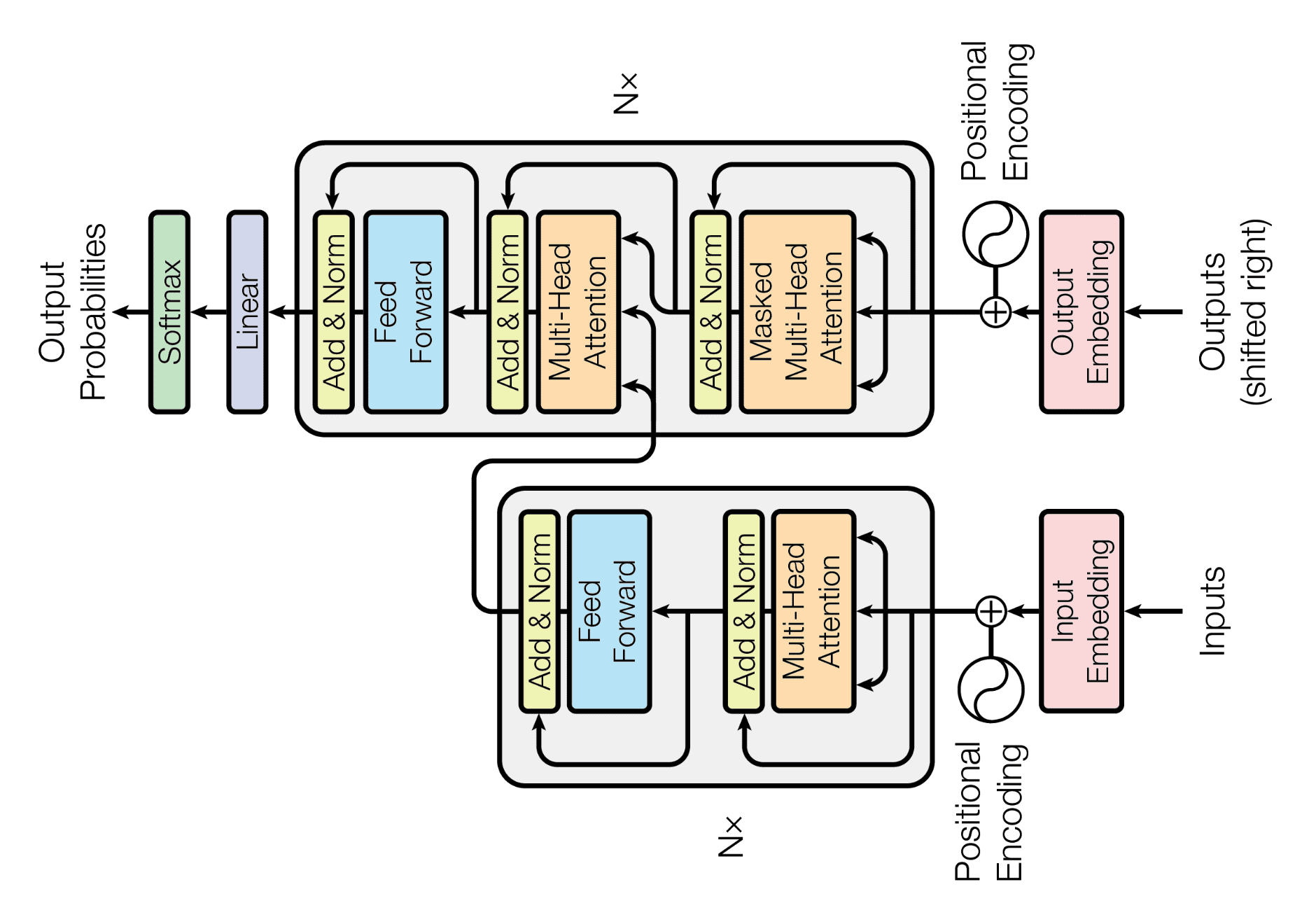}
    \caption[Transformer architecture]{Transformer architecture \protect\cite{vaswani2017attention}. Each encoder–decoder block is composed of multi-head self-attention layers that model contextual dependencies, followed by position-wise feed-forward networks. Residual connections and layer normalization ensure stable training and effective gradient flow, enabling scalability to very large models.}
    \label{fig:transformer_architecture}
\end{figure}

\subsection*{Vision-Language Models (VLMs)}

VLMs integrate image encoders (CNNs or Vision Transformers) with text encoders (Transformers) to align visual and textual modalities. This enables tasks such as image captioning, visual question answering, and cross-modal retrieval. CLIP \cite{radford2021learning} and LLaVA \cite{liu2024llava} are prominent examples.

\begin{figure}[h]
    \centering
    \includegraphics[width=0.75\linewidth]{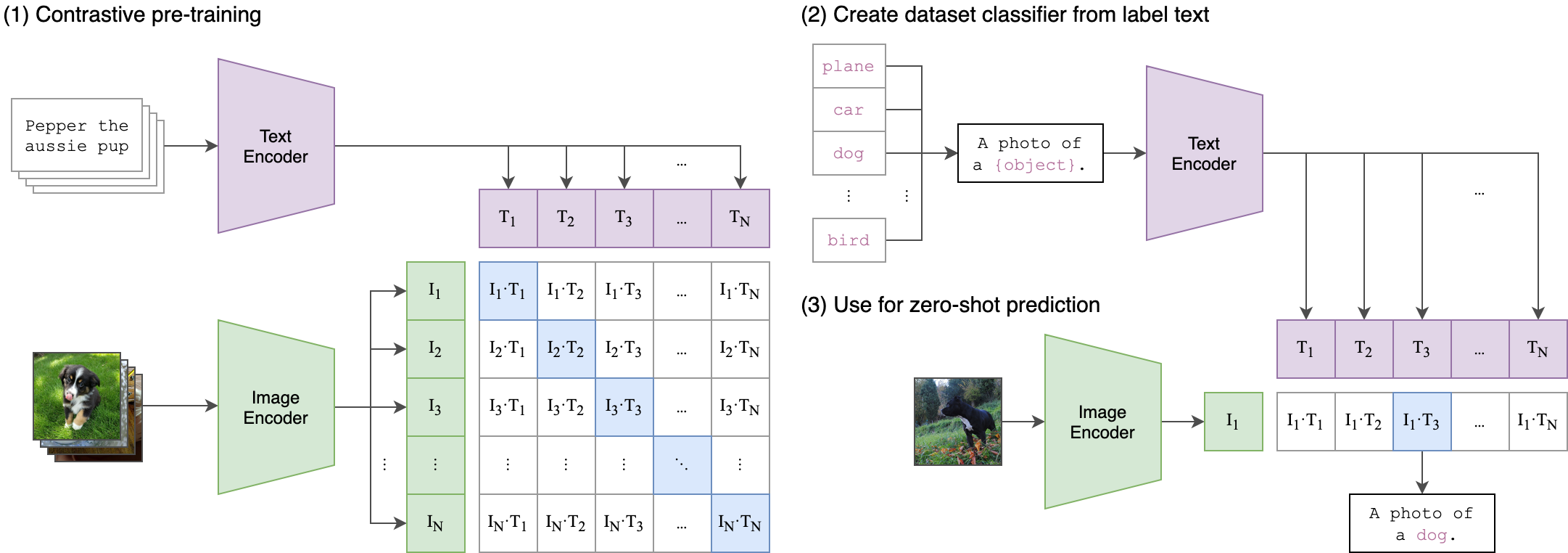}
    \caption[Vision-language model (VLM) schematic]{Vision-language model (VLM) schematic. CLIP (Contrastive Language–Image Pretraining) \protect\cite{radford2021learning} aligns images and text by projecting them into a shared latent space. The architecture consists of two encoders: a Vision Transformer (or CNN) for images and a Transformer for text. During training, the model learns to maximize the cosine similarity between embeddings of matching image-text pairs while minimizing it for mismatched pairs, using a contrastive loss. This enables zero-shot classification by comparing the similarity of an image's embedding to embeddings of text descriptions of potential classes. CLIP's ability to generalize to unseen tasks without fine-tuning has made it a cornerstone for multimodal AI.}
    \label{fig:vlm_architecture}
\end{figure}

\subsection*{The Double-Edged Sword of Overparameterization}

Overparameterization is a typical trait of modern neural networks. It enables models to interpolate training data while still generalizing well, a phenomenon described by the `double descent'' curve \cite{belkin2019reconciling}, but also brings inefficiency, high energy cost, and deployment challenges. From the perspective of the bias--variance trade-off \cite{geman1992neural}, increasing complexity should reduce bias but raise variance, implying an optimal model size. Yet deep networks often challenge this intuition: despite extreme overparameterization, they can achieve both low bias and low variance with a sufficient amount of training data. Explaining this paradox remains an active area of research, with spectral methods offering one promising approach.

%% file: sections/2_background/2_spectral.tex
\section{Background: Mathematical Foundations of Spectral Geometry}
\label{background:math_spectral_geometry}

\subsection{Eigenvalues and Eigenvectors Foundations}

At the foundation of spectral geometry are \textbf{eigenvalues} and \textbf{eigenvectors}.  
For a square matrix $A \in \mathbb{R}^{n \times n}$, an eigenvector $v \neq 0$ and corresponding eigenvalue $\lambda \in \mathbb{R}$ satisfy the relation $A v = \lambda v$. This identifies directions in which the linear transformation $A$ acts by pure scaling, without rotation or distortion. The scalar $\lambda$ measures the amount of stretching or compression along the eigenvector. Eigenvalues are obtained by solving the characteristic equation $\det(A - \lambda I) = 0$, whose roots $\lambda_1, \ldots, \lambda_n$ represent the eigenvalues of $A$ \cite{horn2012matrix}. In practice, direct computation from the determinant is numerically unstable for large matrices, and one typically relies on methods such as QR decomposition, singular value decomposition (SVD), or iterative algorithms like the Lanczos procedure and power iteration \cite{golub1996matrix}.

\subsection{Spectral Decomposition}

For symmetric matrices, common in machine learning (e.g., covariance matrices), the spectral theorem guarantees a decomposition $A = Q \Lambda Q^T$, where $Q$ contains the orthogonal eigenvectors and $\Lambda$ is a diagonal matrix of eigenvalues. Thus, any symmetric transformation can be expressed as scaling along orthogonal directions \cite{horn2012matrix}.

\subsection*{Interpretation of Eigenvalues}

Eigenvalues provide deep geometric and statistical insight. Geometrically, large values correspond to directions in which vectors are stretched, while small values correspond to nearly collapsed directions. For covariance matrices, eigenvalues quantify the variance of the data along the associated eigenvector directions. A few dominant eigenvalues indicate that most variability lies in a low-dimensional subspace, while a flatter spectrum reflects variance distributed across many directions. Hence, eigenvalues reveal intrinsic dimensionality and help distinguish meaningful structure from noise in high-dimensional representations \cite{trefethen1997numerical}.

\subsection{Dimensionality Reduction with Eigenvalues}

An important application of eigenvalues is dimensionality reduction, especially PCA (Principal Component Analysis). Consider a dataset $X \in \mathbb{R}^{n \times d}$, where each row is a data point in $d$ dimensions. After centering the data, the covariance matrix is computed as $C = \tfrac{1}{n} X^T X \in \mathbb{R}^{d \times d}$. Let $\lambda_1 \geq \lambda_2 \geq \cdots \geq \lambda_d$ be the eigenvalues of $C$ with corresponding orthonormal eigenvectors $v_1, \ldots, v_d$. To reduce the dimensionality to $k < d$, we construct the projection matrix $V_k = [v_1 \; v_2 \; \cdots \; v_k] \in \mathbb{R}^{d \times k}$ by concatenating the top-$k$ eigenvectors. The reduced representation of the data is then $X_{red} = X V_k \in \mathbb{R}^{n \times k}$. This projection maximizes the preserved variance, given by $\text{Var}(X_{red}) = \sum_{i=1}^k \lambda_i$, which is the largest possible among all $k$-dimensional linear projections. A more efficient approach uses SVD (Singula Values Decomposition) directly on $X$ to obtain the principal components without explicitly forming the covariance matrix.

%% file: sections/2_background/3_rmt.tex
\section{Background: Random Matrix Theory}
\label{background:rmt}

\subsection{Introduction to Random Matrix Theory}

Random Matrix Theory (RMT) studies the spectral properties of large matrices whose entries are random variables. Its central goal is to characterize the empirical spectral distribution (ESD), defined as the normalized histogram of eigenvalues of a random matrix as its dimensions grow. Let $\mathbf{M} \in \mathbb{R}^{p \times p}$ be a random symmetric matrix with eigenvalues $\lambda_1, \ldots, \lambda_p$. The empirical spectral distribution is:
$\mu_{\mathbf{M}}(\lambda) = \frac{1}{p} \sum_{i=1}^p \delta(\lambda - \lambda_i)$ where $\delta$ is the Dirac delta. As $p \to \infty$, $\mu_{\mathbf{M}}$ converges (in probability) to a deterministic limiting law, such as the Wigner semicircle or Marchenko–Pastur distribution, depending on the matrix ensemble.

\begin{figure}[h]
    \centering
    \includegraphics[width=0.75\linewidth]{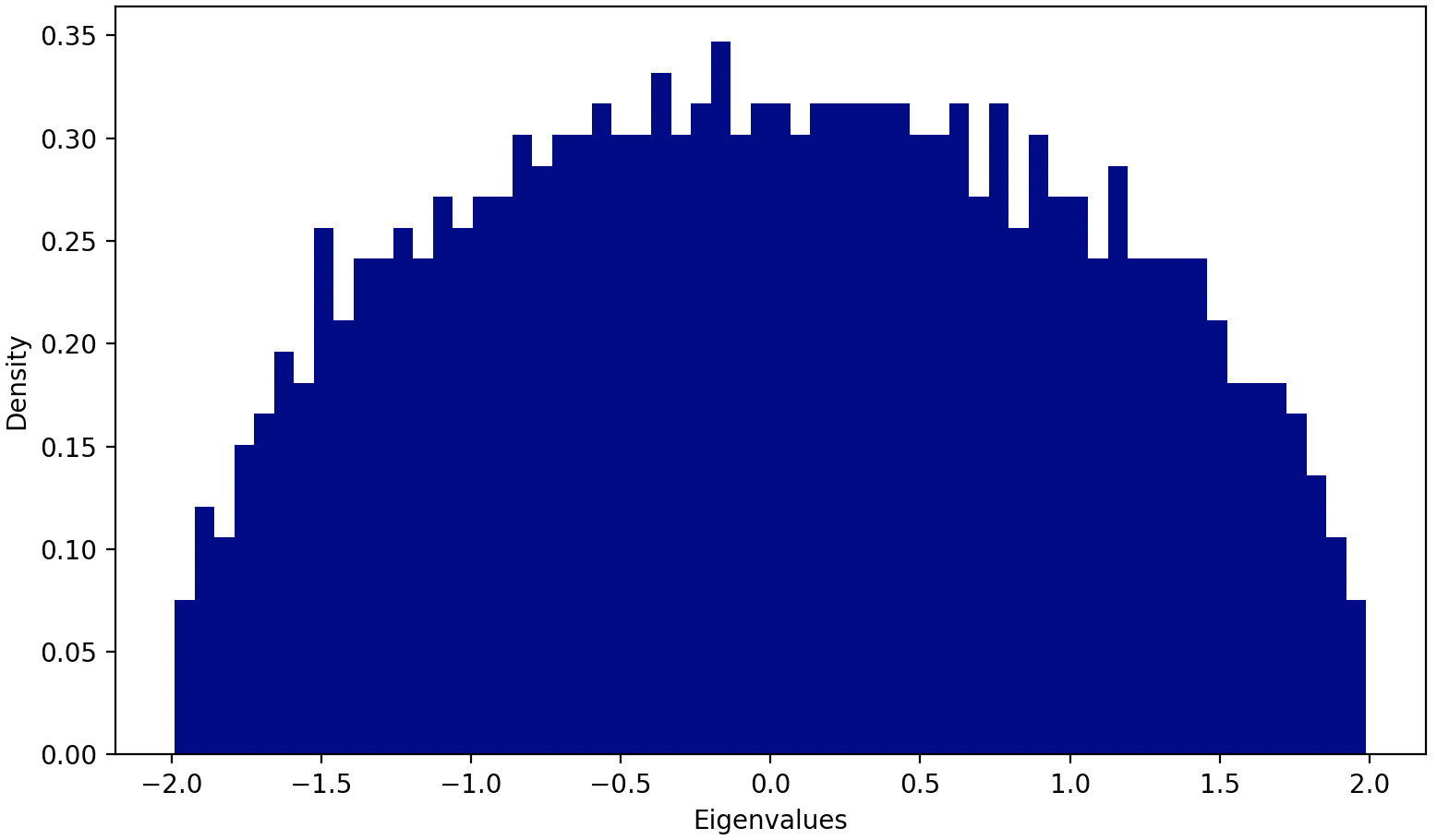}
    \caption[Wigner semicircle law]{Wigner semicircle law: eigenvalue density of large symmetric random matrices.}
    \label{fig:wigner}
\end{figure}

\begin{figure}[h]
    \centering
    \includegraphics[width=0.75\linewidth]{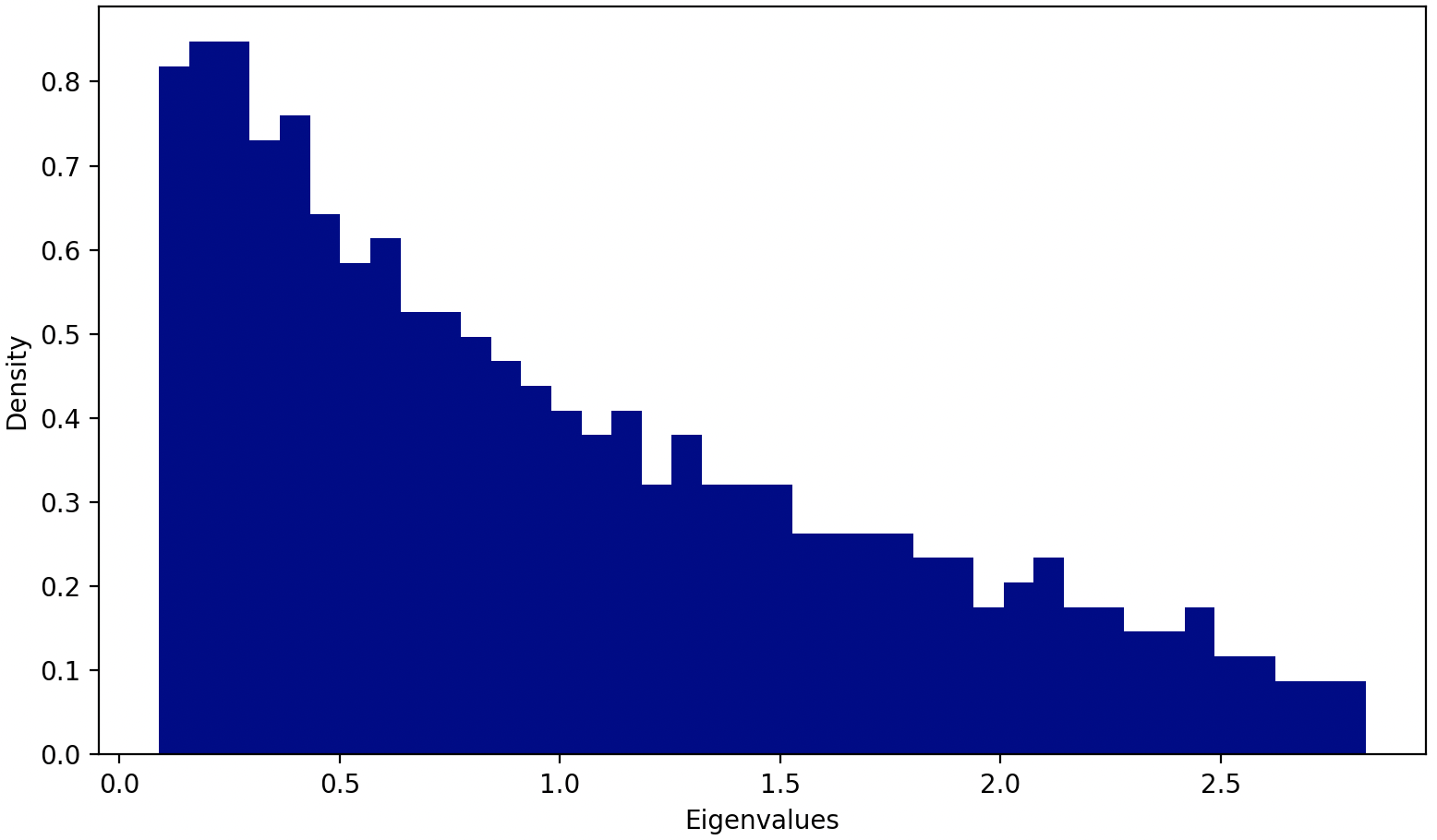}
    \caption[Marchenko–Pastur distribution]{Marchenko–Pastur distribution: eigenvalue density of sample covariance matrices.}
    \label{fig:mp}
\end{figure}

\subsection*{Wigner Semicircle Law}

The Wigner Semicircle Law describes the eigenvalue density of large symmetric matrices with i.i.d. entries of mean zero and variance $\sigma^2$ \cite{chan1992wigner}.  
Let $\mathbf{W} \in \mathbb{R}^{p \times p}$ with entries $W_{ij}$ drawn i.i.d. and $\mathbf{W} = \mathbf{W}^T$. Then, as $p \to \infty$, the ESD converges to:
$f_{\text{Wigner}}(\lambda) = \frac{1}{2 \pi \sigma^2} \sqrt{4\sigma^2 - \lambda^2} \ \text{for } |\lambda| \leq 2\sigma,\ \text{and } 0 \text{ otherwise}$ . This implies that the eigenvalues of a purely random symmetric matrix are asymptotically confined to the interval $[-2\sigma, 2\sigma]$, establishing a universal ``noise floor.''  

\subsection*{Marchenko–Pastur Distribution}

For covariance-type matrices, the limiting spectrum is described by the Marchenko–Pastur (MP) law \cite{yaskov2016short}. Let $\mathbf{X} \in \mathbb{R}^{n \times p}$ have i.i.d. entries with mean zero and variance $\sigma^2$, and define the sample covariance matrix $\mathbf{C} = \frac{1}{n} \mathbf{X}^T \mathbf{X}$.  
If $n, p \to \infty$ with ratio $p/n \to c > 0$, then the ESD of $\mathbf{C}$ converges to:
$f_{\text{MP}}(\lambda) = \frac{1}{2\pi c \sigma^2 \lambda} \sqrt{(\lambda_+ - \lambda)(\lambda - \lambda_-)}, \quad \lambda \in [\lambda_-, \lambda_+]$  with support 
$\lambda_{\pm} = \sigma^2 (1 \pm \sqrt{c})^2$. Thus, all eigenvalues are confined within $[\lambda_-, \lambda_+]$, forming the ``bulk spectrum'' of random covariance matrices.

\subsection*{Tracy-Widom Distribution}
\begin{figure}[h]
    \centering
    \includegraphics[width=0.75\linewidth]{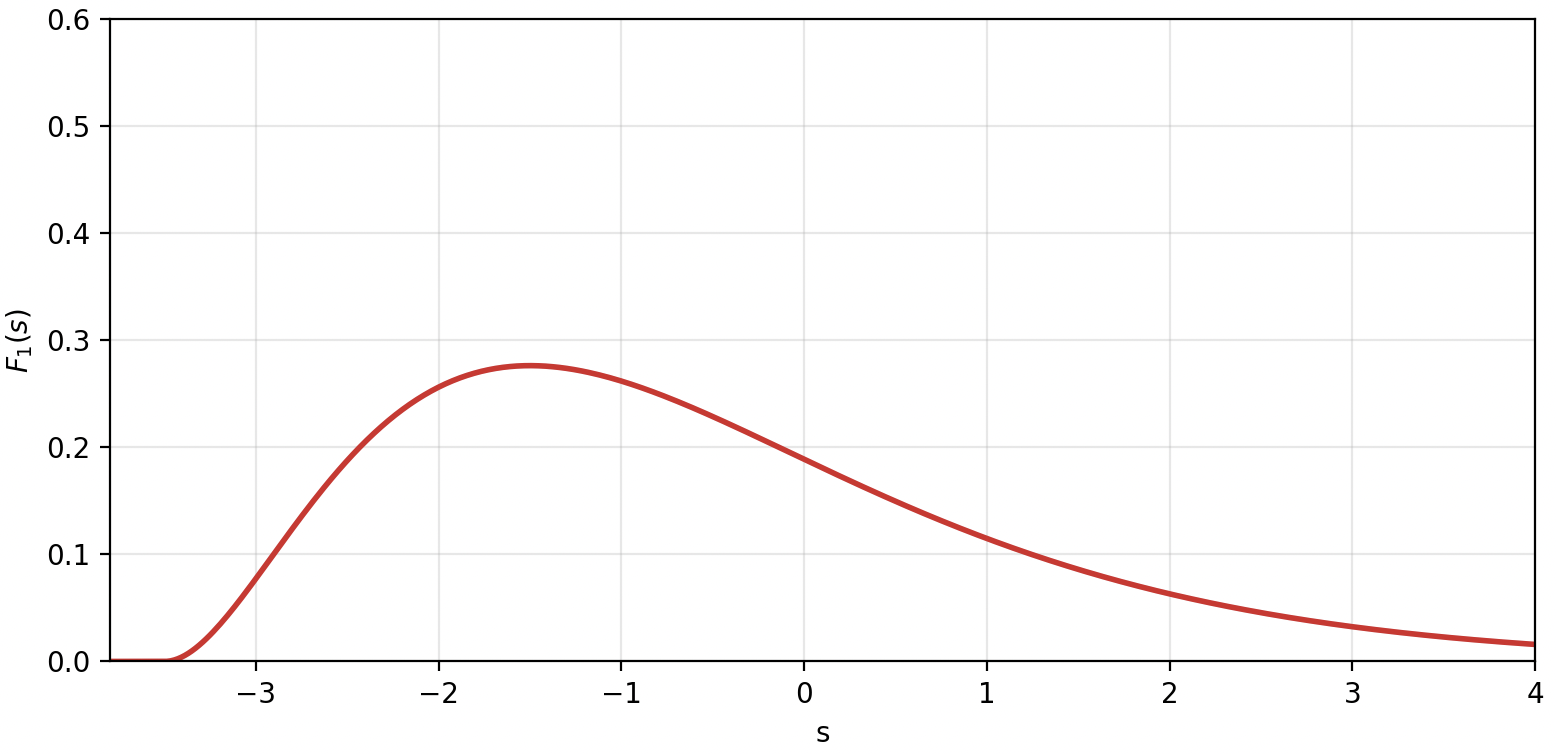}
    \caption[Tracy-Widom distribution]{Tracy-Widom distribution: The variant of the tracy-widom distribution when $\beta=1$ (matrices composed of real numbers). }
    \label{fig:mp_law_background}
\end{figure}

The Tracy--Widom distribution $F_\beta(s)$ describes the limiting fluctuations of the largest eigenvalue $\lambda_{\max}$ of large random matrices after proper centering and scaling, typically as $s = (\lambda_{\max} - \mu_n)/\sigma_n$ where $\mu_n$ is the spectral edge and $\sigma_n \sim n^{-2/3}$, which in case of the MP distribution is \(\lambda_{\max} = \sigma^2(1 + \sqrt{c})^{2} + \alpha n^{-2/3} X\) (where X follows the Tracy–Widom distribution and $\alpha$ is a costant factor). It arises universally for Gaussian matrices with index $\beta = 1, 2, 4$ corresponding to real symmetric (GOE), complex Hermitian (GUE), and quaternionic (GSE) cases. The distribution is defined through the Painlevé~II equation $q''(x) = x\,q(x) + 2\,q(x)^3$ with boundary condition $q(x) \sim \mathrm{Ai}(x)$ as $x \to +\infty$, where $\mathrm{Ai}(x)$ is the Airy function. The cumulative form for $\beta = 2$ is $F_2(s) = \exp[-\!\int_s^{\infty} (x - s)\, q(x)^2\, dx]$, and for $\beta = 1$ it satisfies $F_1(s) = \sqrt{F_2(s)}\, \exp[-\tfrac{1}{2}\!\int_s^{\infty} q(x)\, dx]$. This law captures the universal edge behavior of eigenvalues in random matrix theory and appears in diverse high-dimensional systems.

\subsection{Spiked Covariance Model}

A fundamental extension of the MP model is the spiked covariance model \cite{paul2007asymptotics}, where a low-rank signal is embedded in isotropic noise. The population covariance is: $\boldsymbol{\Sigma} = \sigma^2 \mathbf{I}_p + \sum_{i=1}^k \theta_i u_i u_i^T$  
where $\theta_i$ are spike strengths and $u_i$ are orthonormal signal directions. We use Random Matrix Theory on the sample covariance matrices obtained from the LLMs activations on the datasets to identify the spikes components and the associated signal direction, which is useful for compression by projection and for comparing with the noise baseline.  

\subsection{BBP Phase Transition}
The Baik–Ben Arous–Péché (BBP) phase transition \cite{baik2005phase} states that if the spike strength is below a critical threshold, the corresponding sample eigenvalues remain buried inside the MP bulk. Only when $\theta > \sigma^2 (1+\sqrt{c})$ (equivalently, when population eigenvalues exceed $\lambda_+ = \sigma^2 (1 + \sqrt{c})^2$), do sample eigenvalues separate from the bulk and emerge as outliers: $\lambda_{\text{outlier}} > \lambda_+$.
These detached eigenvalues carry information about the underlying signal, while eigenvalues within the MP bulk represent noise.

The BBP transition emerges directly from the upper edge $\lambda_+$ of the Marchenko-Pastur bulk by considering the asymptotic location of a sample eigenvalue $\hat{\lambda}$ stemming from an isolated population spike $\theta$. This location is given by the mapping $\hat{\lambda}(\theta) \approx \sigma^2 (\theta + \frac{c\theta}{\theta - 1})$ for $\theta > 1$. The phase transition occurs precisely when this isolated eigenvalue detaches from the continuous bulk, i.e., when $\hat{\lambda}(\theta) = \lambda_+$. Substituting $\lambda_+ = \sigma^2 (1 + \sqrt{c})^2$ and solving the equation $\sigma^2(\theta + \frac{c\theta}{\theta - 1}) = \sigma^2(1 + \sqrt{c})^2$ for $\theta$ yields the critical threshold $\theta_{\text{BBP}} = \sigma^2(1 + \sqrt{c})$. Thus, the BBP threshold is derived by finding the minimal population signal strength required to push a sample eigenvalue beyond the theoretical maximum $\lambda_+$ imposed by the noise.

%% file: sections/2_background/4_hallucination.tex
\section{Background: Hallucinations in AI Models}
\label{background:hallucinationsy}

Large-scale neural networks have achieved remarkable success across natural language processing, computer vision, and multimodal reasoning. However, their reliability remains a central challenge. One of the most prominent issues is the phenomenon of \textit{hallucination}, where a model generates fluent but factually incorrect or unsupported content. Hallucinations can emerge even in high-performing systems due to the mismatch between training distributions and real-world usage, or because models prioritize linguistic plausibility over factual grounding. This raises significant concerns in domains such as healthcare, law, and scientific discovery, where incorrect outputs may lead to serious consequences.

Closely related is the problem of \textit{out-of-distribution} (OOD) generalization. Models are trained on large but finite corpora and may encounter inputs at inference that lie outside the training distribution. In these cases, internal representations can become unstable, often leading to either low-quality predictions or hallucinated outputs. This behavior illustrates a fundamental limitation: deep models tend to interpolate well within their training support but extrapolate poorly outside of it.

\subsection{Uncertanty Definition}
Understanding these challenges requires considering the broader notion of \textit{uncertainty} in machine learning. Uncertainty can be decomposed into two main types: \textit{aleatoric} and \textit{epistemic}. Aleatoric uncertainty arises from inherent noise or ambiguity in the data itself, such as ambiguous sentences or low-quality images. Epistemic uncertainty, on the other hand, reflects the model’s lack of knowledge and tends to dominate when the model faces OOD inputs or insufficient training coverage. Formally, if a model produces a predictive distribution $p(y|x,\theta)$ given parameters $\theta$, aleatoric uncertainty is tied to the conditional variability in $y$ for fixed $\theta$, whereas epistemic uncertainty corresponds to variability across different plausible parameter settings.
In practice, distinguishing between these two types is critical for reliability. For example, high aleatoric uncertainty may be unavoidable and can be communicated through probabilistic predictions, while epistemic uncertainty suggests the need for mechanisms such as abstention, retrieval, or further model adaptation. However, modern neural networks often underestimate their uncertainty because maximum-likelihood training encourages overconfident predictions. This miscalibration increases the likelihood that models generate hallucinations, as epistemic uncertainty is often underestimated in their predictive distributions. Such underestimation directly undermines reliability, especially in high-stakes applications.

%% file: sections/2_background/5_compression.tex
\section{Background: Compression in Deep Networks}
\label{background:compression}

The success of deep neural networks is partly due to their overparameterization, which provides strong representational capacity and generalization. However, this leads to high memory usage, inference latency, and energy costs. To mitigate these issues, several compression techniques have been proposed. In this section, we outline four main approaches: knowledge distillation, pruning, quantization, and sparsity. Pruning methods aim to remove redundant or unimportant parameters, typically weights or entire channels, from a trained model. The underlying assumption is that many connections in overparameterized networks contribute little to predictive performance. By eliminating these parameters, the model becomes smaller and faster while maintaining accuracy within acceptable bounds \cite{han2015deep}. Quantization reduces the precision of weights and activations, replacing high-precision floating-point representations with lower-bit formats such as 8-bit integers. This decreases both the memory footprint and the cost of arithmetic operations, enabling deployment on resource-constrained devices. While aggressive quantization can cause accuracy degradation, careful design and post-training calibration mitigate this effect \cite{gholami2021survey}. Sparsity-based approaches enforce or exploit structured or unstructured sparsity in network parameters. Unlike pruning, which is often applied after training, sparse methods may involve training models under sparsity constraints or using specialized architectures. Sparse representations reduce storage and allow faster computation on hardware that supports efficient sparse operations \cite{evci2020rigging}.

\subsection{Knowledge Distillation} Knowledge distillation transfers knowledge from a large, pre-trained ``teacher'' model to a smaller ``student'' model. The student is trained not only on ground-truth labels but also to match the softened output distributions of the teacher, obtained via a temperature-scaled softmax. Given teacher logits $z^{(T)}$ and student logits $z^{(S)}$, the softened probabilities are $p^{(T)}_i = \frac{\exp(z^{(T)}_i / \tau)}{\sum_j \exp(z^{(T)}_j / \tau)}$ and $p^{(S)}_i = \frac{\exp(z^{(S)}_i / \tau)}{\sum_j \exp(z^{(S)}_j / \tau)}$, where $\tau$ is the temperature parameter. The training objective combines the standard cross-entropy with the true labels $y$ and a distillation loss, typically the Kullback–Leibler divergence between $p^{(T)}$ and $p^{(S)}$: $\mathcal{L} = \alpha \, \text{CE}(y, p^{(S)}) + (1-\alpha) \, \tau^2 \, \text{KL}(p^{(T)} \parallel p^{(S)})$. This allows the student to capture not only the hard targets but also the relative similarity structure encoded in the teacher's predictions, leading to compact models with competitive accuracy \cite{hinton2015distilling}.

%% file: sections/2_background/6_spectral.tex
\section{Background: Spectral Methods in Large Models}
\label{background:spectral_methods_in_large_models}

Spectral methods study how information in high-dimensional representations is distributed across directions or frequencies by inspecting the eigenstructure of matrices derived from data or model internals (e.g., activation covariances, Jacobians, or attention maps). In deep learning, these analyses help characterize global geometry (e.g., low-rank structure, anisotropy, and concentration), diagnose training dynamics and generalization, and suggest principled simplifications such as dimensionality reduction or subspace projections \cite{papyan2020powerlaw,martin2021implicit}.

\subsection{Spectral lenses on neural representations} A common approach is to form an activation matrix by stacking hidden features across samples or time and examining its covariance spectrum. The eigenvalues of this covariance describe how variance concentrates along principal directions; rapid decay indicates implicit low-rank structure, while heavy tails suggest richer, multi-scale features \cite{papyan2020powerlaw,martin2021implicit}. Tools like representational similarity based on canonical correlations or centered kernel alignment compute pairwise relationships between layer representations without requiring supervision, enabling comparisons across models, layers, or training stages \cite{raghu2017svcca,kornblith2019cka}. These perspectives complement token-level or pixel-level uncertainty because spectra aggregate information across many units and inputs, producing compact, interpretable descriptors of representation quality.

\textbf{Links to random matrix theory.} When activations are approximately mean-zero and isotropic at a given scale, the empirical covariance spectrum exhibits a bulk consistent with random matrix predictions. The Marchenko–Pastur law describes the asymptotic eigenvalue density of sample covariance matrices with aspect ratio parameter and noise variance, providing a principled baseline for distinguishing noise-dominated directions from signal-bearing outliers; deviations from this baseline, such as outlier eigenvalues or widened gaps, indicate emergent structure \cite{MarchenkoPastur1967}. For symmetric weight or Hessian-like operators with independent fluctuations, the Wigner semicircle distribution offers a corresponding null model \cite{wigner1958}. In practice, deep networks rarely behave like pure noise, but these laws give calibrated reference points: directions with eigenvalues near the bulk edge are more likely to be correlational or redundant, while detached eigenvalues often track semantic or task-relevant factors.

\subsection*{Frequency and complexity biases}Spectral viewpoints also illuminate which functions neural networks learn most readily. In overparameterized regimes, many models exhibit a ``spectral bias,'' fitting low-frequency or smoother components first before higher-frequency detail as training progresses. This bias can be observed in the singular value structure of learned features and in Fourier analyses of fitted functions, with implications for generalization and robustness \cite{rahaman2019spectral}. Spectral diagnostics thus connect training dynamics to hypothesis complexity and can guide early stopping or curriculum choices.

\subsection{Applications to vision and language.} In computer vision, spectra of convolutional features often display heavy-tailed behavior consistent with implicit regularization, and class-conditional structure can concentrate along a small number of directions, reflecting semantic disentanglement in later layers \cite{papyan2020powerlaw,martin2021implicit}. In language models, representation similarity analyses quantify how linguistic abstractions emerge across depth and how attention blocks reconfigure information, while covariance spectra reveal anisotropy and low-rank tendencies in token embeddings and intermediate states \cite{raghu2017svcca,kornblith2019cka}. Across both modalities, spectral characterization offers a unifying language to reason about layer roles, information flow, and the trade-off between expressivity and compression.

\begin{figure}[h]
    \centering
    \includegraphics[width=0.75\linewidth]{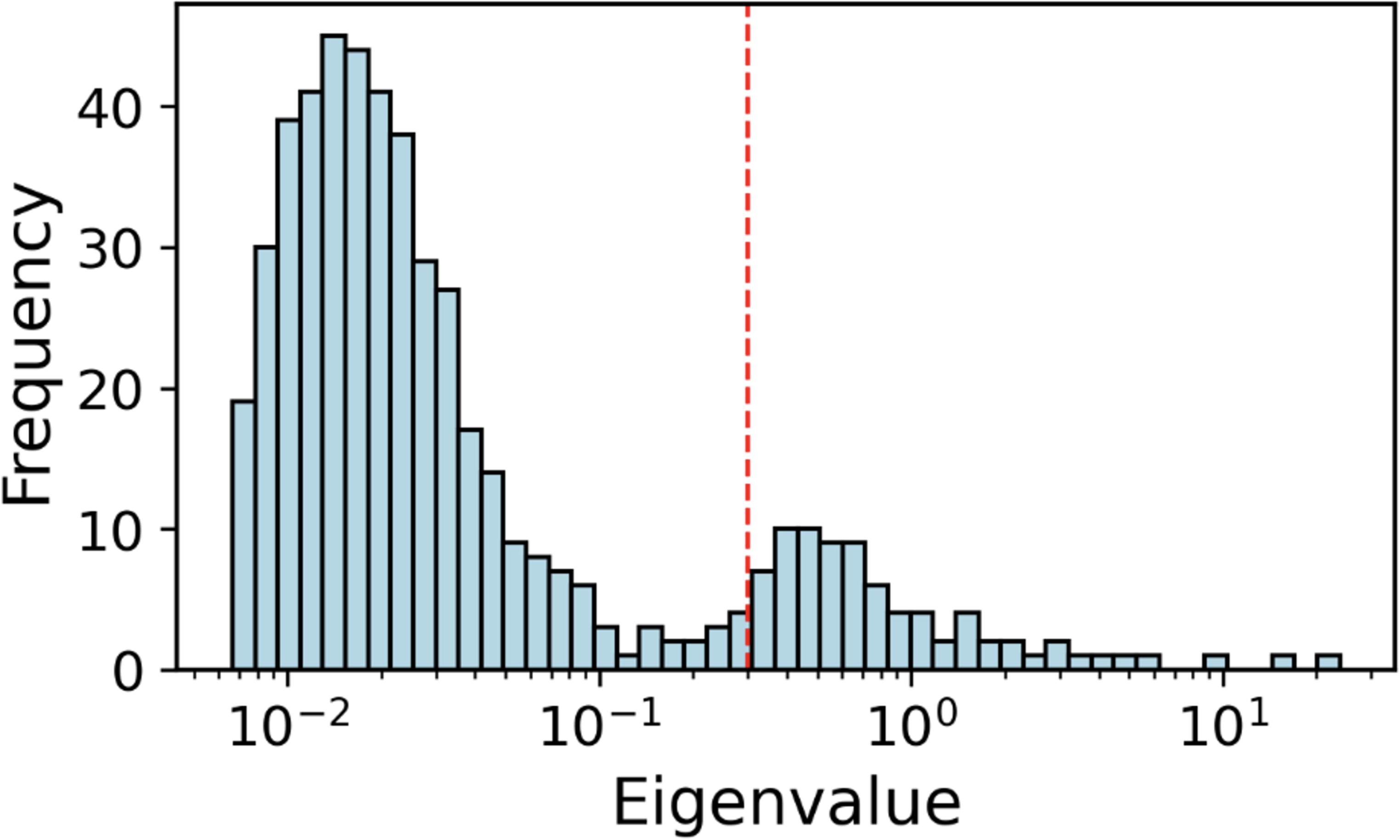}
    \caption[Empirical eigenvalue distribution]{Empirical eigenvalue distribution: spectrum of hidden layer activations in a deep network. 
  The histogram shows a bulk of small eigenvalues consistent with noise-like directions, while a few outliers (to the right of the red line) indicate informative, structured components. 
  Such separation between bulk and outliers is the basis of spectral analyses in deep learning.}
    \label{fig:mp_law_background}
\end{figure}

\textbf{Implications for practice.} Because spectral summaries are global and compact, they are useful for monitoring training health, identifying distributional shifts, and motivating architecture or optimization choices. For example, a pronounced low-rank structure motivates low-rank adapters and subspace updates, where model changes are confined to a small number of directions while preserving overall behavior \cite{hu2022lora}. More broadly, comparing empirical spectra to random matrix baselines provides quantitative signals for when representations become structured, when they remain noise-like, and how complexity evolves with data and compute.

%% file: sections/2_background/7_bridge.tex
\section{Background: Efficiency and Reliability in Large AI Models}
\label{background:efficiency_reliability}

The rapid scaling of artificial intelligence (AI) systems has brought remarkable advances in natural language processing, computer vision, and multimodal learning. At the same time, this growth has exposed two central and often competing requirements: \textit{efficiency}, the ability to deploy and operate models within practical computational budgets, and \textit{reliability}, the assurance that predictions are robust, trustworthy, and aligned with human expectations. While these dimensions are frequently studied in isolation, they are deeply interconnected. Both can be analyzed through the lens of \textit{spectral geometry}, which provides a principled framework for understanding the dynamics of high-dimensional neural representations.

\subsection*{Efficiency in Large-Scale AI Models}
Efficiency concerns the computational and resource costs associated with training and deploying large models. With the rise of billion-parameter networks, challenges include inference latency, memory footprint, and energy consumption \cite{strubell2019energy}. Scaling laws suggest that performance often grows predictably with model size and dataset scale \cite{kaplan2020scaling}, but this trend comes at significant financial and environmental cost. Techniques to address efficiency aim to reduce redundancy in representations, compress parameters, or exploit low-rank structures without compromising accuracy \cite{han2016deep}. Fundamentally, efficiency measures whether a model can operate in real-world conditions, from data centers to edge devices, while preserving its utility.

\subsection*{Reliability in Large-Scale AI Models}
Reliability addresses the trustworthiness of model predictions. Large neural networks are known to hallucinate, produce spurious correlations, and degrade sharply on inputs outside the training distribution \cite{hendrycks2017baseline,ovadia2019can}. Failures can arise from sensitivity to noise, adversarial perturbations, or distributional shifts. Traditional confidence scores, such as softmax probabilities, are poorly calibrated and fail to capture deeper instabilities in learned representations \cite{guo2017calibration}. Reliability, therefore, encompasses robustness, calibration, interpretability, and alignment with ground-truth semantics. It is increasingly critical as AI models are deployed in high-stakes domains such as healthcare, finance, and law.

\subsection{Spectral Geometry as a Bridge}
Although efficiency and reliability may seem distinct, both are tied to the geometry of neural representations. High-dimensional activations often concentrate around low-rank subspaces, and their covariance spectra reveal whether learned features are structured or dominated by noise \cite{martin2021implicit}. Efficient models seek to eliminate redundant directions, while reliable models must ensure that remaining directions capture stable and causal signals. Spectral geometry provides the mathematical lens for unifying these perspectives: eigenvalue distributions and spectral entropy measure redundancy, while spectral gaps and deviations from random matrix baselines indicate stability and robustness. By analyzing activations through this shared framework, it becomes possible to design models that are both compact and trustworthy.

%% file: sections/3_related/0_index.tex
\chapter{Related Work}
\label{chap:related}

\input{sections/3_related/1_hallucination}

\input{sections/3_related/2_ood}

\input{sections/3_related/3_compression}

\input{sections/3_related/4_rmt}

\input{sections/3_related/5_multimodal}

%% file: sections/3_related/1_hallucination.tex
\section{Related: Hallucination Detection} \label{related:hallucination_detection}
Large language models exhibit ``hallucinations'', namely fluent statements that are unfaithful to sources or unverifiable by background knowledge, this jeopardizes safety in high–stakes domains, erodes user trust through confidently wrong explanations, and can propagate misinformation at scale, empirical evidence suggests that hallucinations persist even as models scale and fine–tune, because next–token prediction does not explicitly penalize confidently fabricated content and because training data contain human misconceptions \cite{Ji2023HallucinationSurvey,Lin2021TruthfulQA}. Recent research therefore focuses on \emph{detection} rather than elimination, aiming to flag risky generations with minimal latency and without full model access, the literature can be organized into black–box, grey–box, white–box, and spectral families that trade off access assumptions, cost, and generality.

\subsection{Black-box Methods}
Black–box detectors only observe input, output, and optionally multiple samples, without logits or hidden states, a representative line leverages self–consistency across stochastic re–generations, SelfCheckGPT queries the model multiple times and measures agreement with sentence–level QA, NLI, and semantic similarity probes to infer whether a claim is supported, reporting strong zero–resource performance across QA and summarization \cite{manakul2023selfcheckgpt}. Complementary work turns disagreement into a calibrated probability of hallucination, a cost–effective multi–scoring framework aggregates diverse uncertainty signals, then performs conditional calibration to obtain risk–aware thresholds that are competitive with heavier pipelines while reducing compute \cite{Valentin2024CostEffective}. Another strand uses chains of natural language inference, CoNLI decomposes a response into atomic claims and checks entailment against sources, then post–edits ungrounded spans, providing a plug–and–play pipeline without task–specific fine–tuning \cite{lei2023conli}. Black–box methods are appealing for proprietary models and production since they require no internals and are easy to retrofit, however they can be expensive due to multi–sample decoding, may fail on \emph{self–consistent} errors where the model repeats the same falsehood across samples, and can be sensitive to prompt format and decoding temperature \cite{Ji2023HallucinationSurvey,Lin2021TruthfulQA}.

\subsection{Grey-box}
Grey–box detectors exploit limited internals such as token log–probabilities while avoiding access to hidden states, DetectGPT observes that model–generated text tends to occupy negative curvature regions of the log–likelihood landscape, it perturbs the text and measures average change in log–probability to separate machine from human text \cite{mitchell2023detectgpt}, Fast–DetectGPT replaces masked perturbations with conditional probability curvature, substantially accelerating detection while preserving accuracy \cite{bao2023fastdetectgpt}. Because many commercial APIs expose token log–probs only intermittently, recent work predicts fuller distributions from partial observations to extend white–box–style criteria into proprietary settings, Glimpse learns to reconstruct probability information and thereby enables rank, entropy, and curvature–based detectors on API–only models \cite{bao2025glimpse}. Grey–box approaches are efficient when logits are available and avoid re–generating many samples, yet they may degrade if APIs obscure probabilities, curvature signals can correlate with distribution shift unrelated to factuality, and they do not directly capture conflicts with external evidence \cite{Valentin2024CostEffective}.

\subsection{White-box Methods}
White–box detectors use internal activations, attention maps, or intermediate logits, often enabling real–time alarms, MIND trains an unsupervised detector on internal states collected during inference, pairing automatic data labeling with a lightweight classifier for on–the–fly detection, and reports competitive latency–accuracy trade–offs \cite{Su2024MIND}. INSIDE argues that dense internal states retain semantic information lost in surface tokens, introduces an \emph{EigenScore} derived from the eigenvalues of the covariance of sentence embeddings to quantify response self–consistency, and proposes test–time feature clipping to curb overconfident hallucinations \cite{Chen2024INSIDE}. Faithfulness scoring from attention patterns estimates whether the tokens that should support an answer receive proportionate attention, improving detection of unsupported rationales and explanations \cite{sriramanan2024attentionscore}. Finally, learning from unlabeled outputs in the wild, HaloScope estimates membership in truthful versus untruthful subpopulations among raw LLM generations, enabling a practical truthfulness classifier trained without manual labels and showing strong generalization across datasets \cite{Du2024HaloScope}. White–box methods can be accurate and low–latency because they align with model computation, nevertheless portability to closed–source APIs is limited, internal probes risk picking up spurious correlates, and detectors trained on one family of models may require recalibration for another \cite{Ji2023HallucinationSurvey}.

\subsection{Spectral Methods}
Spectral detectors focus on eigenvalue structure rather than token probabilities or raw attention weights, the central idea is that under benign conditions hidden representations behave approximately like high–dimensional noise with a bulk eigenvalue distribution predicted by random matrix theory, for instance the Marchenko–Pastur law for sample covariance, while emergent low–rank correlations under failure modes produce outlier eigenvalues and widened spectral gaps, deviations that can be monitored during generation \cite{MarchenkoPastur1967,BBP2005}. The implementation vary with the object being analyzed, attention–based spectral methods interpret attention as a graph and compute eigenvalues of the Laplacian, the top–$k$ Laplacian eigenvalues or their gaps feed compact probes that show state–of–the–art accuracy among attention–centric detectors and preserve interpretability through eigengap analysis \cite{Binkowski2025LapEigvals}. White–box consistency metrics based on eigenvalues, for example the EigenScore, use the spectrum of internal sentence–embedding covariances to assess diversity versus concentration in the generated content, with higher dispersion or bulk–like spectra indicating weaker semantic grounding \cite{Chen2024INSIDE}. Spectral methods are attractive because eigenvalue statistics compress high–dimensional geometry into a few interpretable numbers, they are efficient to compute with truncated SVD and admit principled calibration against theoretical baselines, at the same time they require careful control for sequence length, windowing, and layer choice to avoid conflating harmless stylistic shifts with true hallucinations, and theoretical references such as Marchenko–Pastur and the Baik–Ben Arous–P\'ech\'e phase transition provide guidance for separating bulk from spikes but must be adapted to non–ideal neural settings \cite{MarchenkoPastur1967,BBP2005}.

%% file: sections/3_related/2_ood.tex
\section{Related: Out-of-Distribution Detection} \label{related:ood_detection}
Modern deep networks, especially large language and vision–language models, often operate under the implicit assumption that test data follow the same distribution as training data. When this assumption is violated, model predictions can become unreliable or even nonsensical, a condition known as the out-of-distribution (OOD) problem \cite{Yang2021GeneralizedOOD}. OOD detection aims to identify when an input lies outside the support of the training distribution so that the system can abstain, trigger fallback mechanisms, or re-calibrate uncertainty estimates. In LLMs and multimodal architectures, OOD behavior manifests as factual hallucinations, overconfident predictions on unseen topics, or unstable performance under domain shifts. Since full retraining is costly, OOD detection modules are a practical route toward safer deployment. This section reviews three major classes of methods: score-based, representation-based, and spectral/statistical approaches, summarizing their core principles, effectiveness, and limitations.

\subsection{Score-based Methods}
Score-based OOD detectors rely on output statistics such as confidence, logit margins, or energy, treating extreme values as indicators of abnormality. The most classical baseline, maximum softmax probability (MSP), measures confidence as $\max_i p(y_i|x)$, where low values suggest OOD samples \cite{hendrycks2017baseline}. Although simple, MSP often underestimates uncertainty because deep networks are overconfident even on random noise. Temperature scaling and input perturbations were later introduced in ODIN, which perturbs the input slightly in the direction of the gradient and rescales logits by a temperature parameter to separate in- and out-distribution samples more effectively \cite{Liang2018ODIN}. The Mahalanobis method computes layerwise Gaussian scores by modeling class-conditional feature distributions and measuring Mahalanobis distance to the closest class mean \cite{Lee2018Mahalanobis}. These approaches are efficient and can be applied post-hoc without retraining. However, their calibration is architecture- and dataset-dependent, often failing when the model's softmax layer saturates.

Energy-based scores generalize confidence by defining an “energy” $E(x) = -T \log \sum_i e^{z_i/T}$ from logits $z_i$, where in-distribution samples concentrate near low energy \cite{Liu2020EnergyOOD}. Energy-based OOD detection achieves strong results in computer vision and has been extended to transformers and LLMs \cite{Jiang2023EnergyLLM}, offering smoother uncertainty surfaces than probability-based scores. More recently, normalized cosine similarity and logit margin distributions have been proposed for sequence models \cite{Xiao2024SeqOOD}. Score-based methods are attractive for their simplicity and compatibility with black-box systems but require careful threshold tuning and can conflate epistemic and aleatoric uncertainty, limiting interpretability.

\subsection{Representation-based Methods}
Representation-based detectors leverage the geometry of internal embeddings or subspaces, motivated by the observation that in-distribution data form compact manifolds in hidden space while OOD inputs deviate from them. Early work modeled feature embeddings with Gaussian mixtures or k-nearest neighbors, classifying high-distance samples as OOD \cite{Sun2022ReAct}. Deep kNN (DkNN) retrieves the labels of nearest training features at multiple layers to estimate conformity and provides layerwise reliability scores \cite{Papernot2018DkNN}. Subsequent work developed centroid-based methods such as kNN cosine similarity, Center Loss embeddings, and Local Outlier Factor variants \cite{Tack2020CSI}.

Advanced subspace approaches analyze the rank or span of hidden representations. SVCCA and CKA metrics have been used to compare activation subspaces across layers and models, showing that OOD inputs distort canonical correlations and occupy low-similarity directions \cite{raghu2017svcca,kornblith2019cka}. Representation self-similarity metrics like Mahalanobis with shrinkage or spectral norm regularization further refine detection by discounting correlated noise. Recent studies in transformer-based encoders use attention entropy and hidden-state variance as signals of uncertainty, achieving strong transfer to text and multimodal data \cite{Yilmaz2022OODText}. Representation-based approaches provide interpretability by localizing anomalies to specific layers or embeddings, yet they require storing or approximating training features, which may be memory-intensive for large-scale models, and they may degrade under feature collapse caused by overfitting or adversarial fine-tuning.

\subsection{Spectral and Statistical Methods}
Spectral and statistical approaches analyze the eigenvalue spectrum or higher-order statistics of activation covariances, connecting OOD detection to Random Matrix Theory (RMT). The empirical covariance $\Sigma = \frac{1}{n} XX^\top$ of hidden activations $X \in \mathbb{R}^{d \times n}$ is expected to follow a bulk spectrum approximated by the Marchenko–Pastur distribution under in-distribution conditions, with deviations indicating structured or anomalous behavior. When an input is OOD, spurious correlations can induce outlier eigenvalues or inflate the spectral tail beyond the bulk limit. This phenomenon was observed in spectral OOD detectors such as RankFeat, which measures the effective rank of covariance matrices as a proxy for representation collapse \cite{Sun2022RankFeat}. SpectralGap expands on this by computing the difference between consecutive eigenvalues (the eigengap) and thresholding on sharp transitions that suggest emergent low-dimensional structure \cite{Park2024SpectralGap}. SNoJoE, an efficient singular value–based detector, evaluates joint energy across layers by tracking the spectral norm and nuclear norm evolution of activations, achieving state-of-the-art detection in both CNNs and transformers \cite{Ahmed2024SNoJoE}. 

These methods are attractive because eigenvalue spectra provide an unsupervised and architecture-agnostic summary of distributional structure, allowing theoretical calibration through RMT. Moreover, they align naturally with the spectral diagnostics used for reliability and hallucination detection, offering a principled link between OOD and uncertainty quantification. Nonetheless, spectral detectors can be sensitive to batch size, normalization, and numerical precision when computing eigenvalues at scale. Extensions using randomized eigensolvers and moving-window covariance estimates partially mitigate the cost, enabling online OOD tracking in large models \cite{Martin2021RMTImplicit,Donos2020SpectralShift}. Overall, spectral and statistical approaches form a rapidly growing subfield that unifies deep learning reliability analysis with classical multivariate statistics.

%% file: sections/3_related/3_compression.tex
\section{Related: Model Compression} \label{related:model_compression}
Deploying modern deep networks in latency and energy constrained settings motivates principled compression, the goal is to reduce parameters, memory footprint, and inference cost while preserving accuracy and calibration. Four families dominate the literature, pruning, quantization, low rank factorization, and knowledge distillation. Each family spans classic CNN accelerators and transformer specific variants, and each introduces distinct trade offs between software and hardware efficiency, portability, and accuracy retention.

\subsection{Pruning}
Pruning removes parameters judged unnecessary for task performance. Unstructured pruning zeros individual weights according to saliency, magnitude pruning introduced in the deep compression pipeline shows that iterative prune, retrain and Huffman coding achieves large sparsity with minimal accuracy loss in CNNs \cite{Han2016DeepCompression}, while the lottery ticket hypothesis argues that dense networks contain trainable sparse subnetworks that reach competitive accuracy when reset and fine tuned, revealing overparameterization and the importance of initialization \cite{Frankle2019LotteryTicket}. Gradient and Hessian based criteria estimate sensitivity more precisely, Taylor expansion pruning and WoodFisher approximate the loss increase from removing a weight using first or second order information and can outperform magnitude baselines, especially at high sparsity \cite{Molchanov2017PruningTaylor,Singh2020WoodFisher}. In transformers, movement pruning encourages weights to move toward zero during fine tuning via a dynamic L0 style regularizer, yielding structured sparsity patterns that transfer across tasks \cite{Sanh2020MovementPruning}.

Structured pruning removes whole channels, heads, and blocks to obtain hardware friendly speedups. For CNNs, channel level sparsity via L1 penalty on batch norm scales, known as network slimming, reliably shrinks models with real wall clock gains \cite{Liu2017NetworkSlimming}. Automated search methods, AMC and MetaPruning, couple reinforcement learning or meta networks with a pruning controller to allocate sparsity budgets per layer under latency constraints \cite{He2018AMC,Liu2019MetaPruning}. In transformers, studies show many attention heads are redundant, removing heads and even entire feed forward sublayers can preserve accuracy with careful retraining \cite{Michel2019Heads}. Structured approaches align well with dense libraries and accelerators, however they may leave accuracy on the table relative to fine grained sparsity, and benefits depend on compiler and kernel support for the targeted structure.

\subsection{Quantization}
Quantization maps floating point tensors to low precision integers to reduce memory bandwidth and arithmetic cost. Post training quantization calibrates scales after training without access to labels, uniform 8 bit affine quantization with per channel weights and per tensor activations is a robust baseline across CNNs and transformers \cite{Jacob2018Q8}. Large language models exhibit activation outliers that break naive post training quantization, SmoothQuant shifts range from activations into weights during calibration to suppress outliers, enabling 8 bit activations and 8 bit or 4 bit weights at minimal perplexity cost \cite{Xiao2023SmoothQuant}. GPTQ proposes blockwise second order weight quantization that solves a local least squares objective using approximate Hessian information, delivering strong 3–4 bit results on LLMs with a single calibration pass \cite{Frantar2022GPTQ}. AWQ further observes that a small set of salient channels dominate error, protecting them during quantization improves 4 bit accuracy on a wide range of decoder only models \cite{Lin2023AWQ}. ZeroQuant and ZeroQuant v2 push post training quantization toward 8, 6, and 4 bit on both weights and activations with tensor wise and group wise calibration that scales to billion parameter models \cite{Yao2022ZeroQuant,Yao2022ZeroQuantV2}. 

Quantization aware training inserts fake quantization nodes during fine tuning so the network learns to be robust to quantization noise. Learned step size quantization treats scale as a learnable parameter and achieves state of the art low bit accuracy in CNNs \cite{Esser2019LSQ}, while Q BERT adapts group wise quantization and Hessian guided mixed precision to transformers, reporting strong gains at 4 bit and below \cite{Shen2020QBERT}. Post training methods are attractive for cost and simplicity, yet may require sizable calibration sets for LLMs, whereas QAT yields better extremes at the expense of additional fine tuning and careful optimizer choices. Realized speedups depend on integer kernels, activation quantization, and operator coverage along the whole graph.

\subsection{Low rank factorization}
Low rank factorization exploits linear redundancy by decomposing weight tensors into products of smaller tensors. For fully connected layers, truncated SVD yields two narrow matrices that approximate the original, reducing multiplies without sparsity \cite{Denton2014ExploitingSVD}. For convolutions, spatial separability and tensor decompositions apply, CP and Tucker decompositions factorize 4D kernels into sequences of smaller convolutions, maintaining accuracy after brief fine tuning \cite{Lebedev2014CP,Kim2016Tucker}. In transformers, parameter sharing and projection tying reduce redundancy, ALBERT shares parameters across layers and factorizes the embedding matrix, substantially shrinking model size with modest accuracy changes \cite{Lan2019ALBERT}. 

Adapter style low rank updates enable parameter efficient fine tuning rather than inference time compression, yet the same principle illuminates low intrinsic dimensionality in language models. LoRA injects trainable low rank matrices into attention and MLP projections while freezing the base network, matching or surpassing full fine tuning with a small fraction of trainable parameters \cite{hu2022lora}, extensions such as DoRA decouple direction and magnitude to stabilize training and further reduce rank \cite{Liu2024DoRA}, and analyses of intrinsic dimension argue that many NLP tasks lie in very low dimensional subspaces relative to model size \cite{Aghajanyan2020IntrinsicDim}. When used for compression rather than adaptation, factorization replaces large dense projections with products of thin matrices in the deployed model, offering dense kernel friendliness and predictable latency, at the cost of designing ranks per layer and possible accuracy drops if ranks are set too aggressively.

\subsection{Knowledge distillation}
Knowledge distillation transfers behavior from a high capacity teacher to a smaller student using soft targets and auxiliary hints. The classical formulation minimizes a convex combination of task loss and KL divergence between teacher and student output distributions at a temperature, improving generalization of the student \cite{Hinton2015KD}. FitNets introduced intermediate feature hints to guide deeper students \cite{Romero2015FitNets}, attention transfer encourages students to match teacher attention maps \cite{Zagoruyko2017AT}, and contrastive representation distillation aligns teacher and student embeddings by maximizing mutual information at the instance level \cite{Tian2020CRD}. 

In NLP, distillation underpins many compact transformers. DistilBERT compresses BERT by matching soft logits, intermediate states, and attention distributions during pre training, halving parameters and improving speed with small accuracy loss \cite{Sanh2019DistilBERT}. Patient knowledge distillation matches layer wise representations with patience, i.e., allowing the student to align to multiple teacher layers over time, improving stability and performance \cite{Sun2019PKD}. BERT of Theseus progressively replaces teacher components with student modules during fine tuning, interpreting the process as a Markov chain of network components and delivering competitive small models \cite{Xu2020Theseus}. MiniLM distills deep self attention by matching value relation matrices and query key self attention distributions, yielding strong students with tiny footprints \cite{Wang2020MiniLM}. Beyond fixed students, progressive shrinking in Once for All training learns a super network that can instantiate sub networks of different widths and depths, combining distillation with neural architecture adaptation to hit diverse latency targets without retraining from scratch \cite{Cai2020OFA}. 

In vision, compact students like MobileNets and EfficientNets commonly incorporate KD during training, and recent works explore multi teacher ensembles, self distillation without an external teacher, and task specific distillation for detection and segmentation. Distillation is flexible, it composes with pruning, quantization, and factorization by recovering accuracy after structural changes, and it can target calibration and robustness objectives beyond top 1 accuracy. Its costs include teacher inference during training, potential mismatch across domains, and sensitivity to loss weighting, temperature, and layer mapping choices.

%% file: sections/3_related/4_rmt.tex
\section{Related: Random Matrix Theory in Deep Learning} \label{related:rmt_in_deep_learning}
Random Matrix Theory (RMT) provides a statistical–mechanical lens to analyze deep neural networks, treating weight and activation matrices as high-dimensional random systems. By studying the eigenvalue spectra of these matrices, RMT reveals universal behaviors that shed light on learning dynamics, generalization, and robustness. Unlike heuristic diagnostics, spectral geometry provides analytical tools to quantify phase transitions, implicit regularization, and signal–noise separation in modern networks. This section summarizes key developments applying RMT to deep learning and related areas of reliability and robustness.

\subsection{Phase Transitions}
A central insight from RMT is that eigenvalue distributions of random covariance matrices follow the Marchenko–Pastur (MP) law, which describes the limiting density of eigenvalues for a matrix $X X^\top / n$ when both $n$ and the feature dimension grow large \cite{MarchenkoPastur1967}. In deep networks, deviations from this theoretical bulk indicate structured correlations in learned representations. When training begins, the spectrum typically resembles the MP bulk; as learning progresses, a few eigenvalues detach from the bulk, forming outliers associated with meaningful features or “spikes.” This transition is governed by the Baik–Ben Arous–Péché (BBP) threshold \cite{BBP2005}, which determines when a signal eigenvalue separates from random noise.

Empirical studies demonstrate that layer weight spectra in trained networks display heavy-tailed distributions following power laws rather than pure MP shapes \cite{Martin2021RMTImplicit,Martin2020HeavyTailed}. These heavy tails reflect correlations induced by gradient descent and act as fingerprints of generalization. The phase-transition view thus links RMT to implicit bias: well-generalizing models operate near criticality, balancing order (signal) and chaos (noise). Conversely, overtrained or overregularized models exhibit either degenerate (collapsed) spectra or overly flat ones, signaling underfitting or overparameterization.

\subsection{Implicit Regularization}
RMT also explains the implicit regularization mechanisms that emerge in large-scale optimization. Gradient descent, batch normalization, and dropout collectively push weight matrices toward critical spectral regimes, shaping their eigenvalue decay \cite{Martin2021RMTImplicit}. This self-organized criticality aligns with the notion of \emph{spectral bias}, where networks learn low-frequency, smooth components first. In the spectral domain, weight and gradient covariance spectra often show $1/\lambda^\alpha$ power-law tails with $\alpha$ between 2 and 5, characterizing an intermediate regime between random noise and strong correlations.

Such spectral regularization correlates with model robustness and generalization: networks with heavy-tailed spectra resist overfitting and exhibit better OOD performance \cite{Pennington2018Resurrecting,Martin2023Tikhonov}. Moreover, the Hessian spectra of well-trained models typically have a small number of large eigenvalues corresponding to informative directions and a bulk near zero corresponding to flat minima. This separation implies that stochastic optimization implicitly enforces a spectral cutoff, analogous to Tikhonov regularization in inverse problems. In this view, spectral analysis provides both a diagnostic and a theoretical justification for why simple optimization procedures produce well-behaved models without explicit constraints.

\subsection{Subspace Analysis}
Beyond bulk statistics, RMT provides tools for subspace analysis of learned representations. In deep networks, activations at each layer can be viewed as samples from a high-dimensional manifold embedded in feature space. The covariance spectrum captures how variance is distributed across latent directions. Signal–noise decomposition based on spectral truncation isolates meaningful subspaces from isotropic noise, yielding insights into feature redundancy and causal directions. Studies show that informative subspaces correspond to outlier eigenvectors, while the noise subspace follows MP-like statistics \cite{Sagun2017Empirical,Martin2021RMTImplicit}. Monitoring spectral entropy or eigengaps across layers reveals how information condenses as depth increases, forming hierarchical subspaces of decreasing intrinsic dimension.

Recent work extends subspace spectral analysis to multimodal models. In vision–language transformers, joint embeddings exhibit coupled spectral dynamics alignment across modalities manifests as correlated eigenvalue shifts between text and vision subspaces. Spectral filtering, where activations are projected onto dominant eigenspaces, has been shown to enhance robustness and interpretability. In particular, \emph{EigenShield} introduces causal subspace filtering via RMT principles to defend against adversarial perturbations in vision–language models, identifying and suppressing anomalous eigenmodes while preserving semantic structure \cite{darabi2025eigenshield}. This line of work connects RMT-based subspace analysis to adversarial security, suggesting that spectral geometry can serve both analytical and defensive roles.

%% file: sections/3_related/5_multimodal.tex
\section{Related: Multimodal Models and Hallucination} \label{related:multimodal_models}
The integration of visual, textual, and sometimes auditory modalities has enabled a new generation of foundation models capable of perception and reasoning across heterogeneous data. Vision–language models (VLMs) such as CLIP, BLIP, Flamingo, LLaVA, and GPT-4V have demonstrated impressive zero-shot capabilities on diverse multimodal tasks including captioning, VQA, and visual reasoning \cite{Radford2021CLIP,Li2022BLIP,Alayrac2022Flamingo,Liu2023LLaVA,OpenAI2023GPT4V}. However, the same scaling that endows such flexibility also amplifies problems of reliability. Multimodal hallucination, the generation of textual or visual content inconsistent with the actual input image or video, emerges as a major limitation \cite{Li2023SurveyMultimodalHallucination}. These hallucinations can manifest as fabricated objects in image captions, incorrect visual attributes, or semantically inconsistent reasoning steps, threatening the deployment of VLMs in sensitive contexts such as medical imaging, autonomous driving, or assistive technology.

\subsection{Nature of Multimodal Hallucinations}
Multimodal hallucination differs fundamentally from text-only hallucination because it stems from misalignment between modalities rather than internal linguistic overconfidence alone. Empirical studies classify hallucinations into three categories: \emph{object hallucination}, where models invent entities absent from the visual scene; \emph{attribute hallucination}, where they misdescribe color, shape, or position; and \emph{contextual hallucination}, where textual reasoning contradicts visual evidence \cite{Rohrbach2018ObjectHallucination,Li2023SurveyMultimodalHallucination}. The underlying causes include dataset biases (co-occurrence correlations that teach the model spurious associations), imbalance between vision and language encoders, and autoregressive decoding that privileges linguistic priors over perceptual grounding.

Early VLMs such as OSCAR and VinVL relied heavily on object detection features, which constrained hallucination but limited generality \cite{Li2020OSCAR,Zhang2021VinVL}. End-to-end pretrained architectures like CLIP and BLIP2 improved generalization but exhibited increased hallucination due to weaker explicit grounding. Recent works have shown that model scale and instruction-tuning can paradoxically increase hallucination frequency even while improving benchmark accuracy, as seen in GPT-4V and other multimodal LLMs \cite{Liu2023LLaVA,OpenAI2023GPT4V}. This tension highlights a key challenge: multimodal grounding requires both accurate fusion and balanced modality dominance.

\subsection{Detection and Mitigation Strategies}
Research into multimodal hallucination detection adapts both linguistic and perceptual uncertainty measures. Black-box metrics evaluate semantic alignment between generated captions and ground-truth regions using object detectors or CLIP-based similarity scores. The VisualHallucination benchmark proposed quantitative metrics for hallucinated object frequency and visual faithfulness \cite{Li2023SurveyMultimodalHallucination}. Gray-box approaches leverage attention entropy or cross-modal similarity matrices computed from intermediate layers to detect when text tokens attend to irrelevant visual regions \cite{Yin2023MHalDet}. White-box or spectral detectors extend this by analyzing hidden covariance spectra of multimodal fusion layers: deviations from expected eigenvalue distributions or sharp spectral gaps correspond to loss of alignment, providing unsupervised hallucination signals consistent with Random Matrix Theory-based diagnostics.

Mitigation techniques span architectural, training, and decoding interventions. Architecturally, contrastive pretraining with balanced image–text pairs (e.g., CLIP, ALIGN) reduces dataset-induced biases \cite{Radford2021CLIP,Jia2021ALIGN}. Training-level solutions such as grounding instruction tuning explicitly penalize ungrounded generations by augmenting prompts with object tags or region features \cite{Liu2023LLaVA,Li2023InstructBLIP}. Reinforcement learning with human feedback (RLHF) has also been adapted to multimodal settings, optimizing for human-judged visual faithfulness instead of text coherence \cite{OpenAI2023GPT4V}. Decoding-level methods, including constrained beam search or re-ranking by cross-modal consistency, filter out hallucinated outputs post hoc \cite{Rohrbach2018ObjectHallucination}.

\subsection{Spectral and Causal Perspectives}
From a spectral viewpoint, multimodal fusion layers can be viewed as coupling two high-dimensional subspaces, visual and textual representations, whose joint covariance structure governs alignment. When properly grounded, the joint covariance spectrum exhibits well-matched eigenmodes with smooth decay; hallucinations correlate with the emergence of spurious outlier eigenvalues, representing latent directions dominated by linguistic noise or irrelevant visual activations. This aligns with recent analyses of eigenvalue gaps and cross-modal covariance geometry in transformer blocks \cite{Martin2021RMTImplicit,darabi2025eigenshield}. Spectral regularization and subspace filtering, as explored in the context of adversarial robustness, can thus mitigate multimodal hallucinations by projecting activations onto stable causal eigenspaces.

Beyond diagnosis, such RMT-informed methods offer interpretability: by tracking which eigenvectors carry cross-modal signal versus noise, one can localize the onset of hallucination and analyze its modality-specific source. This spectral–causal perspective suggests that multimodal hallucination is not an isolated failure mode but an emergent property of unbalanced eigenspectra, where linguistic variance overwhelms visual evidence. As multimodal models continue to scale, bridging RMT analysis, causal reasoning, and alignment training will be crucial to achieving reliable cross-modal intelligence.

%% file: sections/4_methodology/0_index.tex
\chapter{EigenTrack: Spectral Detection Framework}
\label{chap:methodology_eigentrack}

\noindent\textit{This chapter is based in part on the author’s publication 
``EigenTrack: Spectral Activation Feature Tracking for Hallucination and Out-of-Distribution Detection in LLMs and VLMs'' \cite{ettori2025eigentrackspectralactivationfeature}. 
The text and figures have been expanded and reformulated for clarity and completeness. Paper available on arXiv and submitted to ICASSP 2026 conference, under review. See Appendix~A}

\input{sections/4_methodology/eigentrack_1}

\input{sections/4_methodology/eigentrack_2}
\input{sections/4_methodology/eigentrack_3}
\input{sections/4_methodology/eigentrack_4}
\input{sections/4_methodology/eigentrack_5}

\chapter{RMT-KD: Random Matrix Distillation}
\label{chap:methodology_r bmtkd}

\noindent\textit{This chapter is based in part on the author’s publication 
``RMT-KD: Random Matrix Theoretic Causal Knowledge Distillation'' 
\cite{ettori2025rmtkdrandommatrixtheoretic}. 
The text and figures have been expanded and reformulated for clarity and completeness. Paper available on arXiv and submitted to ICASSP 2026 conference, under review. See Appendix~A}

\input{sections/4_methodology/rmtkd_1}
\input{sections/4_methodology/rmtkd_2}

\input{sections/4_methodology/rmtkd_3}

\input{sections/4_methodology/rmtkd_4}

\input{sections/4_methodology/rmtkd_5}

%% file: sections/4_methodology/eigentrack_1.tex
\section{Methodology}
\label{methodology:eigentrack_methodology}

EigenTrack introduces a real-time, interpretable framework for detecting hallucinations and out-of-distribution (OOD) behavior in large language and vision–language models. Rather than relying on surface-level probabilities, EigenTrack monitors how the internal geometry of activations evolves during generation. The key idea is that when a model begins to hallucinate or drift from the training distribution, its internal representations lose structure and become increasingly isotropic. This degradation can be quantified through the eigenvalue spectrum of hidden-layer covariances. 

Hallucinations in large generative models often emerge gradually, not as isolated anomalies. During early steps of generation, activations are structured and highly correlated, reflecting a coherent internal reasoning process. As hallucination begins, these correlations decay and the internal state becomes more random. Random Matrix Theory (RMT) provides a natural lens for analyzing this behavior. In- distribution activations produce covariance spectra with a few large, dominant eigenvalues. When representations lose structure, the spectrum collapses toward the Marchenko–Pastur (MP) law, which characterizes uncorrelated random noise. By tracking spectral statistics over time, one can therefore detect when the model’s reasoning dynamics begin to diverge from stable behavior.

\begin{figure}[h]
    \centering
    \includegraphics[width=0.65\textwidth]{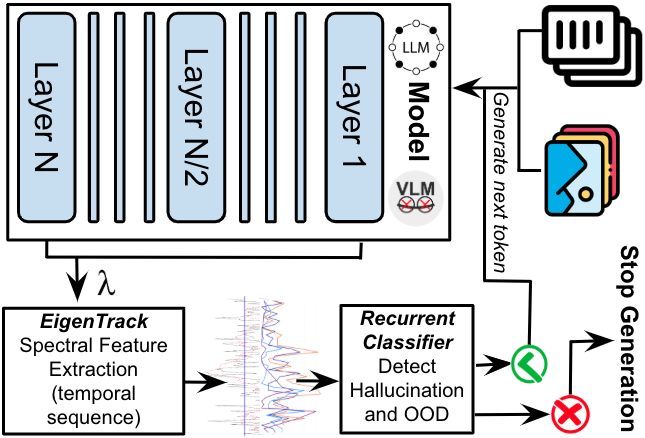}
    \caption[General architecture of EigenTrack]{General architecture of EigenTrack: Spectral features extracted from hidden activations are streamed into a recurrent discrepancy detector, which outputs early warnings of hallucination or OOD drift. In this pipeline, the problem is framed as binary classification.}
    \label{fig:eigentrack_general_architecture}
\end{figure}

\subsection*{Overview of the Framework}

EigenTrack operates as an auxiliary monitoring head attached to any pretrained model, requiring no modification of model parameters.  
The pipeline comprises three components:

\begin{itemize}
    \item \textbf{Spectral Feature Extraction:}  
    Hidden activations from selected layers are collected in a sliding temporal window. Their covariance matrices are analyzed through a truncated singular value decomposition (SVD) to obtain eigenvalues describing the structure of the hidden-state manifold.

    \item \textbf{Spectral Discrepancy Detection:}  
    From each window, a compact vector of spectral descriptors is computed, including entropy, leading eigenvalues, spectral gaps, and divergence from the MP distribution. These descriptors reflect how structured or random the current representations are.

    \item \textbf{Temporal Modeling and Early Warning:}  
    A lightweight recurrent network observes the evolution of spectral descriptors over time. By learning patterns associated with representational collapse, it outputs a probability that the model is entering a hallucinatory or OOD regime.
\end{itemize}

\subsection*{Spectral Feature Extraction from Hidden Representations}

Let the model layers be \( L_1, L_2, \ldots, L_m \).  
At each generation step \( t \), activations from the selected layers are concatenated: $v_t = [h_{1,t} \Vert h_{2,t} \Vert \cdots \Vert h_{m,t}], \quad h_{\ell,t} \in \mathbb{R}^{d}$.
To capture short-term temporal evolution, EigenTrack maintains a sliding window of the most recent \( N \) steps: $H_t = \begin{bmatrix} v_{t-N+1},\, \dots,\, v_t \end{bmatrix}^\top \in \mathbb{R}^{N \times md}$.
 The empirical covariance is then \( C_t = \frac{1}{N} H_t^{\top} H_t \)
 and its eigenvalues are obtained efficiently via the truncated singular value decomposition \( H_t = U_t \Sigma_t V_t^{\top}, \quad \lambda_{t,i} = \frac{\sigma_{t,i}^2}{N} \)
 Because \( N \ll md \), the SVD is computationally cheap, and randomized or incremental updates allow near real-time operation during autoregressive decoding. Each \( \lambda_{t,i} \) measures how variance is distributed across activation directions, providing a geometric fingerprint of the model’s internal dynamics.

\begin{figure}[h]
    \centering
    \includegraphics[width=\textwidth]{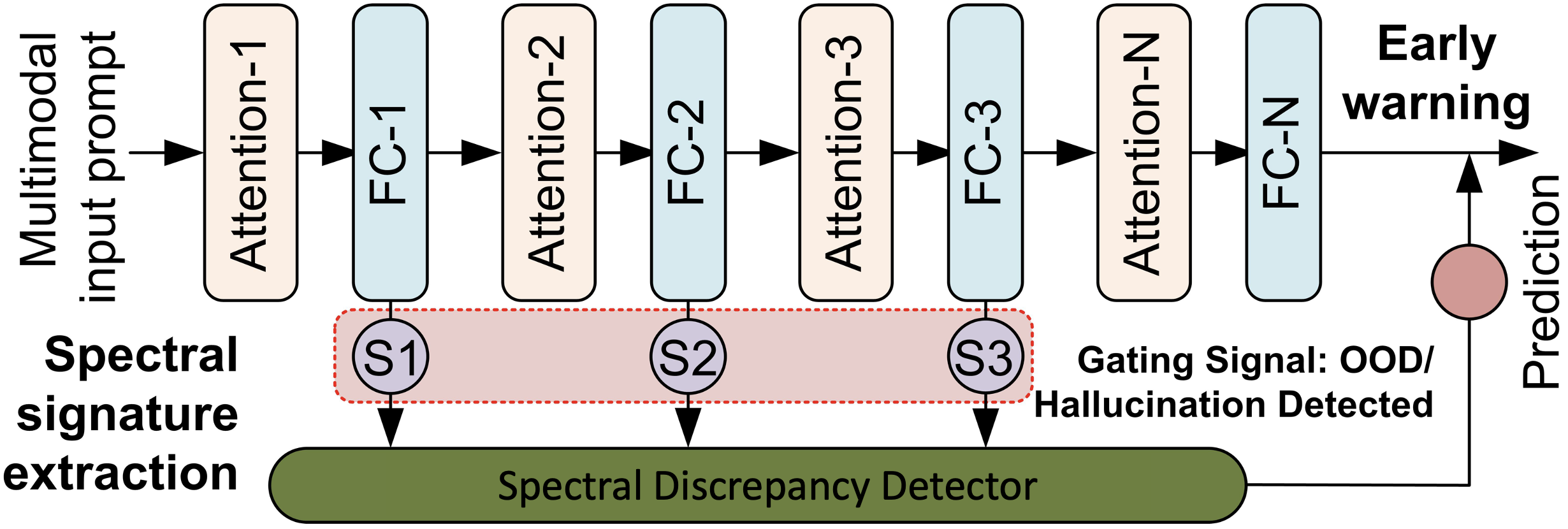}
    \vspace{5pt}
    \caption[Layer-level spectral signature extraction]{Layer-level spectral signature: extraction and temporal tracking. Selected layers feed activations into the spectral feature extractor, which computes eigenvalue-based descriptors and streams them to the discrepancy detector.}
    \label{fig:eigentrack_layer_architecture}
\end{figure}

\subsection*{Spectral Descriptors}

From the spectrum \( \{\lambda_{t,i}\} \), EigenTrack derives a compact vector \( F_t \) of interpretable descriptors. It includes 22        features in total, among which we have:

\begin{itemize}
\item \textbf{Leading Eigenvalue Sum:}
    \begin{align}
        s_t = \sum_{i=1}^{5} \lambda_{t,i}
    \end{align}
    measuring how much variance is concentrated in the top principal directions. Larger values indicate low-rank, structured representations, while smaller values suggest diffuse or noisy activations.

    \item \textbf{Spectral Entropy:}
    \begin{align}
        S_t = -\sum_{i} p_{t,i} \log p_{t,i}, \qquad p_{t,i} = \frac{\lambda_{t,i}}{\sum_{j} \lambda_{t,j}}
    \end{align}
    where high entropy indicates isotropic, less informative embeddings, and low entropy reflects concentrated, coherent feature directions.

    \item \textbf{KL Divergence from the Marchenko--Pastur Baseline:}
    \begin{align}
        D_{KL}\big(p(\lambda_t) \,\Vert\, \rho_{MP}(\lambda)\big)
    \end{align}
    where $\rho_{MP}(\lambda)$ denotes the Marchenko--Pastur eigenvalue density. Small divergence suggests noise-like, unstructured activations; larger divergence indicates meaningful statistical structure. The same also applies when quantifying the difference in distributions with the \textbf{Wasserstein distance}, which is another feature we consider in our classifier.

    \item \textbf{Tracy--Widom Fluctuation of the Leading Eigenvalue:} 
    The probability that $\lambda_{t,1}$ significantly deviates from the MP bulk edge $\lambda_{+}$:
    \begin{align}
        \mathbb{P}\big(\lambda_{t,1} > \lambda_{+} + \delta\big)
    \end{align}
    A substantial deviation indicates emergence of a dominant, non-random direction (signal-bearing subspace).

    \item \textbf{Spectral Gap Ratio:}
    \begin{align}
        g_{t,i} = \frac{\lambda_{t,i}}{\lambda_{t,i+1}}, \qquad i \in \{1,\dots,k\}
    \end{align}
    capturing how sharply leading eigenvalues detach from the rest. Larger gaps mark transitions between structured and noise-dominated subspaces.

    \item \textbf{Spectral Skewness:}
    \begin{align}
        \gamma_t = \frac{1}{d} \sum_{i=1}^{d} 
        \left(\frac{\lambda_{t,i} - \bar{\lambda}_t}{\sigma_t}\right)^3
    \end{align}
    Positive skewness implies a heavy tail of large eigenvalues (structured information), while near-zero skewness suggests noise-like balance.
\end{itemize}

\begin{figure}[h]
    \centering
    \includegraphics[width=\textwidth]{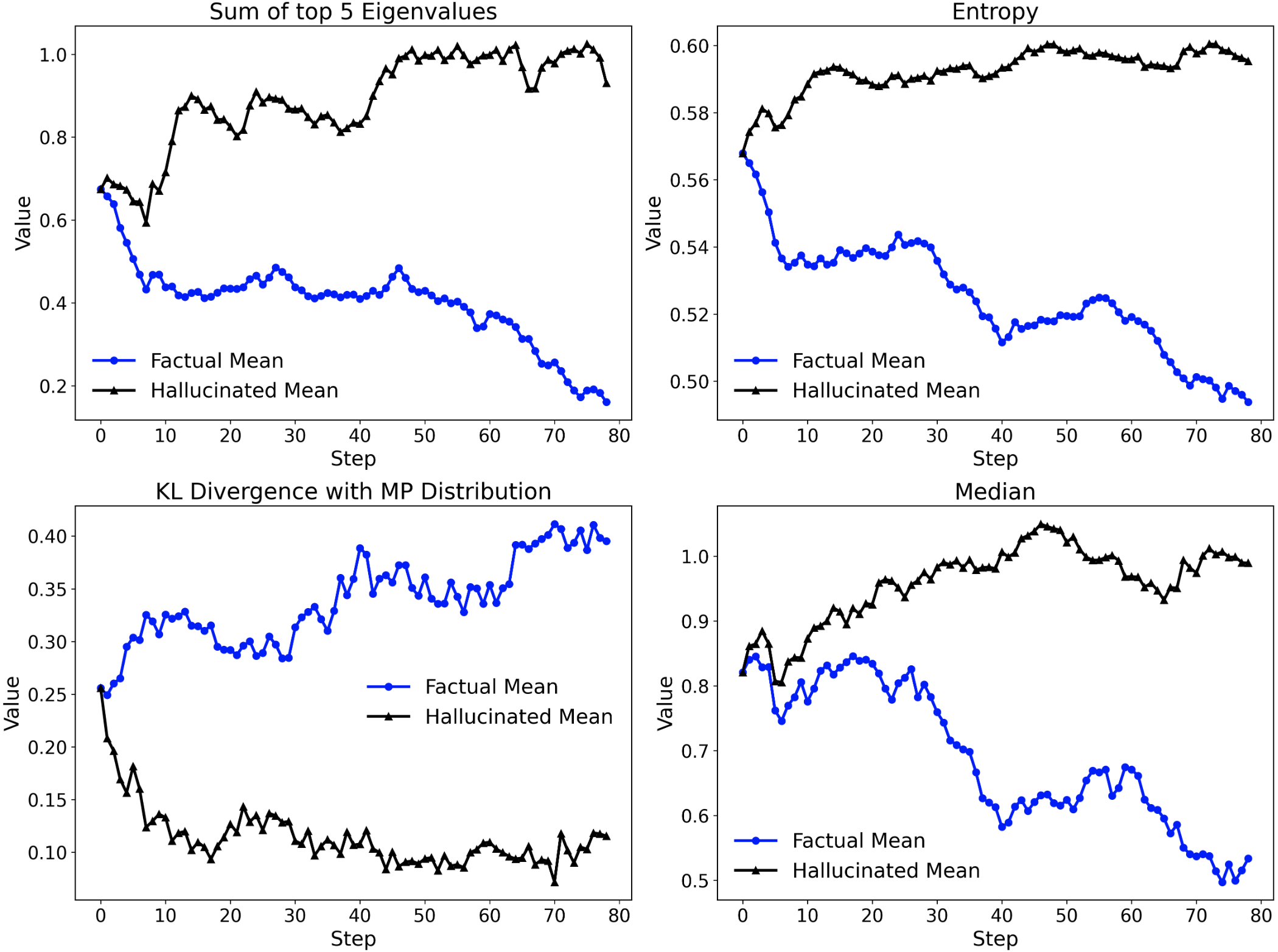}
    \caption[Temporal evolution of spectral statistics]{Temporal evolution of spectral statistics: The plots illustrate how spectral features evolve over time. Hallucinated sequences concentrate variance in a few dominant directions, producing higher cumulative sums of the top eigenvalues (top-left). Their spectra are flatter and more dispersed, resulting in elevated entropy values (top-right). From a random matrix theory perspective, if activations were purely uncorrelated noise, their eigenvalue distribution would follow the Marčenko–Pastur law \cite{MarchenkoPastur1967}. Relative to this baseline, hallucinated sequences remain closer to the noise-like regime, yielding lower KL divergence values (bottom-left), while factual sequences diverge more strongly, consistent with structured and informative dynamics. The median eigenvalue (bottom-right) is higher and more stable for hallucinations, whereas factual sequences show lower and more variable medians that correlate with model confidence.}
    \label{fig:eigentrack_spectral_evolution}
\end{figure}

\subsection*{Temporal Modeling with Recurrent Tracking}

A single spectral snapshot can reveal instability, but hallucinations typically evolve gradually as the internal representation quality degrades over time. To capture this temporal evolution, EigenTrack treats the descriptors as a time series \( \{F_1, F_2, \ldots, F_T\} \) and processes them using a small recurrent neural network. The feature sequence \( (F_1, \dots, F_T) \) is modeled as a multivariate time series and processed by a lightweight recurrent model (RNN, GRU, or LSTM). At each step, \( F_t \) enters the recurrent cell, hidden states propagate contextual information, and a feed-forward head outputs a binary logit corresponding to in-distribution versus anomalous context. Weight sharing across time ensures that the parameter count remains independent of sequence length, allowing the model to learn characteristic spectral signatures associated with stable or unstable dynamics.

This recurrent formulation provides three main advantages. First, it preserves temporal continuity by maintaining a memory of previous spectral states, enabling the detection of gradual drifts in representation quality. Second, it is computationally efficient, with updates occurring in constant time per step and minimal additional overhead. Third, it preserves causality by processing only past information, making it suitable for real-time inference and early warning during generation. At each generation step, the hidden state evolves according to \( z_t = \text{RNN}(F_t, z_{t-1}) \), and the anomaly probability is computed as \( \hat{y}_t = \sigma(W z_t + b) \). If \( \hat{y}_t > \tau \), the detector raises a gating signal that can intervene by halting decoding, triggering retrieval grounding, or lowering the generation temperature. The recurrent head is highly compact, containing only a few thousand parameters, and can operate concurrently with model inference without significant latency.

EigenTrack supports tuning of the monitored layer set \( L \), window length \( N \), number of spectral features \( k \), and recurrent hidden size to balance accuracy and latency. The pipeline also extends naturally to multimodal architectures by constructing \( H_t \) from cross-modal fusion layers or vision encoder blocks, enabling spectral monitoring across both language and visual representations within a unified latent space.

\subsection*{Efficiency and Practical Considerations}

EigenTrack is designed to be lightweight, modular, and fast enough to operate in real time alongside large generative models. In practice, only a small subset of layers is monitored, typically one every three or four transformer blocks, since adjacent layers exhibit highly correlated spectral behavior. This sampling strategy minimizes data transfer without sacrificing sensitivity. Covariance estimation is implemented through truncated or incremental SVD, which avoids full matrix reconstruction and enables efficient eigenvalue updates at each decoding step. The recurrent head itself contains only a few thousand parameters and performs constant-time updates, making it several orders of magnitude smaller and faster than transformer-based temporal models. For this reason, recurrent networks were preferred over transformers: they preserve causal processing, require negligible memory, and operate with minimal computational overhead. As a result, EigenTrack can be integrated directly into autoregressive decoding loops, providing continuous reliability monitoring with only a small additional latency.

%% file: sections/4_methodology/eigentrack_2.tex
\section{Theoretical Justification}
\label{methodology:eigentrack_theoretical_justification}

\subsection*{Spectral Geometry as a Diagnostic Lens}

The theoretical foundations of EigenTrack arise from the observation that large neural networks, 
despite their nonlinear nature, exhibit emergent regularities that can be studied through the 
geometry of their hidden representations. 
The key insight is that during normal operation, the network’s representations inhabit a structured, 
low-dimensional manifold, while under distributional shift or hallucination-inducing conditions, 
this manifold collapses toward isotropic randomness. 

Modern decoder-only LLMs apply LayerNorm at every transformer block \cite{radford2019language}, 
centering activations and scaling them to unit variance. Combined with the near-orthogonality of 
projection matrices induced by weight decay and adaptive optimization methods such as Adam 
\cite{kobayashi2024weight}, this makes per-token activations approximately mean-zero and isotropic. 
Under these assumptions, the activations of each layer can be modeled as random samples from an 
approximately isotropic Gaussian ensemble, and their covariance spectra follow well-known laws from Random Matrix Theory (RMT). Departures from this baseline indicate the emergence of structure or instability within the model’s 
internal representations. When the network encounters out-of-distribution (OOD) inputs or 
hallucination-inducing contexts, hidden activations exhibit correlated, low-rank perturbations that 
follow the spiked covariance model. 

\subsection*{The Random Matrix Baseline}

Random Matrix Theory provides analytical tools to model the spectrum of large, unstructured covariance matrices. 
Consider a matrix $X \in \mathbb{R}^{N \times d}$ whose entries are independent and identically distributed 
with zero mean and variance $\sigma^2$. In the asymptotic limit where $N, d \to \infty$ and the aspect ratio 
$q = d / N$ is fixed, the empirical distribution of eigenvalues of the sample covariance 
$C = \frac{1}{N} X^{\top} X$ converges to the \emph{Marchenko--Pastur (MP) law} \cite{MarchenkoPastur1967}. 
The density of eigenvalues under this null model is given by
\begin{align}
    \rho_{\text{MP}}(\lambda) = 
\frac{1}{2\pi \sigma^2 q \lambda}
\sqrt{(\lambda_{+} - \lambda)(\lambda - \lambda_{-})},
\quad \lambda \in [\lambda_{-}, \lambda_{+}]    
\end{align}
with the support edges defined as \( \lambda_{\pm} = \sigma^2 (1 \pm \sqrt{q})^2 \) This distribution characterizes the ``noise floor'' of high-dimensional representations. 
Any activation covariance spectrum that aligns closely with $\rho_{\text{MP}}$ 
can be interpreted as statistically equivalent to a random ensemble with no significant correlations. 
Conversely, systematic deviations, such as the emergence of outlier eigenvalues or heavy tails, 
signal the existence of structured correlations and dependencies in the underlying representations. A similar concept, related to the top eigenvalue deviation, is modeled by the Tracy-Widom distribution.

An analogous principle holds for uncentered activations or correlation matrices 
whose entries are symmetrized rather than formed from $X^{\top}X$. 
In this setting, the eigenvalue density converges to the \emph{Wigner semicircle law} \cite{wigner1958distribution}, 
which describes the distribution of eigenvalues of symmetric random matrices. 
The Marchenko--Pastur, Tracy-Widom and Wigner laws form the theoretical backbone of EigenTrack’s null model: 
they represent the spectral geometry of complete randomness against which meaningful structure can be contrasted.

\subsection*{Signal Emergence on the Spiked Covariance Model}

The empirical covariance of a trained model’s activations rarely conforms to the pure MP law. 
Instead, it follows a \emph{spiked covariance model} \cite{johnstone2001distribution}, 
in which a low-rank signal subspace is embedded within a high-dimensional noisy background: \( C_t = \sigma^2 I + \sum_{i=1}^{k} \theta_i u_i u_i^{\top} \)
 Here, $\sigma^2 I$ represents isotropic noise, and 
$\{u_i\}_{i=1}^{k}$ are orthogonal directions capturing coherent, semantically meaningful structure. 
The scalars $\theta_i$ denote the signal strengths along those directions. 
In the spectral domain, the presence of such low-rank perturbations produces 
outlier eigenvalues that detach from the MP bulk. 
This phenomenon is governed by the \emph{Baik--Ben Arous--Péché (BBP) phase transition} \cite{baik2005phase}, 
which defines a critical signal-to-noise threshold:
\begin{align}
    \lambda_s > \sigma^2 (1 + \sqrt{q}) \quad \Longrightarrow \quad 
\text{signal eigenvalue separates from the bulk.}    
\end{align}
When $\lambda_s$ falls below this threshold, the signal direction becomes indistinguishable from noise, 
and its corresponding eigenvalue reabsorbs into the random bulk. 
This threshold delineates the boundary between ordered and disordered regimes in the representation space. 
Within deep neural networks, such transitions correspond to phases where 
latent representations lose alignment with meaningful semantic directions—an effect that empirically correlates 
with hallucinations or failures of grounding in LLMs and VLMs.

\subsection*{From Static Spectra to Temporal Dynamics}

Traditional RMT analyses treat $C_t$ as a static object, characterizing one snapshot of representational geometry. 
However, in autoregressive models, activations evolve dynamically with each generated token. 
Let $\{\rho_t(\lambda)\}_{t=1}^{T}$ denote the sequence of spectral densities observed across a generation trace. 
EigenTrack extends the RMT framework by modeling this sequence as a stochastic process in the space of spectral measures. 
This approach captures not only instantaneous anomalies but also the temporal precursors of instability.

Formally, define a set of spectral statistics $\phi_t = f(\rho_t)$ summarizing 
the current spectral state (e.g., curvature, skewness, or divergence from MP baseline). 
The trajectory $\{\phi_t\}$ can then be regarded as a dynamical system: \( \phi_{t+1} = \mathcal{F}(\phi_t) + \epsilon_t \) where $\mathcal{F}$ denotes the underlying transition operator induced by the model’s internal dynamics 
and $\epsilon_t$ is a stochastic perturbation. 
When the model processes coherent, in-distribution input, $\mathcal{F}$ is approximately stationary: 
the geometry of the activation manifold evolves smoothly. 
Under hallucination or distributional shift, however, the transition dynamics 
become unstable, producing abrupt spectral fluctuations and non-stationarity in $\phi_t$. 
These deviations serve as early-warning signals of representational collapse.

\subsection*{Implications for Model Reliability}

In summary, the theoretical justification of EigenTrack rests on three pillars:

\begin{enumerate}
    \item \textbf{Random Matrix Theory:} establishes the null hypothesis of isotropic randomness, 
    represented by the Marchenko--Pastur and Tracy--Widom distribution. 

    \item \textbf{Spiked Covariance Theory:} explains how deviations from the MP law arise from coherent, 
    task-aligned structure, and how their disappearance signals representational collapse.

    \item \textbf{Temporal Spectral Dynamics:} connects the evolution of eigenvalue distributions 
    to the stability of the model’s internal manifold during generation.
\end{enumerate}

Together, these principles provide a rigorous mathematical framework linking 
spectral geometry to reliability in deep generative models. 
EigenTrack operationalizes this connection by tracking the time evolution of the spectrum, 
transforming abstract random-matrix theory into a practical diagnostic of hallucination risk.

%% file: sections/4_methodology/eigentrack_3.tex
\section{Experimental Setup}
\label{methodology:eigentrack_experimental_setup}

\subsection*{Overview of Evaluation Protocol}

The experimental evaluation of EigenTrack is designed to assess its ability to detect hallucinations and out-of-distribution (OOD) behavior across a wide range of open-source large language models (LLMs) and vision-language models (VLMs). All experiments are conducted on publicly available architectures from the HuggingFace Hub, spanning models from 1B to 8B parameters. The evaluated families include LLaMa, Qwen, Mistral, and LLaVa, each tested in both base and instruction-tuned variants.  

For every model, generation is limited to sequences of up to 128 tokens. During inference, the full stream of hidden activations across layers is captured in real time. To facilitate efficient computation, caching mechanisms are activated and configured to retain all intermediate hidden states. The generation temperature is fixed at 0.2 to ensure deterministic and stable token sampling, reducing stochastic variability across runs.

\subsection*{Hallucination Detection Setup}

To evaluate hallucination detection, we adopt a controlled and reproducible QA-based pipeline built upon the HaluEval dataset, which derives from HotpotQA. Each passage is paired with two types of questions: its true, factually grounded question, and a randomly selected unrelated question generated by LLaMa-8B. The corresponding answers to these paired questions yield factual (non-hallucinated) and hallucinated model outputs respectively.  

This setup involves three distinct interacting components, instantiated as separate models:

\begin{itemize}
    \item \textbf{Main Model:} The model under analysis (typically the smallest one in the family) whose hidden activations are monitored by EigenTrack.
    \item \textbf{Question Generator:} A larger model (LLaMa-8B) used to produce semantically unrelated or misleading questions that trigger hallucinations.
    \item \textbf{Answer Judge:} A separate LLM employed to automatically verify factual consistency between the question, passage, and generated response.
\end{itemize}

Through this tri-model interaction, the system constructs a large, automatically labeled dataset of hallucinated versus truthful generations, without the need for manual annotation.  
For multimodal evaluation, the same procedure is extended to VLMs such as LLaVa, where text passages are combined with images from the Flickr8k dataset. Again, LLM-as-a-judge is employed to verify if the answer was factual or hallucinated.

\subsection*{Out-of-Distribution Detection Setup}

EigenTrack is also evaluated on OOD detection tasks to test its sensitivity to semantic domain shift.  
For this purpose, the WebQuestions dataset is used as the in-distribution (ID) source, while the Eurlex dataset, composed of European legal-domain queries, is treated as the OOD counterpart.  
Since Eurlex data lies outside the pretraining distribution of the tested models, it naturally induces OOD behavior.  

As in the hallucination experiments, each model receives one question at a time and generates a textual response of up to 128 tokens, with activations streamed during decoding.  
For VLMs, we give images from Flickr8k as context and a prompt to describe them for ID samples, for OOD ones, we ask to respond to EurLex questions.  
This consistent setup allows EigenTrack to be tested across both unimodal and multimodal configurations without architecture-specific modifications.

\subsection*{Classifier Training}

At each generation step, EigenTrack converts the hidden activations into a compact spectral representation.  
This produces a time series of spectral descriptors, already defined in the previous section, which capture the temporal evolution of the representation geometry.  
Each sequence of spectral vectors serves as input to a lightweight recurrent classifier designed to distinguish between stable (in-distribution, factual) and unstable (OOD, hallucinated) trajectories. Three recurrent architectures are considered: a simple RNN, a GRU, and an LSTM.  
Each classifier consists of a linear input projection, a single recurrent layer, and a binary output head producing a probability score for anomalous behavior.  
The recurrent hidden state dimension is set to 16.  
All classifiers are trained using the Adam optimizer with learning rate \(10^{-3}\) and weight decay \(10^{-4}\).  
Training is performed with a binary cross-entropy loss over labeled sequences.  
A final linear projection layer is applied after the recurrent unit to map the hidden representation to the two output logits.

\subsection*{Evaluation Metrics}

Detection performance is measured using the area under the receiver operating characteristic curve (AUROC).  
This metric quantifies the separability between the positive (hallucinated or OOD) and negative (factual or ID) classes.  
An AUROC of 0.5 corresponds to random guessing, while a score of 1.0 indicates perfect discrimination.  
All results are reported as mean AUROC values averaged over multiple runs and datasets for both LLM and VLM settings.  

This evaluation protocol ensures that the results reflect not only instantaneous performance but also the robustness of the proposed spectral tracking method across different architectures, modalities, and failure types.

%% file: sections/4_methodology/eigentrack_4.tex
\section{Results}
\label{methodology:eigentrack_results}

This section reports the empirical performance of EigenTrack on hallucination detection and out-of-distribution (OOD) shift identification across language-only and vision–language backbones. We summarize results with two comparative bar plots and two tables. For each artifact, we provide an interpretive commentary that links the observed trends back to the spectral-temporal design of the method.

\begin{figure}[h]
    \centering
    \includegraphics[width=\linewidth]{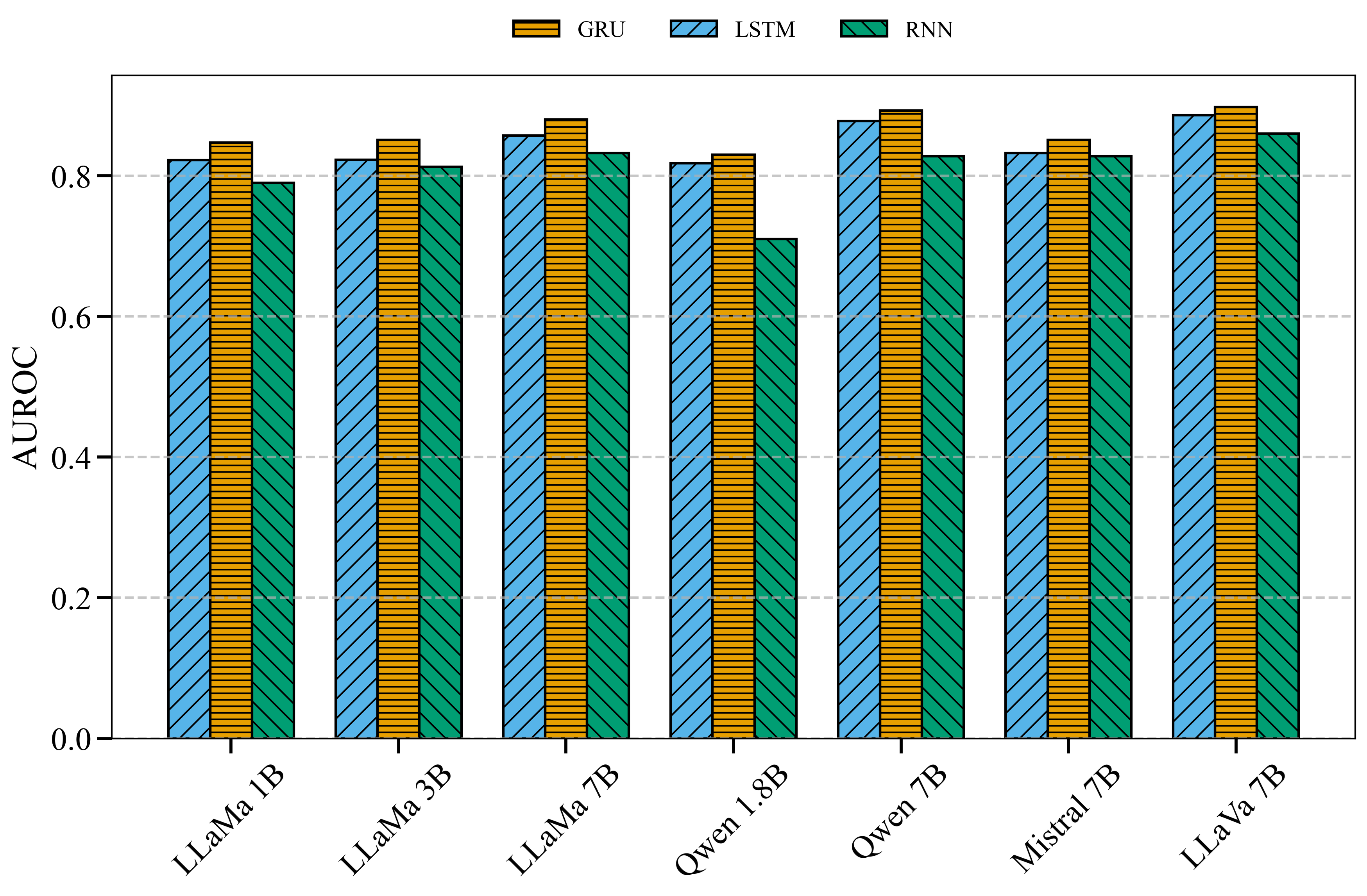}
    \vspace{5pt}
    \caption[AUROC for hallucination detection]{AUROC for hallucination detection: across LLaMa (1B/3B/7B), Qwen (1.8B/7B), Mistral-7B, and LLaVa-7B using RNN, GRU, and LSTM temporal heads.}
    \label{fig:auroc_hallucination}
\end{figure}

~\ref{fig:auroc_hallucination} reveals a consistent and interpretable trend across all evaluated model families. EigenTrack achieves strong performance, maintaining AUROC values between 0.82 and 0.94 across the full spectrum of model sizes. Among the recurrent architectures, GRUs deliver the highest overall performance, followed closely by LSTMs, while vanilla RNNs trail slightly behind. This ranking underscores the advantage of gated recurrent mechanisms—particularly their ability to retain long-term dependencies and selectively filter spectral fluctuations over time. The improved sensitivity of GRUs suggests that hallucination events are preceded by subtle, low-frequency variations in spectral statistics that require controlled temporal integration rather than simple memory recurrence.

A clear correlation emerges between model size and detection performance. Within the LLaMa family, for instance, AUROC improves steadily from approximately 0.84 in LLaMa-1B to nearly 0.89 in LLaMa-7B. This positive scaling trend indicates that larger models, by virtue of their higher representational capacity, generate more structured and separable spectral signatures. These richer eigenvalue dynamics allow EigenTrack’s recurrent back-end to identify deviations from the stable spectral regimes that characterize factual, coherent generations. The strongest results are observed on 7B-scale architectures, particularly Qwen-7B and LLaVa-7B, where GRUs reach AUROC values exceeding 0.93, marking a high degree of discriminative reliability. ~\ref{fig:auroc_hallucination} demonstrates that EigenTrack generalizes effectively across diverse foundation models and that its spectral-temporal framework benefits both from larger model capacities and from the inductive bias of gated recurrence. The consistency of these results supports the hypothesis that hallucinations manifest as measurable perturbations in the eigenvalue geometry of activation covariances, patterns that GRU-based tracking captures with remarkable precision.

\begin{figure}[h]
    \centering
    \includegraphics[width=\linewidth]{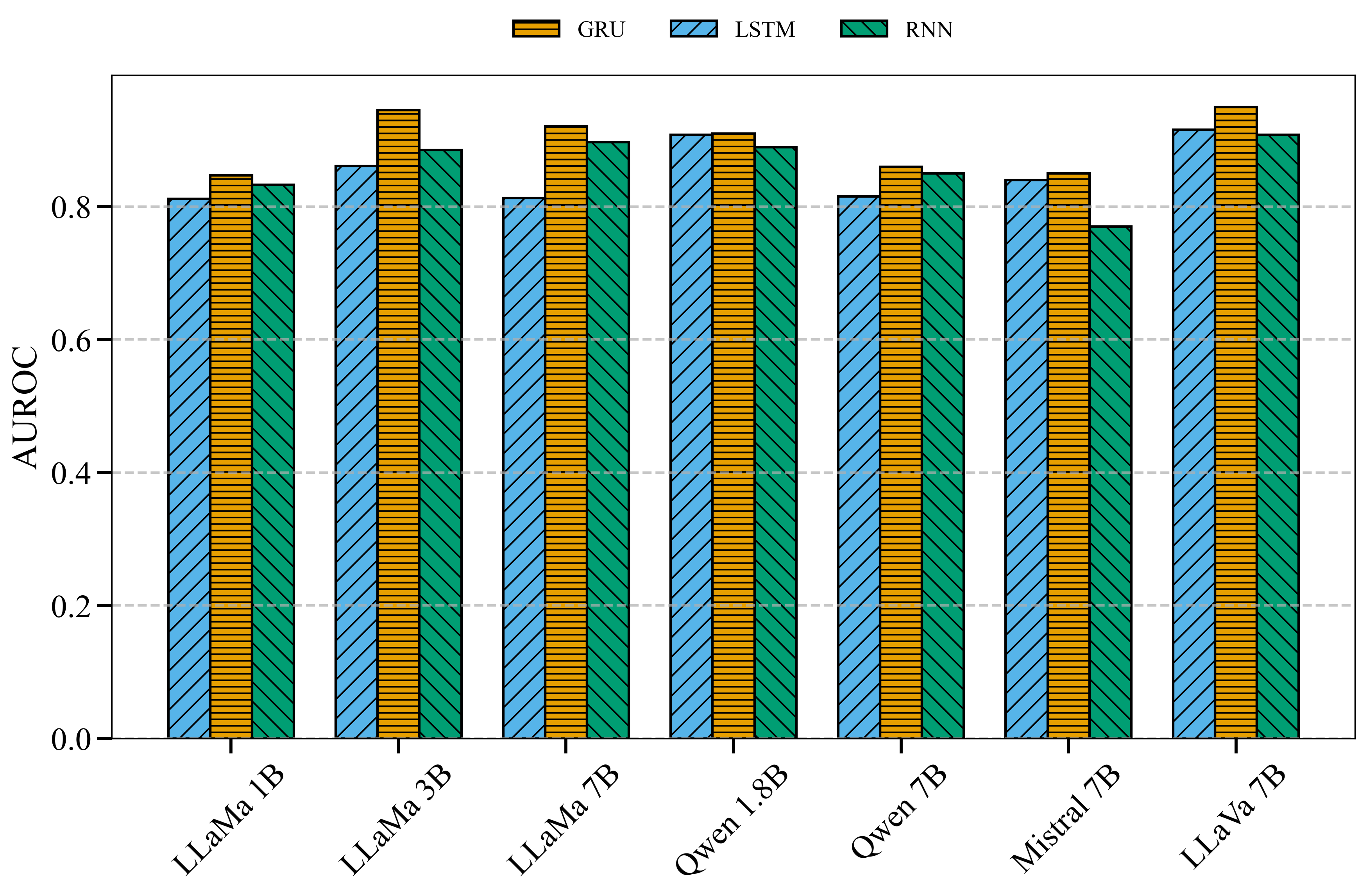}
    \vspace{5pt}
    \caption[AUROC for OOD detection]{AUROC for OOD detection: on the same set of models using RNN, GRU, and LSTM temporal heads. Same idea to classify OOD instead of hallucination.}
    \label{fig:auroc_ood}
\end{figure}

The AUROC results for OOD detection, shown in ~\ref{fig:auroc_ood}, exhibit a pattern closely mirroring that of hallucination detection while achieving overall higher performance levels. Across all tested models, EigenTrack attains AUROC scores ranging from approximately 0.85 to 0.96, reflecting robust and consistent OOD sensitivity. Once again, GRU-based classifiers consistently outperform LSTM and vanilla RNN counterparts, underscoring the importance of gated temporal memory in capturing the gradual spectral shifts that accompany domain transitions. The performance gap between GRUs and simpler recurrent cells is less pronounced in smaller models such as LLaMa-1B and Qwen-1.8B, suggesting that even basic recurrence can effectively capture spectral drift when the representational complexity is limited, while gated mechanisms become more advantageous as model scale and spectral richness increase.

A clear upward trend is observed with increasing model scale. Larger architectures such as LLaMa-7B and LLaVa-7B exhibit AUROC values surpassing 0.92, implying that more complex networks produce activation spectra with richer statistical structure and sharper deviations under distributional shift. These results demonstrate that EigenTrack’s spectral features generalize across scales, capturing the transition from in-distribution to out-of-distribution regimes through eigenvalue dispersion and Marchenko–Pastur divergence.

Interestingly, OOD detection tends to outperform hallucination detection overall, as domain shifts often trigger broader and more coherent changes in the global spectral landscape than the subtler local instabilities preceding hallucinations. This observation aligns with the theoretical expectation that shifts in data manifold geometry induce distinct, high-amplitude spectral perturbations that are easier to isolate. By integrating these spectral descriptors with lightweight GRU-based temporal modeling, EigenTrack achieves strong OOD generalization with minimal architectural overhead, validating its efficiency and interpretability as a spectral–temporal reliability framework for large-scale models.

\begin{table}[h]
\centering
\caption[SOTA COMPARISON ON LLAMA
]{SOTA COMPARISON ON LLAMA}
\vspace{20pt}
\label{tab:sota_comparison}
\small
\setlength{\tabcolsep}{10pt}
\renewcommand{\arraystretch}{1.8}
\begin{tabular}{l ccc l ccc}
\toprule
\multicolumn{4}{c}{\textbf{Hallucination Detection}} & \multicolumn{4}{c}{\textbf{OOD Detection}} \\
\cmidrule(lr){1-4}\cmidrule(lr){5-8}
\textbf{Method} & \textbf{1B} & \textbf{3B} & \textbf{7B} &
\textbf{Method} & \textbf{1B} & \textbf{3B} & \textbf{7B} \\
\midrule
\textbf{EigenTrack}  & \textbf{84.2} & \textbf{86.1} & \textbf{89.4} & \textbf{EigenTrack}  & \textbf{85.5} & \textbf{89.2} & \textbf{92.4} \\
LapEigvals           & 78.5 & 81.9 & \underline{87.1} & Cosine Distance & 81.9 & \underline{87.7} & 92.0 \\
INSIDE               & 75.3 & \underline{83.1} & 81.0 & Energy Score    & \underline{83.2} & 85.2 & 89.0 \\
SelfCheckGPT         & 73.9 & 80.4 & 80.9 & Max Softmax Prob & 70.1 & 71.0 & 72.0 \\
HaloScope            & \underline{82.0} & 82.7 & 86.1 & ODIN & 80.1 & 84.2 & \underline{92.1} \\
\bottomrule
\end{tabular}
\end{table}

\noindent ~\ref{tab:sota_comparison} presents the comparison between EigenTrack and representative state-of-the-art detectors on the LLaMa model family. It shows the AUROC comparison on LLaMa models (1B/3B/7B) between EigenTrack and representative baselines for hallucination and OOD detection. Each half of the table shows one task (Hallucination Detection and OOD Detection). 

For hallucination detection, EigenTrack achieves AUROC values of 84.2, 86.1, and 89.4 on the 1B, 3B, and 7B models, respectively, surpassing spectral baselines such as LapEigvals (81.9 on LLaMa-3B, 87.1 on LLaMa-7B) and self-consistency approaches like HaloScope and INSIDE, which remain below 83 on smaller models. The performance gap widens with model scale, emphasizing the capacity of EigenTrack’s temporal spectral modeling to capture evolving internal dynamics that become more pronounced in larger networks.  

In the OOD detection setting, EigenTrack again dominates across scales, reaching 92.4 AUROC on LLaMa-7B compared to 92.0 for Cosine Distance and 89.0 for Energy Score, while simpler baselines such as Max Softmax Probability fall sharply below 72. These differences underscore that static confidence-based metrics saturate quickly, whereas EigenTrack’s use of evolving spectral statistics maintains discriminative power across both subtle and pronounced domain shifts. 

These capture global representation dynamics that surface-level confidence methods (Max Softmax, ODIN) and snapshot spectral analyses (LapEigvals) miss. Baseline OOD methods are score-based without OOD supervision; their AUROC values are obtained by sweeping thresholds over in-distribution scores, ensuring threshold-independent and fair comparison.

\ref{tab:eigentrack_full_results} shows the comprehensive EigenTrack results for hallucination and out-of-distribution (OOD) detection across multiple language and vision–language backbones. The table reports both the AUROC and the F1 score for each model and temporal head. AUROC quantifies the overall discriminative power of EigenTrack in distinguishing normal from anomalous states, providing a threshold-independent indicator of reliability detection quality. The F1 score complements this by measuring the balance between precision and recall at the optimal operating point, reflecting how effectively the detector can identify failure instances without excessive false alarms. Together, these metrics capture both the robustness and practical usability of the spectral–temporal detection framework.

\begin{table}[h]
\centering
\setlength{\tabcolsep}{15pt}
\caption[FULL METRICS ACROSS MODELS]{FULL METRICS ACROSS MODELS}
\vspace{20pt}
\label{tab:eigentrack_full_results}
\small
\renewcommand{\arraystretch}{1.8}
\begin{tabularx}{\textwidth}{c X *{6}{c}}
\toprule
& & \multicolumn{3}{c}{\textbf{AUROC Full}} & \multicolumn{3}{c}{\textbf{F1 Full}} \\
\cmidrule(lr){3-5}\cmidrule(lr){6-8}
& \textbf{Model} & RNN & GRU & LSTM & RNN & GRU & LSTM \\
\midrule
\multirow{7}{*}{\rotatebox[origin=c]{90}{\textbf{Hallucination}}}
& LLaMa 1B   & 0.799 & 0.842 & 0.831 & 0.750 & 0.790 & 0.783 \\
& LLaMa 3B   & 0.832 & 0.861 & 0.844 & 0.779 & 0.808 & 0.794 \\
& LLaMa 7B   & 0.853 & 0.894 & 0.872 & 0.805 & 0.851 & 0.825 \\
& Qwen 1.8B  & 0.724 & 0.824 & 0.821 & 0.672 & 0.798 & 0.774 \\
& Qwen 7B    & 0.842 & 0.931 & 0.922 & 0.794 & 0.881 & 0.870 \\
& Mistral 7B & 0.864 & 0.888 & 0.871 & 0.812 & 0.839 & 0.819 \\
& LLaVa 7B   & 0.902 & 0.941 & 0.934 & 0.853 & 0.892 & 0.887 \\
\midrule
\multirow{7}{*}{\rotatebox[origin=c]{90}{\textbf{OOD}}}
& LLaMa 1B   & 0.825 & 0.855 & 0.852 & 0.776 & 0.814 & 0.802 \\
& LLaMa 3B   & 0.858 & 0.892 & 0.871 & 0.810 & 0.841 & 0.821 \\
& LLaMa 7B   & 0.879 & 0.924 & 0.897 & 0.829 & 0.874 & 0.847 \\
& Qwen 1.8B  & 0.762 & 0.872 & 0.846 & 0.713 & 0.821 & 0.796 \\
& Qwen 7B    & 0.867 & 0.948 & 0.936 & 0.817 & 0.898 & 0.885 \\
& Mistral 7B & 0.883 & 0.906 & 0.892 & 0.832 & 0.855 & 0.842 \\
& LLaVa 7B   & 0.923 & 0.958 & 0.946 & 0.873 & 0.906 & 0.897 \\
\bottomrule
\end{tabularx}
\vspace{-5pt}
\end{table}

%% file: sections/4_methodology/eigentrack_5.tex
\section{Ablation Studies}
\label{methodology:eigentrack_ablation}

\subsection*{Accuracy–Latency Trade-off}

This section investigates how the spectral dynamics captured by EigenTrack depend on temporal and architectural parameters. In particular, we study how the sliding-window size and the number of generated tokens influence the overall detection accuracy, measured by the Area Under the Receiver Operating Characteristic (AUROC). These ablation studies clarify the trade-off between response time and reliability, providing practical guidance for deploying EigenTrack in different operating regimes.

\subsection*{Sliding-Window Length and Latency}

\begin{figure}[h]
    \centering
    \includegraphics[width=\linewidth]{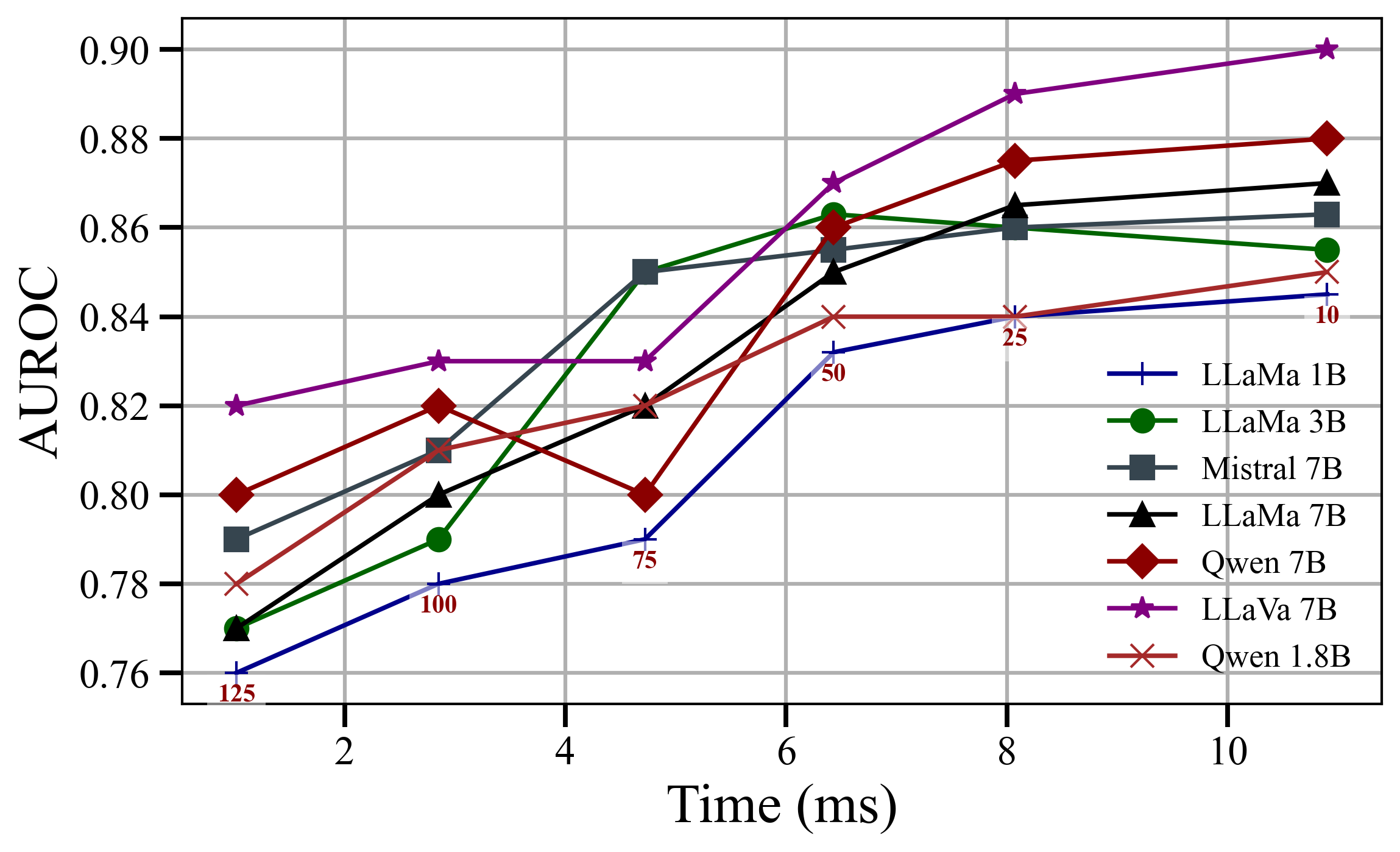}
    \vspace{10pt}
    \caption[AUROC–latency trade-off]{AUROC–latency trade-off: across large language and vision-language models. Shorter windows yield finer temporal resolution but higher latency, while longer windows improve efficiency at minor cost in accuracy. The shaded area indicates the optimal trade-off region. Experiments were conducted on a reduced sample (20\%) of the Hallucination dataset, using GRU as the classification head.}
    \label{fig:auroc_vs_sequence_length}
\end{figure}

Figure \ref{fig:auroc_vs_sequence_length} shows the evolution of AUROC as a function of inference latency across several large language and vision-language models, including LLaMa-1B, LLaMa-3B, LLaMa-7B, Qwen-1.8B, Qwen-7B, Mistral-7B, and LLaVa-7B. Each curve corresponds to a distinct model family evaluated using different sliding-window lengths for the GRU-based classifier that tracks spectral features over time.

Shorter windows enable EigenTrack to capture rapid fluctuations in the covariance spectrum of activations, thereby improving sensitivity to early anomalies. This finer temporal resolution leads to higher AUROC at the cost of increased computational load, since more windows must be processed per sequence. As the window length grows, fewer updates are required, reducing latency but also coarsening the temporal signal. The curves in Figure~\ref{fig:auroc_vs_sequence_length} reveal that accuracy saturates between approximately 25 and 50 tokens, marking an optimal trade-off region where inference time remains below 10 milliseconds while AUROC exceeds 0.88 for most models. This balance highlights that EigenTrack can be flexibly tuned depending on whether the application prioritizes real-time responsiveness or maximum detection precision.

\subsection*{Evolution with Number of Generated Tokens}

\begin{figure}[h]
    \centering
    \includegraphics[width=\linewidth]{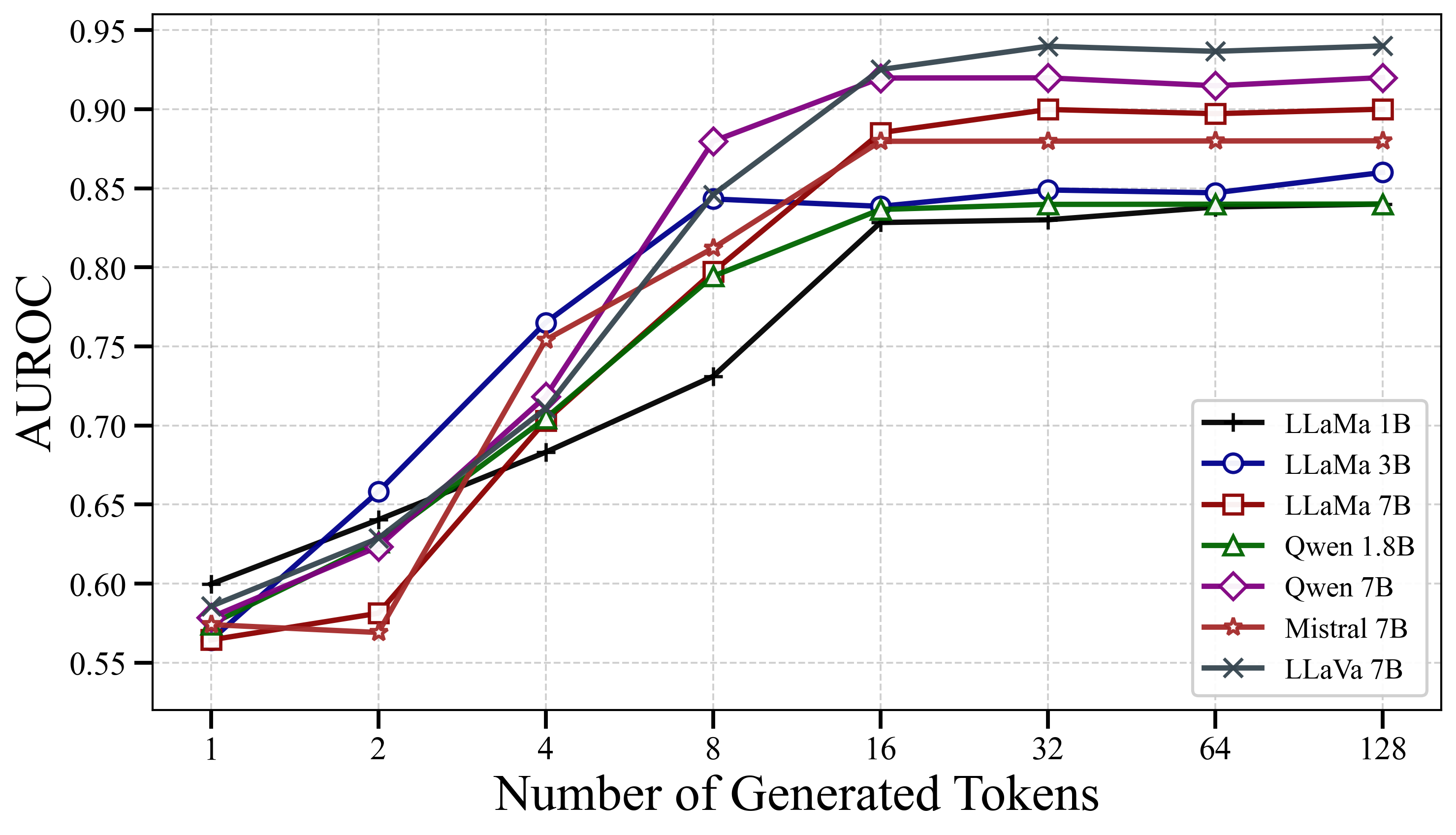}
    \vspace{10pt}
    \caption[AUROC as a function of the number of generated tokens.]{AUROC as a function of the number of generated tokens: accuracy improves sharply during early generation and saturates beyond 32 tokens, indicating that spectral signatures of hallucination and distributional shift arise early in the decoding process. Experiments were conducted on the Hallucination dataset, using GRU as the classification head.}
    \label{fig:auroc_vs_token_length}
\end{figure}

Figure \ref{fig:auroc_vs_token_length} illustrates AUROC as a function of the number of generated tokens. The results show that all evaluated models start near chance-level performance when only a few tokens have been produced, as the internal representations are still weakly informative. As generation progresses, the AUROC rises sharply within the first few tokens, reaching a stable plateau once sufficient contextual information has accumulated in the hidden activations. Across model families, the strongest improvement occurs between 8 and 16 tokens, after which the gains gradually diminish.

This pattern reveals that hallucination and out-of-distribution cues emerge early in the generation process. The stabilization of AUROC beyond approximately 32 tokens implies that additional computation provides limited improvement, suggesting that EigenTrack can detect reliability issues well before they manifest in the output text. In practice, this enables efficient monitoring strategies: short responses can be fully analyzed, whereas for long-form generation only the initial segments need continuous spectral tracking to achieve comparable accuracy.

\subsection*{Discussion}

These ablation studies show that EigenTrack delivers stable, interpretable performance across diverse architectures and temporal setups. Its reliability stems from the spectral evolution of activations rather than parameter tuning. The early AUROC stabilization indicates that hallucination-related spectral shifts appear almost immediately, enabling efficient real-time use. This balance of latency and precision makes EigenTrack a flexible framework, well-suited for both research diagnostics and large-scale safety monitoring in production.

%% file: sections/4_methodology/rmtkd_1.tex
\section{Methodology}
\label{methodology:rmtkd_methodology}

\subsection*{Problem Setting}

The goal of Random Matrix Theoretic Knowledge Distillation (RMT-KD) is to compress a trained network while preserving accuracy, latency, and energy efficiency. RMT-KD departs from sparsity- or heuristic-rank methods by identifying and keeping only the causal directions of hidden activations, as revealed by their spectral geometry. At a high level, training proceeds in stages: once validation accuracy crosses a stability threshold, the current layer is analyzed spectrally on a small calibration subset; outlier eigen-directions beyond the Marchenko–Pastur (MP) noise bulk are deemed informative and retained; a linear projection block implements the reduction; the reduced model self-distills from the previous checkpoint to recover accuracy; the loop repeats layer by layer until a target reduction or a quality bound is reached.

\begin{figure}[h]
    \centering
    \includegraphics[width=\textwidth]{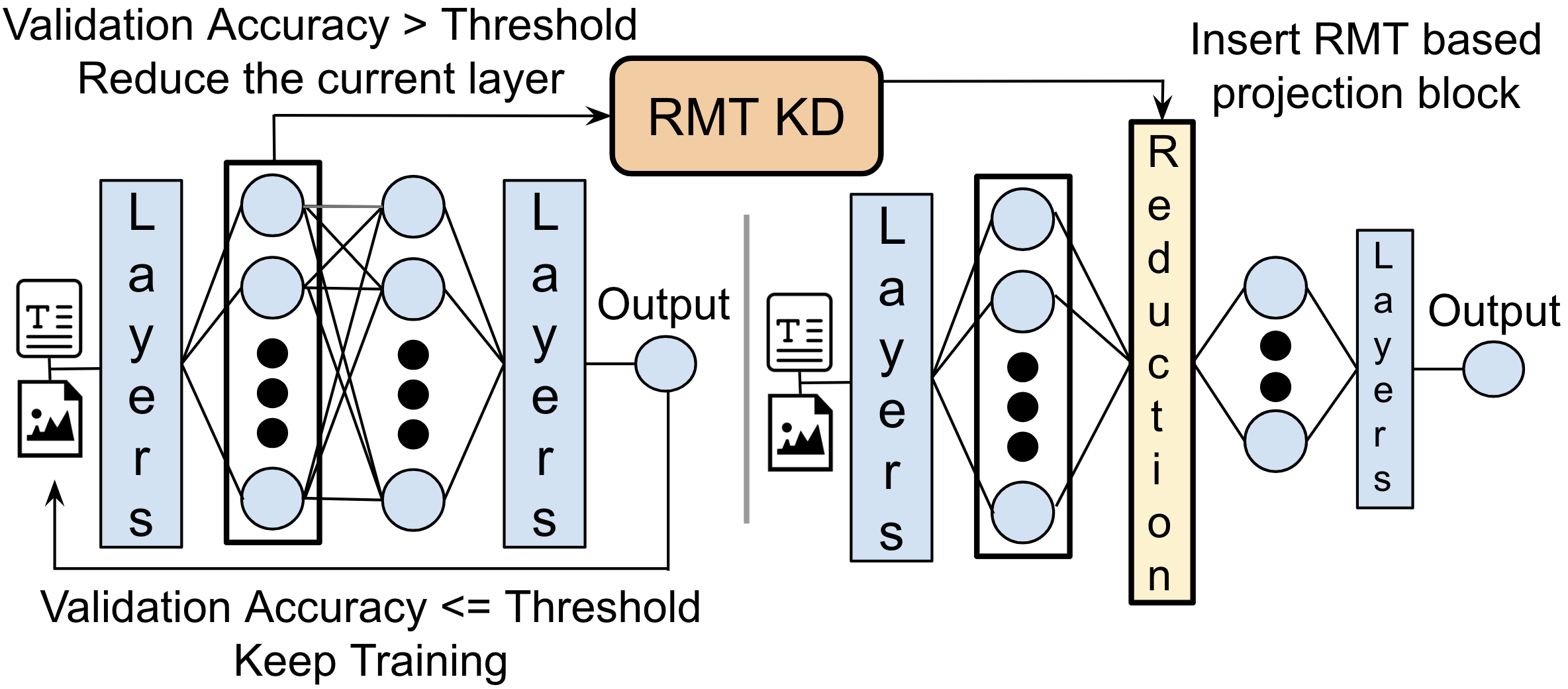}
    \vspace{10pt}
    \caption[Overview of the iterative RMT-KD pipeline.]{Overview of the iterative RMT-KD pipeline: training runs normally until the validation metric exceeds a user-specified threshold; at that moment the selected layer is analyzed with Random Matrix Theory on a held-out calibration subset. The MP bulk edge is estimated to separate noise-like from structure-bearing components. A projection block is inserted that maps activations to the causal subspace spanned by outlier eigenvectors, and downstream widths are resized accordingly. The new (narrower, dense) model is then fine-tuned with a self-distillation objective that aligns its outputs to those of the pre-reduction teacher checkpoint. The same train–analyze–reduce–distill cycle repeats across layers until benefits saturate or targets on parameters, latency, or power are met.}
    \label{fig:rmtkd_architecture}
\end{figure}

\subsection*{Data Collection for Spectral Analysis}

Let a calibration set be formed by sampling a small fraction of the training data that is representative of the task distribution (ten percent by default). Hidden activations from the target layer are collected by a forward pass without gradient updates, forming a column-stacked activation matrix $\mathbf{X} \in \mathbb{R}^{d \times n}$. Here, dimension \( d \) is the layer width and \( n \) is the number of collected samples or tokens. The empirical covariance is then computed $\boldsymbol{\Sigma} = \frac{1}{n}\,\mathbf{X}\mathbf{X}^{\top}$.
For transformer blocks, \(\mathbf{X}\) is taken after the feed-forward or projection sub-block depending on the reduction site; for convolutional networks it is taken channel-wise by flattening spatial dimensions into the sample axis.

\subsection*{Estimating the Noise Bulk via the MP Law}

Under near-isotropic, mean-zero activations typical of normalized deep layers, the eigenvalue distribution of the empirical covariance exhibits a noise bulk captured by the MP density. The aspect ratio is defined as $q = \frac{d}{n}$.
The MP support is parameterized by a noise variance \(\sigma^{2}\) with edges
\begin{align}
    \lambda_{-} \;=\; \sigma^{2}\,(1-\sqrt{q})^{2}
\qquad\text{and}\qquad
\lambda_{+} \;=\; \sigma^{2}\,(1+\sqrt{q})^{2}
\end{align}
RMT-KD initializes \(\sigma^{2}\) from a robust statistic of the spectrum (the median eigenvalue by default), then refines \(\sigma^{2}\) by minimizing the squared error between the empirical histogram and the MP density on the interval \([\lambda_{-},\lambda_{+}]\). The aggressiveness of reduction can be tuned by shifting the initialization quantile upward, which increases \(\lambda_{+}\) and thereby classifies more components as noise.
\begin{align}
    \sigma^{2}_{0} \;=\; \operatorname{Quantile}\!\left(\{\lambda_{i}\}_{i=1}^{d},\,\tau\right)
\quad\text{with}\quad
\tau \in [0.4,0.6]
\end{align}
\begin{align}
\sigma^{2}_{\star} \;=\; \arg\min_{\sigma^{2}} \;\left\|\, \widehat{\rho}(\lambda) \;-\; \rho_{\text{MP}}(\lambda;\sigma^{2},q) \,\right\|_{2}^{2}
\end{align}
\subsection*{Estimating $\sigma^2$ from Eigenvalues of Sample Covariance Matrix}
In random matrix theory, when the data matrix $X \in \mathbb{R}^{N \times p}$ has i.i.d. entries with mean 0 and variance $\sigma^2$, the sample covariance matrix $S = X^T X / N$ has eigenvalues $\lambda_i$ whose empirical mean satisfies $\frac{1}{p} \sum_{i=1}^p \lambda_i = \frac{1}{Np} \sum_{i=1}^N \sum_{j=1}^p X_{ij}^2$. By the Law of Large Numbers, this converges to $\mathbb{E}[X_{ij}^2] = \sigma^2$ (since those are assumed to have 0 mean) as $N, p \to \infty$. Therefore, using the mean eigenvalue to estimate $\sigma^2$ is justified under these ideal conditions. The median eigenvalue also approximates $\sigma^2$, being more robust to outliers in the spikes section. However, this is still an approximation because: (1) finite $N, p$ cause sampling error, (2) neural network activations are not perfectly i.i.d., and (3) the presence of spike eigenvalues from correlated signals biases the mean upward.

\subsection*{Selecting Causal Directions}

\noindent With the refined bulk edge \(\lambda_{+}\) estimated, outliers are identified by thresholding the spectrum.
\begin{align}
    \mathcal{I}_{\text{out}} \;=\; \left\{\, i \in \{1,\ldots,d\} \;:\; \lambda_{i} > \lambda_{+} \,\right\}
\end{align}
Let \(\mathbf{U}_{\text{out}} \in \mathbb{R}^{d \times k}\) denote the eigenvectors corresponding to these \(k=|\mathcal{I}_{\text{out}}|\) outlier eigenvalues. The causal projection is then formed as an orthonormal basis from \(\mathbf{U}_{\text{out}}\) (by construction it is orthonormal) $\mathbf{P} = \mathbf{U}_{\text{out}}^{\top}$. The reduced activation is the image of the original activation through \(\mathbf{P}\), such that $\mathbf{h}^{\prime} = \mathbf{P}\mathbf{h}$.
In code, the projection is implemented as a fixed linear layer initialized with \(\mathbf{P}\) and optionally fine-tuned jointly with the rest of the network. Downstream modules are resized from width \(d\) to width \(k\). For transformers, reductions are applied to token-embedding, intermediate feed-forward, or projection dimensions; for convolutional networks, activations are reshaped so that spatial locations act as samples and channels as features. The resulting activation matrix is projected through the learned matrix $\mathbf{P}$ in feature space and then reshaped back to the original image tensor dimensions, effectively reducing the channel dimension without using explicit convolutional filters.

\subsection*{Self-Distillation for Stability Across Reductions}

\noindent To prevent catastrophic forgetting, each reduction step starts from a teacher checkpoint (pre-reduction) and trains the student (post-reduction) to match both labels and teacher logits. The total loss is a convex combination of task cross-entropy and a Kullback–Leibler alignment term.
\begin{align}
    \mathcal{L} \;=\; \alpha\,\mathcal{L}_{\text{CE}} \;+\; (1-\alpha)\,\operatorname{KL}\!\big(\,\mathbf{p}_{\text{teacher}} \;\|\; \mathbf{p}_{\text{student}}\,\big)
\end{align}
A temperature can be used inside the softmax of both distributions for smoother targets; the teacher is an exponential-moving-average or the last full-precision checkpoint. The coefficient \(\alpha\) is chosen to maintain validation accuracy while allowing the student to adapt to its lower-dimensional subspace.

\subsection*{Progressive Layerwise Schedule}

RMT-KD proceeds over a chosen layer order (shallow-to-deep by default). After each reduction and short fine-tuning, validation is re-checked. The process halts when any of the following happens: target compression or latency is achieved, the retained-outlier ratio falls below a minimum threshold, or validation drops beyond an allowed tolerance.
\begin{align}
    \text{Stop if}\quad
\frac{k}{d} \;<\; \rho_{\min}
\quad\text{or}\quad
\Delta\text{ValAcc} \;<\; -\epsilon
\quad\text{or}\quad
\text{Params} \le \text{Target}
\end{align}
The outlier ratio criterion guards against over-aggressive cuts in deep layers whose spectra carry denser structure.

\subsection*{Computational Complexity}

The spectral step requires covariance formation and an eigen-decomposition. With \(d\) as layer width and \(n\) as calibration size, the costs are
\begin{align}
    \mathcal{O}(n d^{2}) \quad \text{for} \quad \mathbf{X}\mathbf{X}^{\top}
\qquad\text{and}\qquad
\mathcal{O}(d^{3}) \quad \text{for eigen-decomposition.}
\end{align}
Because the calibration set is small and only a few layers are reduced per stage, this overhead is negligible relative to training time. Crucially, the reduced networks remain \emph{dense}: projection uses compact matrix–vector multiplies that map efficiently to standard GPU kernels; no sparse backends or custom kernels are required. On-device memory decreases with width, lowering activation and weight footprints; data movement and compute scale down accordingly, yielding both latency improvements and power savings.

\subsection*{Implementation Notes}

To make the procedure robust in practice, we adopt the following defaults, which were effective across transformers and ResNets:
\begin{align}
    \tau \in [0.4, 0.5] \quad\text{for initializing }\sigma^{2}, 
\qquad
\alpha \in [0.3, 0.7], 
\qquad
\text{temperature} \in [0.5, 1.5]
\end{align}
\begin{align}
    \text{Calibration fraction} \approx 0.1, 
\qquad 
\text{MP fit via }\ell_{2}\text{ histogram matching on }[\lambda_{-}, \lambda_{+}]
\end{align}
For BERT-like models, reductions at the feed-forward inner width and projection dimensions bring the largest gains; for ResNet-50, activations are flattened so that spatial positions serve as samples and channels as features, followed by RMT-based projection in feature space and reshaping back to the original tensor. After each projection insertion, a brief warm-up phase can be applied to stabilizes training before resuming full fine-tuning.

\subsection*{Example Pseudocode of RMT-KD Process}
\vspace{-25pt}
\[
\begin{array}{l}
\textbf{Input:} \ \text{trained checkpoint } \mathcal{M}, \ \text{target layer index } \ell, \ \text{calibration set } \mathcal{D}_{\text{cal}} \\
\textbf{1.} \ \text{Collect activations } \mathbf{X} \text{ on } \mathcal{D}_{\text{cal}} \\
\textbf{2.} \ \boldsymbol{\Sigma} \leftarrow \frac{1}{n}\mathbf{X}\mathbf{X}^{\top} \\
\textbf{3.} \ \text{Initialize } \sigma^{2}_{0} \text{ from spectrum quantile } \tau \\
\textbf{4.} \ \sigma^{2}_{\star} \leftarrow \arg\min_{\sigma^{2}} \|\widehat{\rho} - \rho_{\text{MP}}(\sigma^{2}, q)\|_{2}^{2} \\
\textbf{5.} \ \lambda_{+} \leftarrow \sigma^{2}_{\star}(1+\sqrt{q})^2 \\
\textbf{6.} \ \text{Select outliers } \{\lambda_{i} > \lambda_{+}\} \text{ and eigenvectors } \mathbf{U}_{\text{out}} \\
\textbf{7.} \ \mathbf{P} \leftarrow \mathbf{U}_{\text{out}}^{\top}, \ \ \mathbf{h}^{\prime} \leftarrow \mathbf{P}\mathbf{h} \\
\textbf{8.} \ \text{Insert projection block; resize downstream widths to } k \\
\textbf{9.} \ \text{Fine-tune with } \mathcal{L} = \alpha \mathcal{L}_{\text{CE}} + (1-\alpha)\operatorname{KL}(\mathbf{p}_{\text{teacher}}\|\mathbf{p}_{\text{student}}) \\
\textbf{10.} \ \text{Restart for next layer or stop}
\end{array}
\]

\begin{figure}[h]
    \centering
    \includegraphics[width=\linewidth]{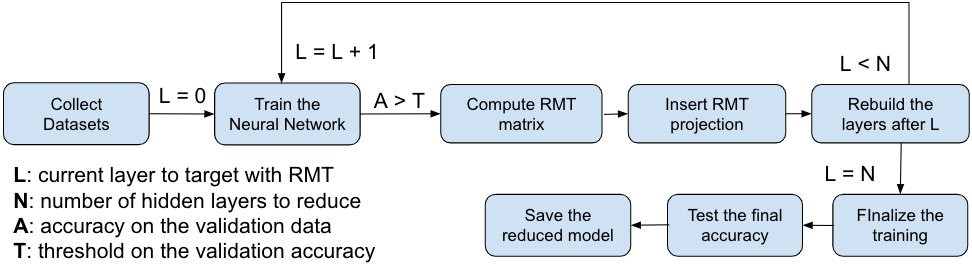}
    \vspace{20pt}
    \caption[Iterative RMT-KD training process]{Iterative RMT-KD training process: the model is trained until the validation accuracy surpasses a threshold, after which Random Matrix Theory is applied layer by layer to compute eigenvalue spectra, insert causal projections, and rebuild the network with reduced dimensions. The cycle continues until all target layers are processed and the final compressed model is validated and saved.}
    \label{fig:training_process}
\end{figure}

\subsection*{Why This Design Is Sensible}

The MP bulk fitting provides a statistically grounded noise floor, eliminating ad hoc cutoffs that plague PCA-style truncation and heuristic rank selection. Selecting directions by outlier tests aligns the model with a spiked covariance picture in which only a small set of eigen-directions carry task structure. Performing projection as a dense linear map preserves hardware efficiency. Finally, self-distillation smooths the optimization landscape across discrete architecture changes, ensuring that compressed models inherit the function learned by their predecessors rather than relearning from scratch.

%% file: sections/4_methodology/rmtkd_2.tex
\section{Theoretical Justification}
\label{methodology:rmtkd_theoretical_justification}

\subsection*{Random Matrix Theory as a Statistical Lens}

Random Matrix Theory (RMT) provides a rigorous mathematical framework to describe the spectral behavior of large matrices whose entries are random or exhibit weak correlations. 
When applied to deep learning, RMT enables the separation of meaningful, task-relevant structures in neural activations from random fluctuations that arise during training.
In high-dimensional regimes, RMT offers asymptotic laws that govern the eigenvalue spectra of covariance matrices.

Consider a hidden activation matrix $X \in \mathbb{R}^{d \times n}$ formed by the activations of $d$ neurons across $n$ samples. 
The empirical covariance matrix is defined as $\Sigma = \frac{1}{n} XX^\top$ whose eigenvalues $\{ \lambda_i \}_{i=1}^d$ encode the variance distribution of the learned representations.
RMT predicts that for large $d, n$, the eigenvalue density of such random covariance matrices converges to the Marchenko–Pastur (MP) distribution \cite{MarchenkoPastur1967}, given by $\rho_{\text{MP}}(\lambda)=\frac{1}{2\pi\lambda q \sigma^2}\sqrt{(\lambda_+-\lambda)(\lambda-\lambda_-)}\,,\ \lambda\in[\lambda_-,\lambda_+]$ where $q = d/n$, $\sigma^2$ is the variance of the entries of $X$, and the bulk edges are defined as $\lambda_{\pm}=\sigma^2(1\pm\sqrt{q})^2$. Eigenvalues within the interval $[\lambda_-, \lambda_+]$ form the \emph{bulk}, corresponding to random noise.
In contrast, eigenvalues $\lambda_i > \lambda_+$ represent statistically significant deviations, so-called outliers, which capture structured or causal information within the activations.
This separation between noise and structure provides a statistically grounded rule for identifying informative directions in representation space.

\subsection*{BBP Phase Transition}

The theoretical foundation for distinguishing signal from noise lies in the spiked covariance model \cite{johnstone2001distribution}.
In this model, the covariance matrix is assumed to consist of a low-rank structured component plus isotropic noise: $\Sigma = \Sigma_s + \sigma^2 I_d$
where $\Sigma_s$ encodes the task-relevant directions (the ``spikes'').
When the signal strength exceeds a critical threshold, that is known as the Baik–Ben Arous–Péché (BBP) transition \cite{baik2005phase}, the corresponding eigenvalues detach from the MP bulk.
The BBP threshold is defined as:
\begin{align}
    \lambda_{\text{BBP}} = \sigma^2 (1 + \sqrt{c}), \quad \text{where } c = \frac{d}{n}.
\end{align}
Eigenvalues $\lambda_i > \lambda_{\text{BBP}}$ correspond to genuine signals whose eigenvectors align with semantically meaningful or causal dimensions of the data distribution.
This mechanism explains how deep representations evolve from random initializations to structured manifolds as learning progresses: early in training, the spectrum follows the MP law closely, whereas later stages exhibit prominent outliers signifying emergent structure.

\subsection*{Application to Neural Representations}

Deep neural networks, particularly large language models (LLMs) and vision models (CNNs, VLMs), produce hidden representations that are extremely high-dimensional but redundantly parameterized.
Empirically, their covariance spectra display a characteristic profile: a large noisy bulk and a small number of outliers that dominate the representational variance \cite{martin2021implicit}.
RMT thus provides a principled criterion for model compression and subspace selection, in contrast to heuristic thresholds used in principal component analysis (PCA) or low-rank approximations.

For a given layer, once the activation covariance $\Sigma$ is computed from a calibration subset, the noise variance $\sigma^2$ is initialized as the median eigenvalue and refined by minimizing the $\ell_2$ distance between the empirical histogram and the MP distribution. 
The resulting $\lambda_+$ acts as a dynamic cutoff separating signal from noise.
All eigenvectors associated with $\lambda_i > \lambda_+$ define a projection matrix $P \in \mathbb{R}^{k \times d}$, where $k < d$, mapping the activations onto a lower-dimensional causal subspace: $Y = P X$.
This transformation eliminates noisy or redundant components while retaining the statistically validated causal structure.
The process is repeated layer by layer, yielding compact, dense networks whose internal representations are aligned with the intrinsic structure of the data.

\subsection*{Empirical Evidence and Spectral Interpretation}

\begin{figure}[h]
    \centering
    \includegraphics[width=\textwidth]{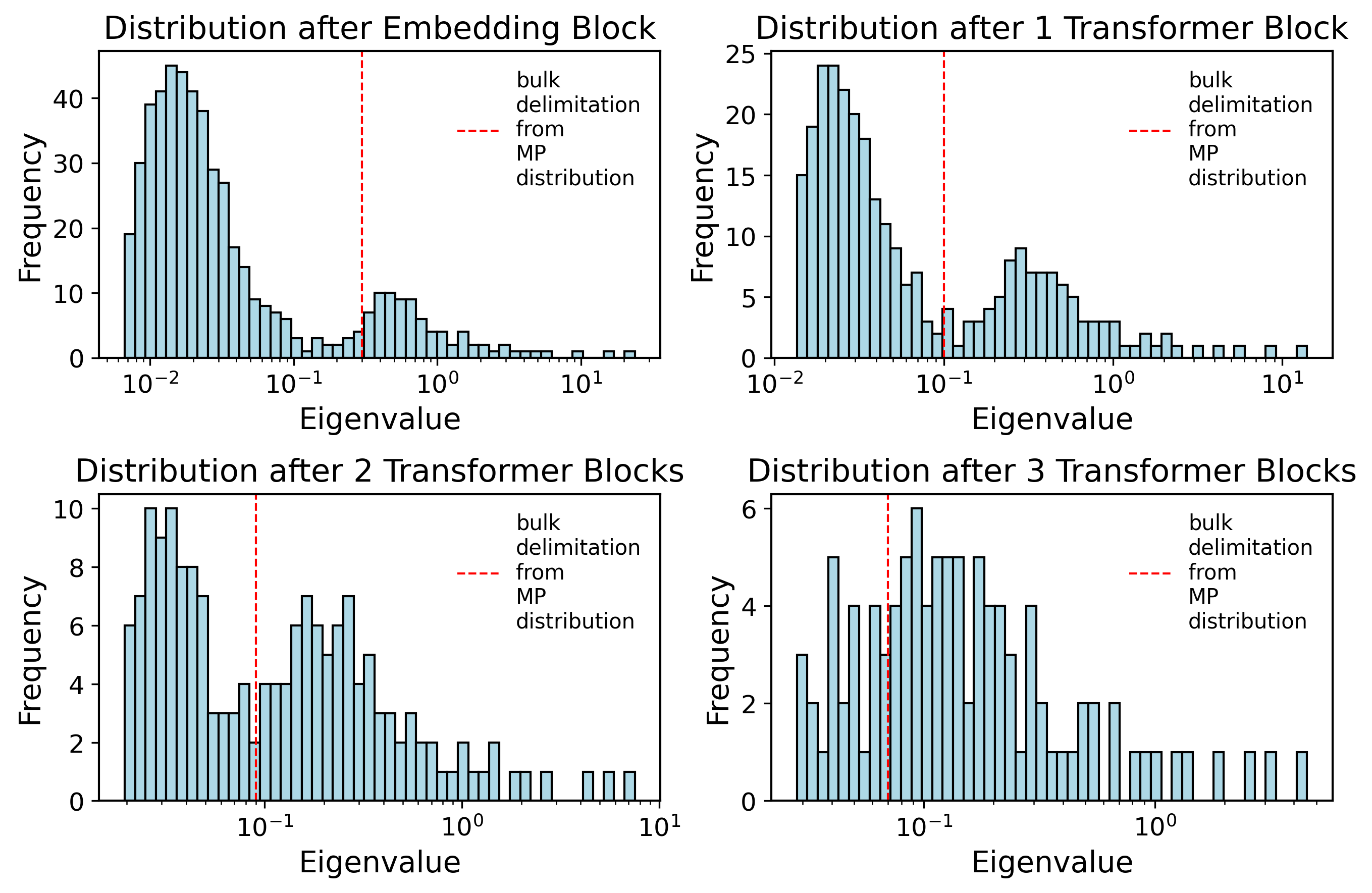}
    \vspace{25pt}
    \caption[Evolution of the empirical eigenvalue distribution]{Evolution of the empirical eigenvalue distribution: activation covariances across BERT-base layers, trained on SST dataset (from GLUE data). The red dashed line denotes the upper bulk edge $\lambda_+$ predicted by the Marchenko–Pastur law. The emergence of outlier eigenvalues across layers indicates the progressive formation of structured, causal representations.}
    \label{fig:eig_distribution}
\end{figure}

Figure \ref{fig:eig_distribution} visualizes the empirical eigenvalue spectra of hidden layer activations in BERT-base at increasing depths.
The red dashed line marks the upper edge $\lambda_+$ predicted by the Marchenko–Pastur law.
In the embedding block (top left), most eigenvalues cluster near zero, with few extending beyond $\lambda_+$, indicative of unstructured, noise-dominated representations.
As training proceeds through successive Transformer blocks, more eigenvalues detach from the bulk, forming distinct outliers that capture meaningful linguistic or contextual information.

This spectral evolution demonstrates that deep layers encode increasingly specialized, semantically rich features, while random-like fluctuations dominate earlier layers.
Hence, RMT-guided filtering provides a natural boundary for compression: early layers can be aggressively reduced without losing meaningful structure, while deeper layers require conservative compression to preserve critical information.

\subsection*{Spectral Geometry for Reliability}

Beyond its role in compression, the spectral analysis of hidden activations connects to broader principles of spectral geometry and information stability.
Large outlier eigenvalues correspond to directions of high curvature or concentrated information flow, whereas the bulk reflects isotropic noise and instability.

This geometric viewpoint suggests that a model’s reliability, its robustness against hallucination or out-of-distribution drift, can also be inferred from spectral signatures.
A stable, well-calibrated model maintains a consistent separation between bulk and outlier regions, indicating a balanced representation of structure and noise.
Conversely, the collapse or inflation of eigenvalue spectra may signal overfitting, undertraining, or instability under perturbations.

Thus, the same spectral framework that governs compression in RMT-KD also underpins diagnostic tools like EigenTrack, establishing a unifying theoretical basis across efficiency and reliability dimensions. In summary, RMT provides:
\begin{itemize}
    \item A statistically grounded threshold ($\lambda_+$ or $\lambda_{\text{BBP}}$) for separating informative structure from random noise.
    \item A causal interpretation of eigenvalue outliers as stable, meaningful directions in representation space.
    \item A framework for constructing low-dimensional, dense, and interpretable subspaces without ad hoc heuristics.
\end{itemize}
By embedding this spectral analysis into an iterative self-distillation loop, RMT-KD transforms deep networks into efficient, causally consistent systems, reducing redundancy while preserving semantic power.

%% file: sections/4_methodology/rmtkd_3.tex
\section{Experimental Setup}
\label{methodology:rmtkd_experimental_setup}

\subsection*{Overview}
To evaluate the effectiveness of the proposed RMT-KD framework, we conduct a series of controlled experiments across both language and vision domains. The goal of this setup is to assess whether Random Matrix Theory (RMT)-guided layer compression can significantly reduce model size and computational cost while preserving accuracy and representational quality. All experiments are implemented in \texttt{PyTorch} using CUDA acceleration, ensuring full reproducibility and compatibility with common training pipelines. 

\subsection*{Model Architectures}

We evaluate RMT-KD on two representative classes of deep networks: Transformers for language modeling and Convolutional Neural Networks (CNNs) for image recognition. This dual evaluation demonstrates the generality of the proposed framework across architectures that differ in structural topology, activation statistics, and spectral properties.

\textbf{BERT-base.}  
We adopt a 12-layer Transformer encoder identical to the BERT-base configuration \cite{devlin2019bert}, comprising 12 self-attention heads per layer, hidden size $d=768$, and feed-forward dimension $d_{ff}=3072$. The model totals approximately 139 million parameters. The training uses WordPiece tokenization with a vocabulary of 30k tokens. The base model serves as the initial teacher in the RMT-KD iterative distillation loop, from which progressively reduced students are generated by projecting activation subspaces based on spectral analysis.

\textbf{BERT-tiny.}  
As a smaller baseline, we also include the 6-layer TinyBERT variant with 44 million parameters. This model has a reduced hidden dimension of $d=384$ and 6 attention heads per layer. Evaluating RMT-KD on BERT-tiny provides insight into the limits of compression in already compact architectures, where redundancy is minimal.

\textbf{ResNet-50.}  
For computer vision experiments, we use the standard 50-layer residual network \cite{he2016resnet} consisting of bottleneck convolutional blocks and batch normalization layers. The architecture contains 23.5 million parameters and achieves high baseline accuracy on CIFAR-10. Applying RMT-KD to this convolutional backbone illustrates the adaptability of the spectral compression principle beyond attention-based architectures.

All models are initialized from scratch using Xavier initialization and trained end-to-end before any compression step. This ensures that the activation covariances analyzed during RMT calibration reflect converged, semantically meaningful features.

\noindent We evaluate performance on three datasets representing complementary domains:

\textbf{GLUE Benchmark.}  
The General Language Understanding Evaluation (GLUE) suite includes multiple text classification and sentence-pair tasks such as SST-2 (sentiment analysis), QNLI (question–answer entailment), and QQP (duplicate question detection). Each subtask uses standard train/validation/test splits. Text sequences are tokenized to a maximum length of 128 tokens and padded to form uniform mini-batches. Training and evaluation follow the official GLUE evaluation protocol, with metrics such as accuracy and F1 score.

\textbf{AG News.}  
This dataset contains 120k news articles categorized into four classes (World, Sports, Business, Sci/Tech). It is used to test the scalability of RMT-KD to large text corpora. Documents are preprocessed using lowercasing and punctuation normalization, and truncated to 256 tokens.

\textbf{CIFAR-10.}  
For vision tasks, we use the CIFAR-10 dataset containing 50k training and 10k test images across ten object categories. Each image has a resolution of $32 \times 32$ pixels. Standard augmentations including random horizontal flip, random crop with padding of 4 pixels, and per-channel normalization are applied. These augmentations ensure that the covariance statistics estimated by RMT reflect robust and diverse activations.

\subsection*{Training Procedure}

All experiments are conducted on a NVIDIA RTX 6000 GPU with 48 GB of memory. GPU power consumption and energy efficiency are measured using NVIDIA’s System Management Interface (\texttt{nvidia-smi}) at 1-second resolution to estimate average draw during inference. Training follows a two-phase procedure: \emph{(i)} baseline model training and \emph{(ii)} RMT-guided compression with self-distillation.

\textbf{Phase I – Baseline Training.}  
Each baseline model is trained using the AdamW optimizer with initial learning rate $\eta_0 = 1 \times 10^{-4}$ for Transformers and $\eta_0 = 1 \times 10^{-3}$ for CNNs. The learning rate decays linearly during the epochs by a factor of $0.1$. We use weight decay $\lambda = 0.0005$ and dropout $p = 0.1$. Batch sizes are set to 32 for language tasks and 128 for vision tasks, while the numeber of epochs range from 5 to 10. Training continues until validation accuracy saturates or exceeds a pre-defined threshold $\tau_{val}$ (typically 90\% of the expected baseline score).

\textbf{Phase II – RMT-KD Compression.}  
Once convergence is reached, a calibration subset $\mathcal{D}_{\text{cal}}$ is sampled, comprising 10\% of the training data. Hidden activations $X \in \mathbb{R}^{d \times n}$ from each target layer are extracted to compute empirical covariance matrices $\Sigma = \frac{1}{n}XX^\top$. The eigenvalue spectrum $\{\lambda_i\}$ of $\Sigma$ is estimated using efficient eigendecomposition routines in \texttt{NumPy}. The empirical distribution is compared to the Marchenko–Pastur (MP) law, and the noise variance $\sigma^2$ is initialized as the median of the eigenvalues (or any other chosen quantile as we'll see in the ablation studies) and then refined by minimizing the $\ell_2$ distance between the histogram of $\{\lambda_i\}$ and the expected MP density. Eigenvalues $\lambda_i > \lambda_+$, where $\lambda_+$ is the upper edge of the MP support, are selected as \emph{causal directions}. Their corresponding eigenvectors form a projection matrix $P \in \mathbb{R}^{k \times d}$, which maps activations to the reduced subspace $\mathbb{R}^k$.

This projection is implemented as a fixed linear layer, inserted between the original layers of the model. After projection, the model undergoes fine-tuning with a self-distillation loss that enforces alignment between the original logits $p_{\text{old}}$ and the new logits $p_{\text{new}}$:
\begin{align}
    \mathcal{L} = \alpha \, \mathcal{L}_{\text{task}} + (1 - \alpha) \, D_{\mathrm{KL}}(p_{\text{old}} \| p_{\text{new}})
\end{align}
where $\alpha = 0.7$ balances the supervised task loss and the distillation regularizer. The procedure is repeated iteratively across multiple layers, halting when the target compression ratio or accuracy threshold is reached.

\subsection*{Evaluation Metrics}

We assess each model on three dimensions:
\begin{itemize}
    \item \textbf{Predictive performance:} classification accuracy and F1 score on the test sets.
    \item \textbf{Computational efficiency:} wall-clock inference time and throughput (samples/sec).
    \item \textbf{Resource usage:} model parameter count, disk size, and GPU power consumption.
\end{itemize}

Inference latency is measured with batch size 1 to emulate real-time applications, averaged over 1000 forward passes. Energy measurements integrate instantaneous power over the full inference window.

All training scripts are implemented in modular PyTorch classes to ensure reusability. Random seeds are fixed for all experiments to ensure determinism. Model checkpoints, spectra, and projection matrices are saved using \texttt{dill} for transparent tracking of compression stages.

%% file: sections/4_methodology/rmtkd_4.tex
\section{Results}
\label{methodology:rmtkd_results}

\subsection*{Overview}
The effectiveness of the proposed RMT-KD framework is assessed across diverse model families and datasets to verify its ability to jointly enhance efficiency and preserve performance. The experiments focus on three representative settings: BERT-base and BERT-tiny trained on the GLUE benchmark tasks (SST, QQP, and QNLI), and ResNet-50 on CIFAR-10. The analysis includes a comprehensive evaluation of accuracy, compression ratio, inference speed, energy and power consumption, as well as memory footprint. 

All models are fine-tuned under identical optimization and data conditions. The RMT-KD variants differ only in their use of spectral projections derived from the eigenvalue structure of hidden activations. This ensures that any observed variation in behavior can be attributed to the Random Matrix–based distillation itself, rather than to external hyperparameter tuning or architectural modifications. The following subsections present quantitative results and interpretive commentary for each major aspect of evaluation.

\subsection*{Accuracy and Parameter Reduction}
\begin{figure}[h]
    \centering
    \includegraphics[width=\textwidth]{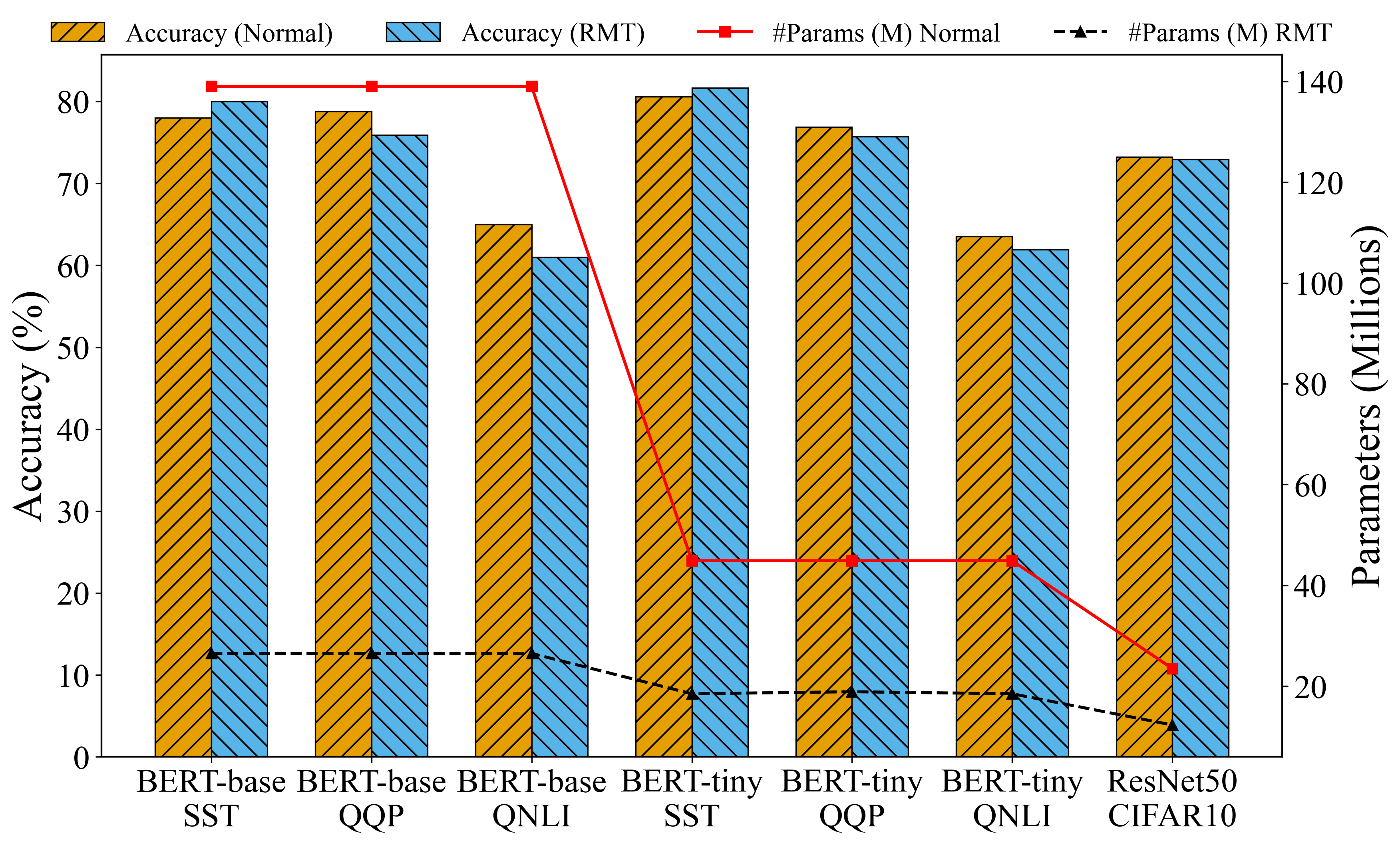}
    \vspace{10pt}
    \caption[Accuracy and parameter reduction across tasks]{Accuracy and parameter reduction across tasks: accuracy comparison between baseline and RMT-KD models (bars, left axis) together with parameter count reduction (lines, right axis). RMT-KD attains up to 80.9\% parameter reduction with negligible or positive accuracy variations.}
    \label{fig:rmtkd_accuracy_parameters}
\end{figure}

Figure~\ref{fig:rmtkd_accuracy_parameters} summarizes the fundamental performance–compression trade-off achieved by RMT-KD. Across all datasets, the Random Matrix–guided projection significantly reduces the number of parameters while maintaining, and in several cases slightly improving, classification accuracy. For BERT-base, parameter counts fall by approximately 81\%, yet the accuracy on SST and QQP increases modestly. This improvement arises from the elimination of noisy and redundant directions in the representation space, allowing the model to operate on a more structured, low-dimensional manifold that retains only causally relevant components. 

For the smaller BERT-tiny, compression remains substantial (around 58\%) but accuracy drops minimally, staying within one percentage point of the baseline. This indicates that even compact models harbor spectral redundancy, and that RMT-KD successfully extracts and preserves the dominant signal modes. The behavior of ResNet-50 on CIFAR-10 further confirms the generality of this principle: despite being convolutional and non-transformer-based, its spectral structure follows the same pattern, leading to almost half the parameters being pruned with negligible performance loss. 

These results demonstrate that the eigenvalue-based selection criterion acts as a robust proxy for representational importance. The alignment between theoretical prediction and empirical observation reinforces the claim that the Marchenko–Pastur bulk effectively characterizes noise-like fluctuations, while the eigenvalue outliers capture task-relevant structure.

\subsection*{Inference Speed and Power Consumption}
\begin{figure}[h]
    \centering
    \includegraphics[width=\textwidth]{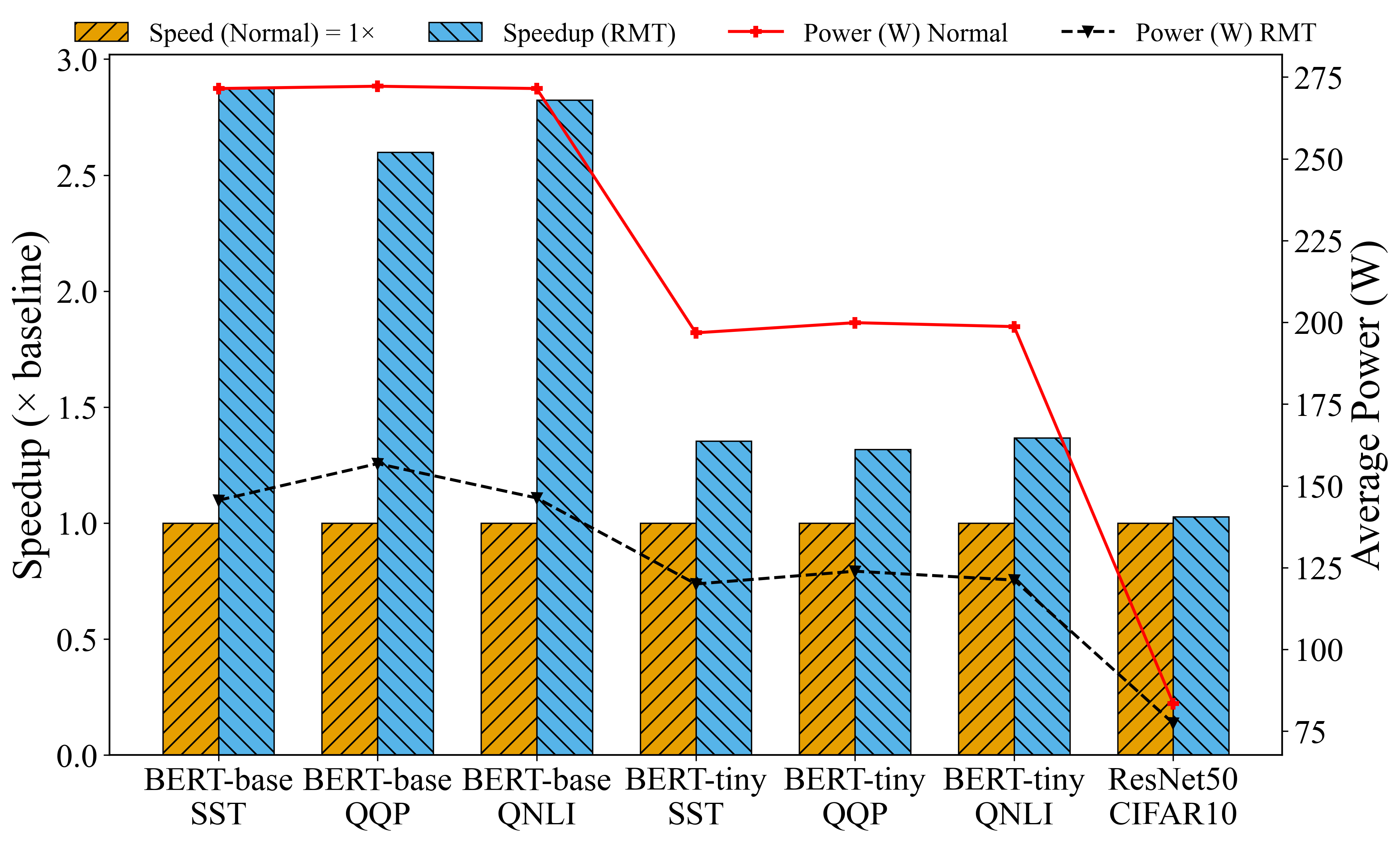}
    \vspace{10pt}
    \caption[Inference speedup and power reduction]{Inference speedup and power reduction: relative inference speedup (bars, left axis) and power consumption (lines, right axis) of RMT-KD models compared with baseline implementations. Values above the bars indicate multiplicative speed gains with respect to the uncompressed model.}
    \label{fig:rmtkd_speed_power}
\end{figure}

The improvements in computational efficiency are illustrated in Figure~\ref{fig:rmtkd_speed_power}. The RMT-KD models consistently achieve faster inference across all configurations, confirming that the spectral projections yield measurable hardware-level benefits. On BERT-base, inference throughput increases nearly threefold on SST and QNLI, reaching speedups of 2.88× and 2.82× respectively. This gain originates from the substantial dimensionality reduction of intermediate representations, which reduces the cost of matrix multiplications in both the attention and feedforward blocks. 

Power measurements show a correlated reduction, with average consumption during inference decreasing by nearly half. The overall power profile becomes smoother, with fewer peaks corresponding to heavy tensor operations. This behavior reflects a denser yet lighter model whose computation remains contiguous and well-aligned with GPU acceleration patterns. The smaller BERT-tiny variants, while starting from a more efficient baseline, still benefit from a consistent 30–35\% improvement in runtime and reduced wattage, demonstrating that the method scales proportionally across model sizes. Even the convolutional ResNet-50 sees a slight gain, evidencing that RMT-KD can complement conventional CNN compression methods without architectural redesign. 

From a systems standpoint, these results indicate that spectral compression transforms the efficiency landscape of transformer models. By maintaining dense matrices while shrinking their dimensions according to principled statistical rules, RMT-KD avoids the memory access inefficiencies typical of sparse pruning and quantization. The resulting models exhibit both algorithmic and energy-level optimization, suitable for large-scale deployment.

\subsection*{Memory and Energy Efficiency}
\begin{figure}[h]
    \centering
    \includegraphics[width=\textwidth]{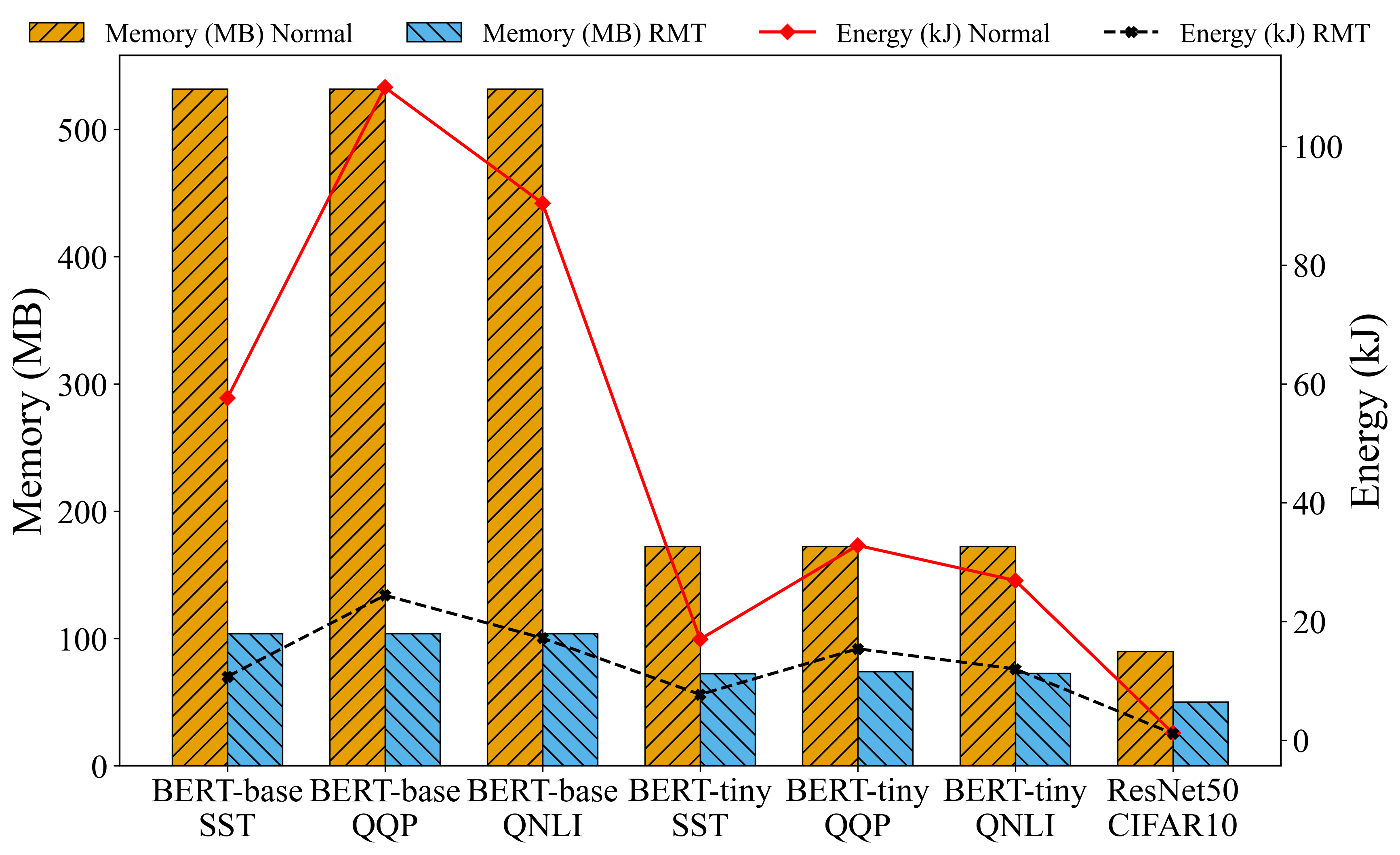}
    \vspace{10pt}
    \caption[Memory footprint and energy efficiency]{Memory footprint and energy efficiency: comparison of memory footprint (bars, left axis) and total energy consumption per inference (lines, right axis) for baseline and RMT-KD models. Energy values represent accumulated consumption over entire forward passes.}
    \label{fig:rmtkd_memory_energy}
\end{figure}

Figure~\ref{fig:rmtkd_memory_energy} highlights the impact of RMT-KD on memory usage and energy expenditure. The compression achieved in the parameter domain directly translates into reduced storage and runtime memory. For the BERT-base family, the total memory footprint drops by roughly 81\%, shrinking from over 500\,MB to near 100\,MB, while maintaining functional equivalence in downstream predictions. The energy cost per inference, measured over full sequences, decreases proportionally, indicating that the reduction in model size and power draw compounds multiplicatively to yield significant energy savings. 

The smaller BERT-tiny models exhibit around 55\% memory reduction and roughly 50\% lower energy requirements, a nontrivial improvement given their already compact nature. In the convolutional setting, ResNet-50 benefits less dramatically due to its structured and spatially constrained design, yet still achieves approximately 10\% energy reduction. Importantly, these gains are obtained without introducing sparsity or hardware-specific optimization; the compressed models remain fully dense, enabling efficient parallel execution on GPUs and TPUs. 

The combined evidence across the three plots underscores the dual virtue of the RMT-KD approach: a substantial acceleration of inference and reduction of computational cost, accompanied by preserved or improved predictive reliability. The compression operates not as a heuristic simplification but as a mathematically grounded projection derived from the spectral geometry of the network itself. By identifying and preserving the eigenmodes that carry causal structure while discarding noise-like components, the method simultaneously enhances both efficiency and interpretability.

The results presented above confirm the central hypothesis of this thesis: Random Matrix Theory provides a rigorous and practical foundation for understanding and optimizing deep neural networks. The consistent correlation between eigenvalue outlier preservation and empirical performance demonstrates that the spectral geometry of neural activations is a reliable indicator of representational content. Through the RMT-KD process, models are distilled not by arbitrary pruning or quantization rules, but through a data-driven, theoretically justified projection that aligns with statistical mechanics principles. The gains reported in Figures~\ref{fig:rmtkd_accuracy_parameters}–\ref{fig:rmtkd_memory_energy} collectively show that spectral methods can reconcile efficiency and reliability, yielding models that are simultaneously faster and smaller.

\begin{table*}[h]
\centering
\caption[PERFORMANCE COMPARISON OF RMT-KD.]
{PERFORMANCE COMPARISON OF RMT-KD}
\vspace{10pt}
\small
\setlength{\tabcolsep}{15pt}   
\renewcommand{\arraystretch}{1.8}  
\begin{tabular}{lcccccc}
\toprule
& \multicolumn{2}{c}{\textbf{BERT-base}} 
& \multicolumn{2}{c}{\textbf{BERT-tiny}} 
& \multicolumn{2}{c}{\textbf{ResNet-50}} \\
\cmidrule(lr){2-3}\cmidrule(lr){4-5}\cmidrule(lr){6-7}
\textbf{Method} 
& \textbf{Red.} & \textbf{Acc.} 
& \textbf{Red.} & \textbf{Acc.} 
& \textbf{Red.} & \textbf{Acc.} \\
\midrule
\textbf{RMT-KD} 
& \textbf{80.9\%} & \textbf{+1.8\%} 
& \textbf{58.8\%} & \textbf{+1.4\%} 
& \textbf{47.7\%} & \textbf{+0.7\%} \\
DistilBERT 
& 42.7\% & +0.2\% 
& \underline{54.8\%} & \underline{+0.4\%} 
& \multicolumn{2}{c}{\textemdash} \\
Theseus 
& \underline{48.3\%} & \underline{+0.6\%} 
& 53.0\% & +0.1\% 
& \multicolumn{2}{c}{\textemdash} \\
PKD 
& 40.5\% & -1.0\% 
& 50.1\% & -0.8\% 
& \multicolumn{2}{c}{\textemdash} \\
AT 
& \multicolumn{2}{c}{\textemdash} 
& \multicolumn{2}{c}{\textemdash} 
& 42.2\% & +0.4\% \\
FitNet 
& \multicolumn{2}{c}{\textemdash} 
& \multicolumn{2}{c}{\textemdash} 
& 40.6\% & +0.2\% \\
CRD 
& \multicolumn{2}{c}{\textemdash} 
& \multicolumn{2}{c}{\textemdash} 
& \underline{45.4\%} & \underline{+0.6\%} \\
\bottomrule
\end{tabular}
\label{comparison_table}
\end{table*}

\noindent
~\ref{comparison_table} shows the BERT results evaluated on the GLUE benchmark, and ResNet-50 results reported on CIFAR-10. 
Theseus = BERT-of-Theseus, PKD = Patient Knowledge Distillation, AT = Attention Transfer, FitNet = FitNets (Hints for Thin Deep Nets), and CRD = Contrastive Representation Distillation. 
Missing entries (—) indicate that NLP and CV baselines are evaluated separately. It summarizes the comparative performance of RMT-KD against several state-of-the-art distillation and compression baselines across both natural language processing (NLP) and computer vision (CV) domains. 
For NLP tasks on the GLUE benchmark, RMT-KD achieves remarkable compression on BERT-base, reducing parameters by over 80\% while simultaneously improving accuracy by +1.8\% compared to the original model. 
Even the lighter BERT-tiny variant benefits substantially, with nearly 60\% reduction and a modest accuracy improvement of +1.4\%. 
In the vision domain, ResNet-50 achieves a 47.7\% reduction with a consistent accuracy gain of +0.7\%, indicating that RMT-KD generalizes effectively across modalities. Compared to popular baselines such as DistilBERT and Theseus, RMT-KD consistently provides larger compression ratios while also improving or maintaining accuracy. 
For instance, while DistilBERT compresses BERT-base by roughly 43\%, its performance improvement is limited to +0.2\%, suggesting that conventional distillation retains redundant subspaces. 
By contrast, RMT-KD identifies and preserves only the causal spectral components, those corresponding to outlier eigenvalues beyond the Marchenko–Pastur threshold, ensuring that only statistically meaningful directions are maintained. 
This theoretically grounded approach enables aggressive compression without the instability or heuristic tuning typical of pruning-based and low-rank factorization methods.

On computer vision benchmarks, RMT-KD also surpasses advanced distillation strategies such as Contrastive Representation Distillation (CRD) and Attention Transfer (AT), achieving both higher compression and slightly better accuracy. 
The consistent positive deltas across all evaluated settings emphasize the robustness of the proposed random-matrix-guided projection and its ability to produce dense, hardware-efficient student models. 
Overall, ~\ref{comparison_table} highlights how RMT-KD bridges the gap between mathematical rigor and practical compression, establishing a new balance between efficiency and performance in deep neural networks.

%% file: sections/4_methodology/rmtkd_5.tex
\section{Ablation Studies}
\label{methodology:rmtkd_ablation}

\subsection*{Quantile Sensitivity Trade-off}

A crucial hyperparameter in RMT-KD is the initialization quantile used to estimate the noise variance $\sigma^2$ in the Marchenko–Pastur (MP) distribution. This quantile determines the upper bulk edge $\lambda_{+}$ and, consequently, the cutoff threshold separating random from causal eigenvalues. Adjusting this quantile directly controls the aggressiveness of compression: higher quantiles enlarge $\lambda_{+}$, classifying more eigenvalues as noise and discarding more dimensions, while lower quantiles retain a broader subspace and yield more conservative reductions.

\begin{figure}[h]
    \centering
    \includegraphics[width=0.95\linewidth]{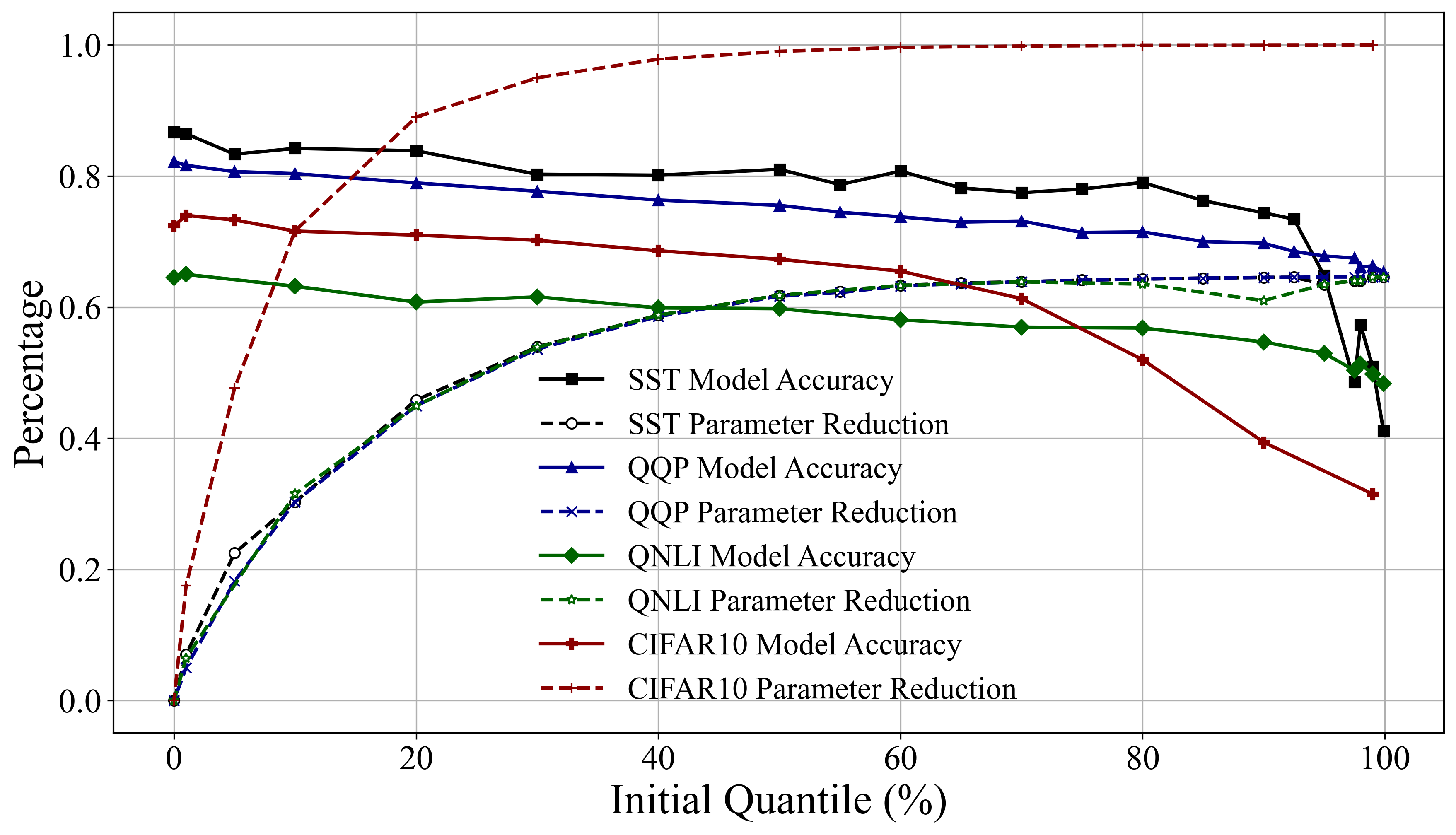}
    \vspace{20pt}
    \caption[Accuracy–reduction tradeoff for RMT-KD.]{Accuracy–reduction tradeoff for RMT-KD: Accuracy–reduction tradeoff as a function of the initial quantile used to initialize $\sigma^2$. The shaded region (40–50\% quantile) marks the regime providing optimal balance between accuracy retention and compression across tasks. Experiments are conducted with BERT-base model on all datasets.}
    \label{fig:quantile_comparison}
\end{figure}

To systematically study the influence of this parameter, we conducted an ablation experiment across a wide range of quantiles, from 0\% to 100\%, on both natural language and vision benchmarks. For each quantile, the complete RMT-KD pipeline was executed, involving iterative self-distillation, RMT-guided projection, and fine-tuning. ~\ref{fig:quantile_comparison} reports model accuracy and parameter reduction percentages for BERT-base on the GLUE datasets (SST, QQP, QNLI) and ResNet-50 on CIFAR-10.

\subsection*{Observations Across Quantiles}

At low quantiles (below 20\%), the estimated $\sigma^2$ is small, leading to narrow MP bulk edges and the preservation of nearly all eigenvalues. In this regime, compression is minimal, parameter reduction remains below 30\%, but task accuracy remains essentially identical to the baseline, confirming that RMT-KD behaves conservatively when the noise threshold is underestimated.

As the quantile increases toward the median (40–50\%), a marked transition occurs. The empirical bulk edges $\lambda_{\pm}$ now align closely with the statistical envelope predicted by the spiked covariance model, effectively filtering out noisy directions while retaining meaningful, structured components of the feature space. Across all datasets, this median regime yields the best compromise between compactness and fidelity: parameter reduction reaches 60–80\% while accuracy degradation remains within 1–2 percentage points. Interestingly, in some cases (notably SST and QNLI), accuracy slightly improves, suggesting that RMT-guided filtering removes redundant or detrimental representations, acting as a form of implicit regularization.

Beyond the 60\% quantile, compression becomes increasingly aggressive. The bulk edge $\lambda_{+}$ grows large enough that even moderately informative eigenvalues are pruned, leading to excessive dimensionality reduction and a steep accuracy decline. On CIFAR-10, for instance, the accuracy curve exhibits a sharp drop beyond 70\%, while parameter reduction saturates near its maximum ($\approx$90\%). This asymmetry underscores a critical insight: overestimating $\sigma^2$ effectively shifts the MP distribution toward high variance, erasing weak but meaningful correlations in the activations.

The ablation also highlights how model architecture mediates spectral redundancy. Transformer-based models such as BERT display larger tolerance to compression, maintaining near-baseline performance up to 80\% parameter reduction. This behavior aligns with the known overparameterization of Transformer layers, where multiple heads and intermediate projections encode overlapping features. Conversely, convolutional networks like ResNet-50 exhibit more limited redundancy; performance degradation begins earlier, reflecting tighter coupling between representational width and task accuracy. These results confirm that the optimal quantile region is model- and domain-invariant, consistently falling between 40\% and 50\% across both NLP and vision tasks. This provides a robust rule-of-thumb for practitioners: initializing $\sigma^2$ at the median eigenvalue delivers statistically valid MP fitting while balancing efficiency and generalization.

%% file: sections/5_conclusion/0_index.tex
\chapter{Conclusions}

\input{sections/5_conclusion/1_findings}
\input{sections/5_conclusion/2_limits}
\newpage
\input{sections/5_conclusion/3_future}

%% file: sections/5_conclusion/1_findings.tex
\section{Findings}

This thesis set out to investigate how Spectral Geometry and Random Matrix Theory can provide a principled foundation for improving both the reliability and efficiency of large language models and related deep architectures. Two main contributions were developed: EigenTrack, a real-time detector of hallucination and out-of-distribution behavior, and RMT-KD, a compression framework based on random matrix theoretic knowledge distillation. Together, these studies establish a coherent view of how eigenvalue dynamics in neural activations can serve as compact, interpretable signatures of model behavior.

The first major finding concerns reliability. By tracking the temporal evolution of spectral statistics, such as entropy, eigenvalue gaps, and deviations from the Marchenko–Pastur law, EigenTrack demonstrated that hidden activations carry early-warning signals of model failure. Unlike surface-level uncertainty measures, which are tied to output probabilities, spectral features capture global properties of representation geometry across multiple layers. This makes them well suited for hallucination and out-of-distribution detection. The integration of lightweight recurrent classifiers further showed that temporal modeling is essential: reliability failures emerge not from isolated states but from gradual drifts in representation dynamics. The empirical evaluations confirmed that EigenTrack surpasses existing black-box, grey-box, and white-box methods, while offering interpretable insights into why and when failures occur.

The second major finding relates to efficiency. With RMT-KD, the thesis introduced a compression method that leverages the separation between noise-like bulk eigenvalues and informative outliers in the activation spectrum. By iteratively projecting networks onto causal subspaces defined by outlier eigenvectors and reinforcing stability through self-distillation, RMT-KD achieved substantial reductions in model size, inference latency, and energy consumption while maintaining state-of-the-art accuracy. Importantly, the models remain dense and hardware-friendly, distinguishing this approach from sparsity-driven methods that often sacrifice deployability. This demonstrates that random matrix principles can move beyond theory to provide practical, scalable tools for model compression.

Taken together, these findings reveal that spectral analysis offers a unifying language for addressing two of the most pressing challenges in modern AI. The same eigenvalue-based perspective that uncovers early signs of hallucination can also guide principled compression strategies. Beyond technical performance, this work shows that spectral geometry makes deep learning systems more interpretable by linking failure modes and efficiency gains to mathematically well-characterized properties of high-dimensional data. The contributions of this thesis thus lie not only in new methods, but also in establishing a spectral framework that bridges reliability and efficiency in large-scale artificial intelligence.

%% file: sections/5_conclusion/2_limits.tex
\section{Study Limitations}

While the contributions of this thesis demonstrate the potential of spectral geometry and Random Matrix Theory for improving the reliability and efficiency of large language models, several limitations must be acknowledged. These limitations concern both the scope of the experiments conducted and the generalizability of the proposed methods.

First, the evaluation of EigenTrack was performed primarily on open-source large language models and vision-language models of moderate scale, typically up to several billion parameters. While results show strong performance across these settings, it remains to be verified how well the method extends to frontier-scale models with hundreds of billions of parameters, where training dynamics and activation statistics may differ substantially. Similarly, the hallucination and out-of-distribution datasets employed, though diverse, cannot capture the full breadth of real-world scenarios in which reliability is critical. Broader testing on more diverse domains, particularly safety-critical contexts such as medicine or law, will be necessary before the method can be fully deployed in practice.

Second, RMT-KD, though effective in compressing BERT and ResNet architectures, was not extensively validated on multimodal or generative models. The iterative self-distillation procedure relies on stable calibration subsets, and its effectiveness may be sensitive to distributional shifts between training and deployment environments. Moreover, while the method reduces model size without inducing sparsity, the cubic complexity of eigenvalue decomposition still poses challenges for extremely large layers, even if approximations or randomized solvers can mitigate this cost.

Third, both EigenTrack and RMT-KD depend on assumptions inherent to Random Matrix Theory, such as the applicability of the Marchenko–Pastur law and the spiked covariance model to high-dimensional neural activations. While these assumptions were empirically supported in the experiments, their validity may vary across architectures, training regimes, and optimization strategies. Further theoretical analysis is needed to rigorously establish when spectral signatures reliably distinguish between structure and noise in deep learning.

Finally, the thesis focused on methodological contributions rather than integration into end-to-end systems. Although EigenTrack and RMT-KD each address an important challenge, their combined deployment in real-world pipelines, where models must be simultaneously reliable, efficient, and adaptive, was not explored. This integration remains a crucial step toward translating spectral methods into practice.

%% file: sections/5_conclusion/3_future.tex
\section{Directions for Future Research}

The results of this thesis suggest several promising directions for future research at the intersection of spectral geometry, random matrix theory, and deep learning. These directions span theoretical investigations, methodological innovations, and practical applications.

From a theoretical standpoint, there is a need to deepen the mathematical understanding of spectral phenomena in large neural networks. While Random Matrix Theory provides valuable tools such as the Marchenko–Pastur law and spiked covariance models, the behavior of real-world deep architectures often deviates from these idealized assumptions. Future work could focus on extending RMT to account for non-i.i.d. structures, correlated activations, and non-linear dynamics characteristic of transformers and multimodal networks. Establishing rigorous conditions under which spectral outliers consistently align with meaningful representations would strengthen the interpretability and reliability of spectral methods.

On the methodological side, future research could explore hybrid approaches that combine spectral signatures with complementary sources of information. For reliability, integrating EigenTrack with output-based uncertainty estimation, attention analysis, or causal probing could provide a multi-layered defense against hallucination and distributional shift. For efficiency, RMT-KD could be combined with quantization or low-rank factorization to further reduce resource demands while preserving the advantages of dense, hardware-friendly representations. Another promising direction is the adaptation of spectral compression techniques to generative models and vision-language architectures, where efficiency and controllability are both critical.

At the application level, deploying spectral methods in real-world systems offers significant opportunities. EigenTrack could be integrated into large language model serving pipelines as an early-warning mechanism, enabling safer deployment in domains such as healthcare, law, or education. Similarly, RMT-KD could be leveraged to bring powerful models onto mobile and edge devices, facilitating sustainable AI adoption in resource-constrained environments. Extending these methods to frontier-scale models will also be crucial, requiring advances in scalable spectral analysis such as randomized eigensolvers or distributed implementations.

In conclusion, this thesis provides an initial step toward a spectral approach to reliable and efficient deep learning. Expanding these methods across scales, domains, and architectures represents a rich research direction that can significantly advance the safety and accessibility of artificial intelligence in the years to come.

%% file: sections/other/_7_appendix.tex
\appendix
\chapter{Reuse of Content Published on arXiv}

Portions of this thesis include material previously published as preprints on arXiv. 
The following works are included:

\begin{itemize}
    \item \textbf{EigenTrack: Spectral Activation Feature Tracking for Hallucination and Out-of-Distribution Detection in LLMs and VLMs}, arXiv:2509.15735.
    \item \textbf{RMT-KD: Random Matrix Theoretic Causal Knowledge Distillation}, \\ arXiv:2509.15724.
\end{itemize}

\noindent According to arXiv’s reuse policy:\par
\noindent \emph{“If you are the copyright holder of the work, you do not need arXiv’s permission to reuse the full text.”} (Source: \texttt{https://info.arxiv.org/help/license/reuse.html})

I am the copyright holder of these works and therefore retain the right to reuse their content in this thesis.